\documentclass[11pt,twoside]{article}

\usepackage{amsfonts,epsfig,graphicx,subfigure}
\usepackage{amsmath,amssymb,amsthm,latexsym}
\usepackage{fullpage}
\usepackage[normalem]{ulem}
\usepackage{epic,eepic}
\usepackage{color}

\usepackage{epsf}
\usepackage{graphics}
\usepackage{epic}
\usepackage{eepic}
\usepackage{eepicemu}
\usepackage{psfrag}
\usepackage{hyperref}

\newcommand{\opnorm}[1]{\ensuremath{\matsnorm{#1}{\myop}}}

\newcommand{\myop}{\ensuremath{\operatorname{op}}}
\newcommand{\argmin}{\operatornamewithlimits{argmin}}
\newcommand{\RegInfConvPlain}{\ensuremath{\Phi}}
\newcommand{\RegInfConv}[1]{\RegInfConvPlain(#1)}
\newcommand{\RegInfConvSq}[1]{\ensuremath{\RegInfConvPlain^2}(#1)}
\newcommand{\BEE}{\ensuremath{B}}
\newcommand{\BEESTAR}{\ensuremath{\BEE^*}}
\newcommand{\EX}{\ensuremath{X}}
\newcommand{\ex}{\EX}
\newcommand{\mineig}{\ensuremath{\sigma_{\min}}}
\newcommand{\maxeig}{\ensuremath{\sigma_{\max}}}
\newcommand{\maxcol}{\ensuremath{\kappa_{\max}}}

\theoremstyle{plain}

\newtheorem{theo}{Theorem}[section]

\newtheorem{lem}{Lemma}[section]
\newtheorem{prop}{Proposition}[section]
\newtheorem{cor}{Corollary}[section]

\theoremstyle{definition} 

\newtheorem{nota}{Notation}[section]
\newtheorem{de}{Definition}[section]
\newtheorem{exa}{Example}[section]
\newtheorem{as}{Assumption}[section]
\newtheorem{alg}{Algorithm}[section]

\newcommand{\btheo}{\begin{theo}}
\newcommand{\bde}{\begin{de}}
\newcommand{\ble}{\begin{lem}}
\newcommand{\bpr}{\begin{prop}}
\newcommand{\bno}{\begin{nota}}
\newcommand{\bex}{\begin{exa}}
\newcommand{\bcor}{\begin{cor}}
\newcommand{\spro}{\begin{proof}}
\newcommand{\bas}{\begin{as}}
\newcommand{\balg}{\begin{alg}}

\newcommand{\etheo}{\end{theo}}
\newcommand{\ede}{\end{de}}
\newcommand{\ele}{\end{lem}}
\newcommand{\epr}{\end{prop}}
\newcommand{\eno}{\end{nota}}
\newcommand{\eex}{\end{exa}}
\newcommand{\ecor}{\end{cor}}
\newcommand{\fpro}{\end{proof}}
\newcommand{\eas}{\end{as}}
\newcommand{\ealg}{\end{alg}}

\theoremstyle{plain}

\newtheorem{theos}{Theorem}
\newtheorem{props}{Proposition}
\newtheorem{lems}{Lemma}
\newtheorem{cors}{Corollary}

\theoremstyle{definition}
\newtheorem{exas}{Example}
\newtheorem{algs}{Algorithm}
\newtheorem{asss}{Assumption}
\newtheorem{defns}{Definition}

\newcommand{\btheos}{\begin{theos}}
\newcommand{\etheos}{\end{theos}}
\newcommand{\bprops}{\begin{props}}
\newcommand{\eprops}{\end{props}}
\newcommand{\bdes}{\begin{defns}}
\newcommand{\edes}{\end{defns}}
\newcommand{\blems}{\begin{lems}}
\newcommand{\elems}{\end{lems}}
\newcommand{\bcors}{\begin{cors}}
\newcommand{\ecors}{\end{cors}}
\newcommand{\bexs}{\begin{exas}}
\newcommand{\eexs}{\end{exas}}
\newcommand{\balgs}{\begin{algs}}
\newcommand{\ealgs}{\end{algs}}
\newcommand{\bass}{\begin{asss}}
\newcommand{\eass}{\end{asss}}

\newlength{\widebarargwidth}
\newlength{\widebarargheight}
\newlength{\widebarargdepth}

\newcommand{\matsnorm}[2]{|\!|\!| #1 | \! | \!|_{{#2}}}

\newcommand{\real}{\ensuremath{\mathbb{R}}}

\newenvironment{carlist}
 {\begin{list}{$\bullet$}
 {\setlength{\topsep}{0in} \setlength{\partopsep}{0in}
  \setlength{\parsep}{0in} \setlength{\itemsep}{\parskip}
  \setlength{\leftmargin}{0.07in} \setlength{\rightmargin}{0.08in}
  \setlength{\listparindent}{0in} \setlength{\labelwidth}{0.08in}
  \setlength{\labelsep}{0.1in} \setlength{\itemindent}{0in}}}
 {\end{list}}

\newcommand{\bcar}{\begin{carlist}}
\newcommand{\ecar}{\end{carlist}}

\newcommand{\tracer}[2]{\ensuremath{\langle \!\langle {#1}, \; {#2}
\rangle \!\rangle}}

\newcommand{\numobs}{\ensuremath{n}}

\newcommand{\lossrsc}{\ensuremath{\gamma}}
\newcommand{\FUNLOW}{{\tau_\numobs}}
\newcommand{\FUNLOWL}{{\FUNLOW}}

\makeatletter
\long\def\@makecaption#1#2{
        \vskip 0.8ex
        \setbox\@tempboxa\hbox{\small {\bf #1:} #2}
        \parindent 1.5em  
        \dimen0=\hsize
        \advance\dimen0 by -3em
        \ifdim \wd\@tempboxa >\dimen0
                \hbox to \hsize{
                        \parindent 0em
                        \hfil 
                        \parbox{\dimen0}{\def\baselinestretch{0.96}\small
                                {\bf #1.} #2
                                } 
                        \hfil}
        \else \hbox to \hsize{\hfil \box\@tempboxa \hfil}
        \fi
        }
\makeatother

\long\def\comment#1{}

\makeatletter
\def\@cite#1#2{[\if@tempswa #2 \fi #1]}
\makeatother

\makeatletter
\long\def\barenote#1{
    \insert\footins{\footnotesize
    \interlinepenalty\interfootnotelinepenalty 
    \splittopskip\footnotesep
    \splitmaxdepth \dp\strutbox \floatingpenalty \@MM
    \hsize\columnwidth \@parboxrestore
    {\rule{\z@}{\footnotesep}\ignorespaces
      #1\strut}}}
\makeatother

\newcommand{\LowRank}{\ensuremath{\Theta}}
\newcommand{\Sparse}{\ensuremath{\Gamma}}

\newcommand{\LowRankHat}{\ensuremath{\widehat{\LowRank}}}
\newcommand{\SparseHat}{\ensuremath{\widehat{\Sparse}}}

\newcommand{\TrueLowRank}{\ensuremath{{\LowRank^\star}}}
\newcommand{\TrueSparse}{\ensuremath{{\Sparse^\star}}}

\newcommand{\DelL}{\ensuremath{\Delta^\LowRank}}
\newcommand{\DelS}{\ensuremath{\Delta^\Sparse}}

\newcommand{\DelHatL}{\ensuremath{\widehat{\Delta}^\LowRank}}
\newcommand{\DelHatS}{\ensuremath{\widehat{\Delta}^\Sparse}}

\newcommand{\DelHat}{\ensuremath{\widehat{\Delta}}}

\newcommand{\nuclear}[1]{\ensuremath{\matsnorm{#1}{\operatorname{N}}}}
\newcommand{\frob}[1]{\ensuremath{\matsnorm{#1}{\operatorname{F}}}}

\newcommand{\ellnorm}[1]{\ensuremath{\| #1\|_1}}
\newcommand{\infnorm}[1]{\ensuremath{\| #1\|_{\infty}}}

\newcommand{\Loss}{\ensuremath{\mathcal{L}}}

\newcommand{\defn}{\ensuremath{: \, =}}

\newcommand{\mdim}{\ensuremath{d}}

\newcommand{\Exs}{\ensuremath{\mathbb{E}}}
\newcommand{\mprob}{\ensuremath{\mathbb{P}}}

\newcommand{\kdim}{\ensuremath{s}}

\newcommand{\rdim}{\ensuremath{r}}

\newcommand{\plaincon}{\ensuremath{c}}

\newcommand{\ALCON}{\ensuremath{\alpha}}

\newcommand{\regpar}{\ensuremath{\lambda_\mdim}}

\newcommand{\BIGREG}{\ensuremath{\mathcal{Q}}}
\newcommand{\Ymat}{\ensuremath{Y}}
\newcommand{\Wmat}{\ensuremath{W}}
\newcommand{\Sset}{\ensuremath{S}}

\newcommand{\rank}{\ensuremath{\operatorname{rank}}}

\newcommand{\noisevar}{\ensuremath{\nu}}

\newcommand{\Ltil}{\ensuremath{\widetilde{\LowRank}}}
\newcommand{\Stil}{\ensuremath{\widetilde{\Sparse}}}
\newcommand{\MINIMAX}{\ensuremath{\mathfrak{M}}}

\newcommand{\DALCON}{\ensuremath{\ALCON}}

\newcommand{\tworeg}{\ensuremath{\mu_\mdim}}
\newcommand{\twopar}{\tworeg}

\newcommand{\supp}{\ensuremath{\operatorname{supp}}}

\newcommand{\DelHatLA}{\ensuremath{\DelHatL_A}}
\newcommand{\DelHatLB}{\ensuremath{\DelHatL_B}}

\newcommand{\pdim}{\ensuremath{p}}

\newcommand{\ellone}[1]{\ellnorm{#1}}

\newcommand{\gennorm}[2]{\ensuremath{\|#1\|_{#2}}}

\newcommand{\mdima}{\ensuremath{{\mdim_1}}}
\newcommand{\mdimb}{\ensuremath{{\mdim_2}}}

\newcommand{\DCON}{\ensuremath{\kappa_\mdim}}
\newcommand{\Dregplain}{\ensuremath{\mathcal{R}^*}}
\newcommand{\Reg}[1]{\ensuremath{\Regplain(#1)}}
\newcommand{\Dreg}[1]{\ensuremath{\Dregplain(#1)}}
\newcommand{\Regplain}{\ensuremath{\mathcal{R}}}
\newcommand{\Dualreg}[1]{\Dreg{#1}}

\newcommand{\goodendex}{\ensuremath{\clubsuit}}

\newcommand{\Model}{\ensuremath{\mathbb{M}}}
\newcommand{\ModelPerp}{\ensuremath{{\Model^\perp}}}
\newcommand{\ProjPerp}[1]{\ensuremath{\Pi_{\ModelPerp}(#1)}}

\newcommand{\Compat}{\ensuremath{\Psi}}

\newcommand{\genspike}{\ensuremath{\varphi_{\Regplain}}}

\newcommand{\trace}{\ensuremath{\operatorname{trace}}}

\newcommand{\DelSModel}{\DelHatS_\Model}
\newcommand{\DelSPerp}{\DelHatS_{\ModelPerp}}

\newcommand{\MyModelSparse}{\Gamma^*_{S}}
\newcommand{\MyNoModelSparse}{\Gamma^*_{S^c}}

\newcommand{\twonorm}[1]{\ensuremath{\|#1\|_2}}
\newcommand{\half}{\ensuremath{\frac{1}{2}}}

\newcommand{\HACK}{\ensuremath{\mathcal{K}}}

\newcommand{\FERRSQ}[2]{\ensuremath{e^2(#1, \, #2)}}

\newcommand{\widgraph}[2]{\includegraphics[keepaspectratio,width=#1]{#2}}

\newcommand{\Colset}{\ensuremath{C}}
\newcommand{\ColSet}{\Colset}

\newcommand{\Fam}{\ensuremath{\mathcal{F}}}
\newcommand{\Famsp}{\ensuremath{\Fam_{\operatorname{sp}}}}
\newcommand{\Famcol}{\ensuremath{\Fam_{\operatorname{col}}}}
\newcommand{\colsupp}{\ensuremath{\operatorname{colsupp}}}
\newcommand{\kull}[2]{\ensuremath{\mathbb{D}(#1 \,\| #2)}}

\newcommand{\Ztil}{\ensuremath{\widetilde{Z}}}

\definecolor{OliveGreen}{cmyk}{0.91,0,0.88,0.2} 

\newcommand{\na}{\ensuremath{n_1}}
\newcommand{\nb}{\ensuremath{n_2}}
\newcommand{\XopPlain}{\ensuremath{\mathfrak{X}}}
\newcommand{\Xop}[1]{\ensuremath{\XopPlain(#1)}}
\newcommand{\XopDual}[1]{\ensuremath{\XopPlain^*(#1)}}

\newcommand{\EXCESSALL}{2 \regpar \sum_{j=\rdim+1}^\mdim \sigma_j(\TrueLowRank) +
  2 \, \tworeg \Reg{\TrueSparse_\ModelPerp}}
\newcommand{\EXCESSNOTWO}{\regpar \sum_{j=\rdim+1}^\mdim \sigma_j(\TrueLowRank) +
  \tworeg \Reg{\TrueSparse_\ModelPerp}}
\newcommand{\EXCESS}{\sum_{j=\rdim+1}^\mdim \sigma_j(\TrueLowRank) +
  \frac{\tworeg}{\regpar} \, \Reg{\TrueSparse_\ModelPerp}} 

\newcommand{\betastar}{\ensuremath{\beta^*}}

\newcommand{\myone}{\ensuremath{\vec{1}}}

\newcommand{\BAD}{\ensuremath{B^*}}

\usepackage{ifthen}

\newcommand{\MYELLONE}[1]{\ensuremath{\|#1\|_1}}

\begin{document}

\begin{center}

{\bf{\LARGE{Noisy matrix decomposition via convex relaxation:
      Optimal rates in high dimensions}}}

\vspace*{.2in}

{\large{
\begin{tabular}{ccc}
Alekh Agarwal$^\dagger$ & Sahand Negahban$^\dagger$ & Martin
J. Wainwright$^{\dagger,\star}$ \\
{\texttt{alekh@eecs.berkeley.edu}} & {\texttt{sahand@eecs.berkeley.edu}} &
{\texttt{wainwrig@stat.berkeley.edu}}
\end{tabular}
}}

\vspace*{.2in}

\begin{tabular}{ccc}
Department of EECS$^\dagger$ & & 
Department of Statistics$^\star$ \\
UC Berkeley & & UC Berkeley \\
\end{tabular}
\vspace*{.2in}

February 2012 (revision of February 2011 version)

\vspace*{.2in}

\begin{tabular}{c}
Technical Report, \\
Department of Statistics,  UC Berkeley
\end{tabular}

\vspace*{.1in}

\begin{abstract}
  We analyze a class of estimators based on convex relaxation for
  solving high-dimensional matrix decomposition problems. The
  observations are noisy realizations of a linear transformation
  $\mathfrak{X}$ of the sum of an (approximately) low rank matrix
  $\Theta^\star$ with a second matrix $\Gamma^\star$ endowed with a
  complementary form of low-dimensional structure; this set-up
  includes many statistical models of interest, including forms of
  factor analysis, multi-task regression with shared structure, and
  robust covariance estimation. We derive a general theorem that gives
  upper bounds on the Frobenius norm error for an estimate of the pair
  $(\Theta^\star, \Gamma^\star)$ obtained by solving a convex
  optimization problem that combines the nuclear norm with a general
  decomposable regularizer.  Our results are based on imposing a
  ``spikiness'' condition that is related to but milder than singular
  vector incoherence. We specialize our general result to two cases
  that have been studied in past work: low rank plus an entrywise
  sparse matrix, and low rank plus a columnwise sparse matrix.  For
  both models, our theory yields non-asymptotic Frobenius error bounds
  for both deterministic and stochastic noise matrices, and applies to
  matrices $\Theta^\star$ that can be exactly or approximately low
  rank, and matrices $\Gamma^\star$ that can be exactly or
  approximately sparse.  Moreover, for the case of stochastic noise
  matrices and the identity observation operator, we establish matching
  lower bounds on the minimax error, showing that our results cannot
  be improved beyond constant factors.  The sharpness of our
  theoretical predictions is confirmed by numerical simulations.
\end{abstract}

\end{center}


\section{Introduction}

The focus of this paper is a class of high-dimensional matrix
decomposition problems of the following variety.  Suppose that we
observe a matrix $\Ymat \in \real^{\mdima \times \mdimb}$ that is
(approximately) equal to the sum of two unknown matrices: how to
recover good estimates of the pair?  Of course, this problem is
ill-posed in general, so that it is necessary to impose some kind of
low-dimensional structure on the matrix components, one example being
rank constraints.  The framework of this paper supposes that one
matrix component (denoted $\TrueLowRank$) is low-rank, either exactly
or in an approximate sense, and allows for general forms of
low-dimensional structure for the second component $\TrueSparse$.  Two
particular cases of structure for $\TrueSparse$ that have been
considered in past work are elementwise
sparsity~\cite{Chand09,ChandraPaWi10,CandesLiMaWr2009} and column-wise
sparsity~\cite{MccoyTr2010,XuCaSa2010}.

Problems of matrix decomposition are motivated by a variety of
applications.  Many classical methods for dimensionality reduction,
among them factor analysis and principal components analysis (PCA),
are based on estimating a low-rank matrix from data.  Different forms
of robust PCA can be formulated in terms of matrix decomposition using
the matrix $\TrueSparse$ to model the gross errors~\cite{Chand09,
  CandesLiMaWr2009, XuCaSa2010}.  Similarly, certain problems of
robust covariance estimation can be described using matrix
decompositions with a column/row-sparse structure, as we describe in
this paper.  The problem of low rank plus sparse matrix decomposition
also arises in Gaussian covariance selection with hidden
variables~\cite{ChandraPaWi10}, in which case the inverse covariance
of the observed vector can be decomposed as the sum of a sparse matrix
with a low rank matrix.  Matrix decompositions also arise in
multi-task regression~\cite{YuaEkiLuMon07,NegWai09, RohTsy10}, which
involve solving a collection of regression problems, referred to as
tasks, over a common set of features.  For some features, one expects
their weighting to be preserved across features, which can be modeled
by a low-rank constraint, whereas other features are expected to vary
across tasks, which can be modeled by a sparse
component~\cite{BliMcdPer06,AndZha05}.  See
Section~\ref{SecApplications} for further discussion of these
motivating applications.

In this paper, we study a noisy linear observation that can be used to
describe a number of applications in a unified way.  Let $\XopPlain$
be a linear operator that maps matrices in $\real^{\mdima \times
  \mdimb}$ to matrices in $\real^{\na \times \nb}$.  In the simplest
of cases, this observation operator is simply the identity mapping, so
that we necessarily have $\na = \mdima$ and $\nb = \mdimb$.  However,
as we discuss in the sequel, it is useful for certain applications,
such as multi-task regression, to consider more general linear
operators of this form.  Hence, we study the problem matrix
decomposition for the general linear observation model
\begin{align}
\label{EqnLinearObs}
\Ymat & = \Xop{\TrueLowRank + \TrueSparse} + \Wmat,
\end{align}
where $\TrueLowRank$ and $\TrueSparse$ are unknown $\mdima \times
\mdimb$ matrices, and $\Wmat \in \real^{\na \times \nb}$ is some type
of observation noise; it is potentially dense, and can either be
deterministic or stochastic.  The matrix $\TrueLowRank$ is assumed to
be either exactly low-rank, or well-approximated by a low-rank matrix,
whereas the matrix $\TrueSparse$ is assumed to have a complementary
type of low-dimensional structure, such as sparsity.  As we discuss in
Section~\ref{SecApplications}, a variety of interesting statistical
models can be formulated as instances of the observation
model~\eqref{EqnLinearObs}.  Such models include versions of factor
analysis involving non-identity noise matrices, robust forms of
covariance estimation, and multi-task regression with some features
shared across tasks, and a sparse subset differing across tasks.
Given this observation model, our goal is to recover accurate
estimates of the decomposition $(\TrueLowRank, \TrueSparse)$ based on
the noisy observations $\Ymat$.  In this paper, we analyze simple
estimators based on convex relaxations involving the nuclear norm, and
a second general norm $\Regplain$.

Most past work on the model~\eqref{EqnLinearObs} has focused on the
noiseless setting ($\Wmat = 0$), and for the identity observation
operator (so that $\Xop{\TrueLowRank + \TrueSparse} = \TrueLowRank +
\TrueSparse$).  Chandrasekaran et al.~\cite{Chand09} studied the case
when $\TrueSparse$ is assumed to sparse, with a relatively small
number \mbox{$\kdim \ll \mdima \mdimb$} of non-zero entries.  In the
noiseless setting, they gave sufficient conditions for exact recovery
for an adversarial sparsity model, meaning the non-zero positions of
$\TrueSparse$ can be arbitrary. Subsequent work by Candes et
al.~\cite{CandesLiMaWr2009} analyzed the same model but under an
assumption of random sparsity, meaning that the non-zero positions are
chosen uniformly at random.  In very recent work, Xu et
al.~\cite{XuCaSa2010} have analyzed a different model, in which the
matrix $\TrueSparse$ is assumed to be columnwise sparse, with a
relatively small number $\kdim \ll \mdimb$ of non-zero columns.  Their
analysis guaranteed approximate recovery for the low-rank matrix, in
particular for the uncorrupted columns.  After initial posting of this
work, we became aware of recent work by Hsu et al.~\cite{HUPAPER}, who
derived Frobenius norm error bounds for the case of exact elementwise
sparsity.  As we discuss in more detail in Section~\ref{SecEllOne}, in
this special case, our bounds are based on milder conditions, and
yield sharper rates for problems where the rank and sparsity scale
with the dimension.

Our main contribution is to provide a general oracle-type result
(Theorem~\ref{ThmConvergence}) on approximate recovery of the unknown
decomposition from noisy observations, valid for structural
constraints on $\TrueSparse$ imposed via a decomposable regularizer.
The class of decomposable regularizers, introduced in past work by
Negahban et al.~\cite{NegRavWaiYu09}, includes the elementwise
$\ell_1$-norm and columnwise $(2,1)$-norm as special cases, as well as
various other regularizers used in practice.  Our main result is
stated in Theorem~\ref{ThmConvergence}: it provides finite-sample
guarantees for estimates obtained by solving a class of convex
programs formed using a composite regularizer.  The resulting
Frobenius norm error bounds consist of multiple terms, each of which
has a natural interpretation in terms of the estimation and
approximation errors associated with the sub-problems of recovering
$\TrueLowRank$ and $\TrueSparse$.  We then specialize
Theorem~\ref{ThmConvergence} to the case of elementwise or columnwise
sparsity models for $\TrueSparse$, thereby obtaining recovery
guarantees for matrices $\TrueLowRank$ that may be either exactly or
approximately low-rank, as well as matrices $\TrueSparse$ that may be
either exactly or approximately sparse.  We provide non-asymptotic
error bounds for general noise matrices $\Wmat$ both for elementwise
and columnwise sparse models (see Corollaries~\ref{CorSparseDet}
through Corollary~\ref{CorColNoisy}).  To the best of our knowledge,
these are the first results that apply to this broad class of models,
allowing for noisiness ($\Wmat \neq 0$) that is either stochastic or
deterministic, matrix components that are only approximately low-rank
and/or sparse, and general forms of the observation operator
$\XopPlain$.

In addition, the error bounds obtained by our analysis are sharp, and
cannot be improved in general.  More precisely, for the case of
stochastic noise matrices and the identity observation operator, we
prove that the squared Frobenius errors achieved by our estimators are
minimax-optimal (see Theorem~\ref{ThmMinimax}).  An interesting
feature of our analysis is that, in contrast to previous
work~\cite{Chand09,XuCaSa2010,CandesLiMaWr2009}, we do \emph{not}
impose incoherence conditions on the singular vectors of
$\TrueLowRank$; rather, we control the interaction with a milder
condition involving the dual norm of the regularizer. In the special
case of elementwise sparsity, this dual norm enforces an upper bound
on the ``spikiness'' of the low-rank component, and has proven useful
in the related setting of noisy matrix completion~\cite{NegWai10b}.
This constraint is not strong enough to guarantee identifiability of
the models (and hence exact recovery in the noiseless setting), but it
does provide a bound on the degree of non-identifiability.  We show
that this same term arises in both the upper and lower bounds on the
problem of approximate recovery that is of interest in the noisy
setting.

The remainder of the paper is organized as follows.  In
Section~\ref{SecMatDecomp}, we set up the problem in a precise way,
and describe the estimators.  Section~\ref{SecMain} is devoted to the
statement of our main result on achievability, as well as its various
corollaries for special cases of the matrix decomposition problem.  We
also state a matching lower bound on the minimax error for matrix
decomposition with stochastic noise.  In Section~\ref{SecExperiment},
we provide numerical simulations that illustrate the sharpness of our
theoretical predictions.  Section~\ref{SecProofs} is devoted to the
proofs of our results, with certain more technical aspects of the
argument deferred to the appendices, and we conclude with a discussion
in Section~\ref{SecDiscuss}.

\paragraph{Notation:}  For the reader's convenience, we summarize here
some of the standard notation used throughout this paper.  For any
matrix $M \in \real^{\mdima \times \mdimb}$, we define the Frobenius
norm $\frob{M} \defn \sqrt{\sum_{j=1}^\mdima \sum_{k=1}^\mdimb
  M^2_{jk}}$, corresponding to the ordinary Euclidean norm of its
entries.  We denote its singular values by $\sigma_1(M) \geq
\sigma_2(M) \geq \cdots \geq \sigma_d(M) \geq 0$, where $d =
\min\{\mdima, \mdimb \}$.  Its nuclear norm is given by $\nuclear{M} =
\sum_{j=1}^d \sigma_j(M)$.

\section{Convex relaxations and matrix decomposition}
\label{SecMatDecomp}

In this paper, we consider a family of regularizers formed by a
combination of the \emph{nuclear norm} $\nuclear{\LowRank} \defn
\sum_{j = 1}^{\min \{\mdima, \mdimb \}} \sigma_j(\LowRank)$, which
acts as a convex surrogate to a rank constraint for $\TrueLowRank$
(e.g., see Recht et al.~\cite{RecFazPar10} and references therein),
with a \emph{norm-based regularizer} \mbox{$\Regplain: \real^{\mdima
    \times \mdimb} \rightarrow \real_+$} used to constrain the
structure of $\TrueSparse$.  We provide a general theorem applicable
to a class of regularizers $\Regplain$ that satisfy a certain
decomposability property~\cite{NegRavWaiYu09}, and then consider in
detail a few particular choices of $\Regplain$ that have been studied
in past work, including the elementwise $\ell_1$-norm, and the
columnwise $(2,1)$-norm (see Examples~\ref{ExaEllOne}
and~\ref{ExaColSparse} below).

\subsection{Some motivating applications}
\label{SecApplications}

We begin with some motivating applications for the general linear
observation model with noise~\eqref{EqnLinearObs}.

\bexs[Factor analysis with sparse noise]
\label{ExaFactor}
In a factor analysis model, random vectors $Z_i \in \real^\mdim$ are
assumed to be generated in an i.i.d. manner from the model
\begin{align}
\label{EqnFactor}
Z_i & = L U_i + \varepsilon_i, \quad \mbox{for $i = 1, 2, \ldots,
  \numobs$},
\end{align}
where $L \in \real^{\mdima \times r}$ is a loading matrix, and the
vectors $U_i \sim N(0, I_{r \times r})$ and $\varepsilon_i \sim N(0,
\TrueSparse)$ are independent.  Given $\numobs$ i.i.d.  samples from
the model~\eqref{EqnFactor}, the goal is to estimate the loading
matrix $L$, or the matrix $L L^T$ that projects onto column span of
$L$.  A simple calculation shows that the covariance matrix of $Z_i$
has the form $\Sigma = L L^T + \TrueSparse$.  Consequently, in the
special case when $\TrueSparse = \sigma^2 I_{\mdim \times \mdim}$,
then the range of $L$ is spanned by the top $\rdim$ eigenvectors of
$\Sigma$, and so we can recover it via standard principal components
analysis.

In other applications, we might no longer be guaranteed that
$\TrueSparse$ is the identity, in which case the top $r$ eigenvectors
of $\Sigma$ need not be close to column span of $L$.  Nonetheless,
when $\TrueSparse$ is a sparse matrix, the problem of estimating $L
L^T$ can be understood as an instance of our general observation
model~\eqref{EqnLinearObs} with $\mdima = \mdimb = \mdim$, and the
identity observation operator $\XopPlain$ (so that $\na = \nb =
\mdim$).  In particular, if the let the observation matrix $Y \in
\real^{\mdim \times \mdim}$ be the sample covariance matrix
$\frac{1}{\numobs} \sum_{i-1}^\numobs Z_i Z_i^T$, then some algebra
shows that $Y = \TrueLowRank + \TrueSparse + W$, where $\TrueLowRank =
L L^T$ is of rank $r$, and the random matrix $W$ is a re-centered form
of Wishart noise~\cite{AndersonStat}---in particular, the zero-mean
matrix
\begin{align}
\label{EqnWishNoise}
W & \defn \frac{1}{\numobs} \sum_{i=1}^\numobs Z_i Z_i^T - \big \{ L
L^T + \TrueSparse \big\}.
\end{align}
When $\TrueSparse$ is assumed to be elementwise sparse (i.e., with
relatively few non-zero entries), then this constraint can be enforced
via the elementwise $\ell_1$-norm (see Example~\ref{ExaEllOne} to
follow).
\hfill \goodendex \eexs

\bexs[Multi-task regression]
\label{ExaMultitask}
Suppose that we are given a collection of $\mdimb$ regression problems
in $\real^\mdima$, each of the form $y_j = X \betastar_j + w_j$ for $j
= 1, 2, \ldots, \mdimb$.  Here each $\betastar_j \in \real^{\mdima}$
is an unknown regression vector, $w_j \in \real^{\numobs}$ is
observation noise, and $X \in \real^{\numobs \times \mdima}$ is the
design matrix.  This family of models can be written in a convenient
matrix form as \mbox{$Y = X B^* + W$,} where $Y = [y_1 \; \cdots \;
  y_\mdimb]$ and $W = [w_1 \; \cdots \; w_\mdimb]$ are both matrices
in $\real^{\numobs \times \mdimb}$ and \mbox{$B^* \defn [\betastar_1
    \; \cdots \; \betastar_\mdimb] \in \real^{\mdima \times \mdimb}$}
is a matrix of regression vectors.  Following standard terminology in
multi-task learning, we refer to each column of $B^*$ as a
\emph{task}, and each row of $B^*$ as a \emph{feature.}

In many applications, it is natural to assume that the feature
weightings---i.e., that is, the vectors $\betastar_j \in
\real^{\mdimb}$---exhibit some degree of shared structure across
tasks~\cite{AndZha05,YuaEkiLuMon07,NegWai09,RohTsy10}. This type of
shared structured can be modeled by imposing a low-rank structure; for
instance, in the extreme case of rank one, it would enforce that each
$\betastar_j$ is a multiple of some common underlying vector.
However, many multi-task learning problems exhibit more complicated
structure, in which some subset of features are shared across tasks,
and some other subset of features vary substantially across
tasks~\cite{AndZha05,Blitzer10}.  For instance, in the Amazon
recommendation system, tasks correspond to different classes of
products, such as books, electronics and so on, and features include
ratings by users.  Some ratings (such as numerical scores) should have
a meaning that is preserved across tasks, whereas other features
(e.g., the label ``boring'') are very meaningful in applications to
some categories (e.g., books) but less so in others (e.g.,
electronics).

This kind of structure can be captured by assuming that the unknown
regression matrix $B^*$ has a low-rank plus sparse
decomposition---namely, $B^* = \TrueLowRank + \TrueSparse$ where
$\TrueLowRank$ is low-rank and $\TrueSparse$ is sparse, with a
relatively small number of non-zero entries, corresponding to
feature/task pairs that that differ significantly from the baseline.
A variant of this model is based on instead assuming that
$\TrueSparse$ is row-sparse, with a small number of non-zero rows. (In
Example~\ref{ExaColSparse} to follow, we discuss an appropriate
regularizer for enforcing such row or column sparsity.)  With this
model structure, we then define the observation operator
\mbox{$\XopPlain: \real^{\mdima \times \mdimb} \rightarrow
  \real^{\numobs \times \mdimb}$} via $A \mapsto X A$, so that $\na =
\numobs$ and $\nb = \mdimb$ in our general notation.  In this way, we
obtain another instance of the linear observation
model~\eqref{EqnLinearObs}.
\hfill \goodendex \eexs 

\vspace*{.01in}

\bexs[Robust covariance estimation]
\label{ExaRobustCovariance}
For $i = 1, 2, \ldots, \numobs$, let $U_i \in \real^\mdim$ be samples
from a zero-mean distribution with unknown covariance matrix
$\TrueLowRank$.  When the vectors $U_i$ are observed without any form
of corruption, then it is straightforward to estimate $\TrueLowRank$
by performing PCA on the sample covariance matrix.  Imagining that $j
\in \{1, 2, \ldots, \mdim \}$ indexes different individuals in the
population, now suppose that the data associated with some subset $S$
of individuals is arbitrarily corrupted.  This adversarial corruption
can be modeled by assuming that we observe the vectors $Z_i = U_i +
v_i$ for $i = 1, \ldots, \numobs$, where each $v_i \in \real^\mdim$ is
a vector supported on the subset $S$.  Letting $Y = \frac{1}{\numobs}
\sum_{i=1}^\numobs Z_i Z_i^T$ be the sample covariance matrix of the
corrupted samples, some algebra shows that it can be decomposed as $Y
= \TrueLowRank + \Delta + \Wmat$, where $\Wmat \defn \frac{1}{\numobs}
\sum_{i=1}^\numobs U_i U_i^T - \TrueLowRank$ is again a type of
re-centered Wishart noise, and the remaining term can be written as
\begin{align}
\label{EqnSparseRC}
\Delta & \defn \frac{1}{\numobs} \sum_{i=1}^\numobs v_i v_i^T +
\frac{1}{\numobs} \sum_{i=1}^\numobs \big( U_i v_i^T + v_i U_i^T).
\end{align}
Note that $\Delta$ itself is not a column-sparse or row-sparse matrix;
however, since each vector $v_i \in \real^\mdim$ is supported only on
some subset $S \subset \{1, 2, \ldots, \mdim \}$, we can write $\Delta
= \TrueSparse + (\TrueSparse)^T$, where $\TrueSparse$ is a
column-sparse matrix with entries only in columns indexed by $S$.
This structure can be enforced by the use of the column-sparse
regularizer~\eqref{EqnColReg}, as described in
Example~\ref{ExaColSparse} to follow.

\hfill \goodendex \eexs

\subsection{Convex relaxation for noisy matrix decomposition}

Given the observation model $Y = \Xop{\TrueLowRank + \TrueSparse} +
\Wmat$, it is natural to consider an estimator based on solving the
regularized least-squares program
\begin{equation*}
\min_{(\LowRank, \Sparse)} \biggr \{ \frac{1}{2} \frob{ \Ymat -
  \Xop{\LowRank + \Sparse}}^2 + \regpar \nuclear{\LowRank} + \tworeg
\Reg{\Sparse} \biggr \}.
\end{equation*}
Here $(\regpar, \tworeg)$ are non-negative regularizer parameters, to
be chosen by the user.  Our theory also provides choices of these
parameters that guarantee good properties of the associated estimator.
Although this estimator is reasonable, it turns out that an additional
constraint yields an equally simple estimator that has attractive
properties, both in theory and in practice. \\

In order to understand the need for an additional constraint, it
should be noted that without further constraints, the
model~\eqref{EqnLinearObs} is unidentifiable, even in the noiseless
setting \mbox{($\Wmat = 0$).}  Indeed, as has been discussed in past
work~\cite{Chand09, CandesLiMaWr2009, XuCaSa2010}, no method can
recover the components $(\TrueLowRank, \TrueSparse)$ unless the
low-rank component is ``incoherent'' with the matrix $\TrueSparse$.
For instance, supposing for the moment that $\TrueSparse$ is a sparse
matrix, consider a rank one matrix with $\Theta^\star_{11} \ne 0$, and
zeros in all other positions.  In this case, it is clearly impossible
to disentangle $\TrueLowRank$ from a sparse matrix.  Past work on both
matrix completion and decomposition~\cite{Chand09, CandesLiMaWr2009,
  XuCaSa2010} has ruled out these types of troublesome cases via
conditions on the singular vectors of the low-rank component
$\TrueLowRank$, and used them to derive sufficient conditions for
exact recovery in the noiseless setting (see the discussion following
Example~\ref{ExaEllOne} for more details).

In this paper, we impose a related but milder condition, previously
introduced in our past work on matrix completion~\cite{NegWai10b},
with the goal of performing approximate recovery. To be clear, this
condition does not guarantee identifiability, but rather provides a
bound on the \emph{radius of non-identifiability.}  It should be noted
that non-identifiability is a feature common to many high-dimensional
statistical models.\footnote{For instance, see the
  paper~\cite{RasWaiYu09} for discussion of non-identifiability in
  high-dimensional sparse regression.} Moreover, in the more realistic
setting of noisy observations and/or matrices that are not exactly
low-rank, such approximate recovery is the best that can be expected.
Indeed, one of our main contributions is to establish
minimax-optimality of our rates, meaning that no algorithm can be
substantially better over the matrix classes that we consider.\\

For a given regularizer $\Regplain$, we define the quantity
$\DCON(\Regplain) \defn \sup_{V \neq 0} \frob{V}/\Reg{V}$, which
measures the relation between the regularizer and the Frobenius norm.
Moreover, we define the associated dual norm
\begin{align}
\Dreg{U} & \defn 
\sup_{\Reg{V} \leq 1} \; \; \tracer{V}{U},
\end{align}
where $\tracer{V}{U} \defn \trace(V^T U)$ is the trace inner product
on the space $\real^{\mdima \times \mdimb}$.  Our estimators are based
on constraining the interaction between the low-rank component and
$\TrueSparse$ via the quantity
\begin{align}
\genspike(\LowRank) & \defn \DCON(\Dregplain) \, \Dualreg{\LowRank}.
\end{align} 
More specifically, we analyze the family of estimators
\begin{align}
\label{EqnGenProg}
\min_{(\LowRank, \Sparse)} \Big \{ \frac{1}{2} \matsnorm{Y - \Xop{\LowRank
  + \Sparse}}{F}^2 \; + \regpar \, \nuclear{\LowRank} + \tworeg \,
\Reg{\Sparse} \Big \},
\end{align}
\mbox{subject to $\genspike(\LowRank) \leq \DALCON$} for some fixed
parameter $\DALCON$.

\subsection{Some examples}

Let us consider some examples to provide intuition for specific forms
of the estimator~\eqref{EqnGenProg}, and the role of the additional
constraint.
\bexs[Sparsity and elementwise $\ell_1$-norm]
\label{ExaEllOne}
Suppose that $\TrueSparse$ is assumed to be sparse, with \mbox{$\kdim
  \ll \mdima \mdimb$} non-zero entries.  In this case, the sum
$\TrueLowRank + \TrueSparse$ corresponds to the sum of a low rank
matrix with a sparse matrix.  Motivating applications include the
problem of factor analysis with a non-identity but sparse noise
covariance, as discussed in Example~\ref{ExaFactor}, as well as
certain formulations of robust PCA~\cite{CandesLiMaWr2009}, and model
selection in Gauss-Markov random fields with hidden
variables~\cite{ChandraPaWi10}.  Given the sparsity of $\TrueSparse$,
an appropriate choice of regularizer is the elementwise $\ell_1$-norm
\begin{align}
\Reg{\Sparse} \; = \; \ellnorm{\Sparse} & \defn \; \sum_{j=1}^\mdima
\sum_{k=1}^\mdimb |\Sparse_{jk}|.
\end{align}
With this choice, it is straightforward to verify that
\begin{align}
\Dreg{Z} \; = \; \infnorm{Z} & \defn \max_{j=1, \ldots, \mdima}
\max_{k = 1, \ldots, \mdimb} |Z_{jk}|,
\end{align}
and moreover, that $\DCON(\Dregplain) = \sqrt{\mdima \mdimb}$.
Consequently, in this specific case, the general convex
program~\eqref{EqnGenProg} takes the form
\begin{align}
\label{EqnSparseProb}
\min_{(\LowRank, \Sparse)} \Big \{ \frac{1}{2} \matsnorm{Y -
  \Xop{\LowRank + \Sparse}}{F}^2 \; + \regpar \, \nuclear{\LowRank} +
\tworeg \, \ellnorm{\Sparse} \Big \} \quad \mbox{such that
  $\infnorm{\LowRank} \leq \frac{\DALCON}{\sqrt{\mdima \, \mdimb}}$.}
\end{align}
The constraint involving $\infnorm{\LowRank}$ serves to control the
``spikiness'' of the low rank component, with larger settings of
$\DALCON$ allowing for more spiky matrices.  Indeed, this type of
spikiness control has proven useful in analysis of nuclear norm
relaxations for noisy matrix completion~\cite{NegWai10b}.  To gain
intuition for the parameter $\DALCON$, if we consider matrices with
$\frob{\LowRank} \approx 1$, as is appropriate to keep a constant
signal-to-noise ratio in the noisy model~\eqref{EqnLinearObs}, then
setting $\DALCON \approx 1$ allows only for matrices for which
$|\LowRank_{jk}| \approx 1/\sqrt{\mdima \mdimb}$ in all entries.  If
we want to permit the maximally spiky matrix with all its mass in a
single position, then the parameter $\DALCON$ must be of the order
$\sqrt{\mdima \mdimb}$.  In practice, we are interested in settings of
$\DALCON$ lying between these two extremes.

\hfill \goodendex \eexs

Past work on $\ell_1$-forms of matrix decomposition has imposed
singular vector incoherence conditions that are related to but
different from our spikiness condition. More concretely, if we write
the SVD of the low-rank component as $\TrueLowRank = U D V^T$ where
$D$ is diagonal, and $U \in \real^{\mdima \times \rdim}$ and $V \in
\real^{\mdimb \times \rdim}$ are matrices of the left and right
singular vectors.  Singular vector incoherence bounds quantities such
as
\begin{align}
\label{EqnIncohere}
\|U U^T - \frac{\rdim}{\mdima} I_{\mdima \times \mdima} \|_\infty,
\quad \|V V^T - \frac{\rdim}{\mdimb} I_{\mdimb \times \mdimb}
\|_\infty, \quad \mbox{and}  \quad \|U V^T\|_\infty.
\end{align}
all of which measure the degree of ``coherence'' between the singular
vectors and the canonical basis.  A remarkable feature of such
conditions is that they have no dependence on the \emph{singular
  values} of $\TrueLowRank$.  This lack of dependence makes sense in
the noiseless setting, where exact recovery is the goal.  For noisy
models, in contrast, one should only be concerned with recovering
components with ``large'' singular values.  In this context, our bound
on the maximum element $\|\TrueLowRank\|_\infty$, or equivalently on
the quantity $\|U D V^T\|_\infty$, is natural.  Note that it imposes
no constraint on the matrices $U U^T$ or $V V^T$, and moreover it uses
the diagonal matrix of singular values as a weight in the
$\ell_\infty$ bound.  Moreover, we note that there are many matrices
for which $\|\TrueLowRank\|_\infty$ satisfies a reasonable bound,
whereas the incoherence measures are poorly behaved (e.g., see Section
3.4.2 in the paper~\cite{NegWai10b} for one example).
\bexs[Column-sparsity and block columnwise regularization] 
\label{ExaColSparse}
Other applications involve models in which $\TrueSparse$ has a
relatively small number $\kdim \ll \mdimb$ of non-zero columns (or a
relatively small number $\kdim \ll \mdima$ of non-zero rows).  Such
applications include the multi-task regression problem from
Example~\ref{ExaMultitask}, the robust covariance problem from
Example~\ref{ExaRobustCovariance}, as well as a form of robust PCA
considered by Xu et al.~\cite{XuCaSa2010}.  In this case, it is
natural to constrain $\Sparse$ via the $(2,1)$-norm regularizer
\begin{align}
\label{EqnColReg}
\Reg{\Sparse} \; = \; \gennorm{\Sparse}{2,1} & \defn \; \sum_{k =
  1}^\mdimb \|\Sparse_k\|_2,
\end{align}
where $\Sparse_k$ is the $k^{th}$ column of $\Sparse$ (or the
$(1,2)$-norm regularizer that enforces the analogous constraint on the
rows of $\Sparse$).  For this choice, it can be verified that
\begin{align}
\Dualreg{U} & = \gennorm{U}{2, \infty} \; \defn \; \max_{k = 1, 2,
  \ldots, \mdimb} \|U_k\|_2,
\end{align}
where $U_k$ denotes the $k^{th}$ column of $U$, and that
$\DCON(\Dregplain) = \sqrt{\mdimb}$. Consequently, in this specific
case, the general convex program~\eqref{EqnGenProg} takes the form
\begin{align}
\label{EqnColProb}
\min_{(\LowRank, \Sparse)} \Big \{ \frac{1}{2} \frob{\Ymat - \Xop{\LowRank
  + \Sparse}}^2 \; + \regpar \, \nuclear{\LowRank} + \tworeg \,
\gennorm{\Sparse}{2,1} \Big \} \quad \mbox{such that
  $\gennorm{\LowRank}{2,\infty} \leq \frac{\DALCON}{\sqrt{\mdimb}}$.}
\end{align}
As before, the constraint $\gennorm{\LowRank}{2, \infty}$ serves to
limit the ``spikiness'' of the low rank component, where in this case,
spikiness is measured in a columnwise manner.  Again, it is natural to
consider matrices such that $\frob{\TrueLowRank} \approx 1$, so that
the signal-to-noise ratio in the observation
model~\eqref{EqnLinearObs} stays fixed.  Thus, if $\DALCON \approx 1$,
then we are restricted to matrices for which $\|\TrueLowRank_k\|_2
\approx \frac{1}{\sqrt{\mdimb}}$ for all columns $k = 1, 2, \ldots,
\mdimb$.  At the other extreme, in order to permit a maximally
``column-spiky'' matrix (i.e., with a single non-zero column of
$\ell_2$-norm roughly $1$), we need to set $\DALCON \approx
\sqrt{\mdimb}$.  As before, of practical interest are settings of
$\DALCON$ lying between these two extremes.
\hfill \goodendex \eexs
%


\section{Main results and their consequences}
\label{SecMain}

In this section, we state our main results, and discuss some of their
consequences.  Our first result applies to the family of convex
programs~\eqref{EqnGenProg} whenever $\Regplain$ belongs to the class
of decomposable regularizers, and the least-squares loss associated
with the observation model satisfies a specific form of restricted
strong convexity~\cite{NegRavWaiYu09}.  Accordingly, we begin in
Section~\ref{SecDecomposable} by defining the notion of
decomposability, and then illustrating how the elementwise-$\ell_1$
and columnwise-$(2,1)$-norms, as discussed in Examples~\ref{ExaEllOne}
and~\ref{ExaColSparse} respectively, are both instances of
decomposable regularizers.  In Section~\ref{SecRSC}, we define the
form of restricted strong convexity appropriate to our setting.
Section~\ref{SecGenResult} contains the statement of our main result
about the $M$-estimator~\eqref{EqnGenProg}, while
Sections~\ref{SecEllOne} and~\ref{SecColSparse} are devoted to its
consequences for the cases of elementwise sparsity and columnwise
sparsity, respectively.  In Section~\ref{SecTwoStep}, we complement
our analysis of the convex program~\eqref{EqnGenProg} by showing that,
in the special case of the identity operator, a simple two-step method
can achieve similar rates (up to constant factors).  We also provide
an example showing that the two-step method can fail for more general
observation operators.  In Section~\ref{SecMinimax}, we state matching
lower bounds on the minimax errors in the case of the identity
operator and Gaussian noise.

\subsection{Decomposable regularizers}
\label{SecDecomposable}

The notion of decomposability is defined in terms of a pair of
subspaces, which (in general) need not be orthogonal complements.
Here we consider a special case of decomposability that is sufficient
to cover the examples of interest in this paper:
\bdes 
Given a subspace $\Model \subseteq \real^{\mdima \times \mdimb}$
and its orthogonal complement $\ModelPerp$, a norm-based regularizer
$\Regplain$ is \emph{decomposable with respect $(\Model, \ModelPerp)$}
if
\begin{align}
\label{EqnDecomposable}
\Reg{U + V} & = \Reg{U} + \Reg{V} \qquad \mbox{for all $U \in \Model$,
  and $V \in \ModelPerp$.}
\end{align}
\edes
\noindent To provide some intuition, the subspace $\Model$ should be
thought of as the nominal \emph{model subspace}; in our results, it
will be chosen such that the matrix $\TrueSparse$ lies within or close
to $\Model$.  The orthogonal complement $\ModelPerp$ represents
deviations away from the model subspace, and the
equality~\eqref{EqnDecomposable} guarantees that such deviations are
penalized as much as possible.

As discussed at more length in Negahban et al.~\cite{NegRavWaiYu09}, a
large class of norms are decomposable with respect to
interesting\footnote{Note that any norm is (trivially) decomposable
  with respect to the pair $(\Model, \ModelPerp) = (\real^{\mdima
    \times \mdimb}, \{0\})$.}  subspace pairs.  Of particular
relevance to us is the decomposability of the elementwise
$\ell_1$-norm $\ellnorm{\Sparse}$ and the columnwise $(2,1)$-norm
$\gennorm{\Sparse}{2,1}$, as previously discussed in
Examples~\ref{ExaEllOne} and~\ref{ExaColSparse} respectively.

\paragraph{Decomposability of $\Reg{\cdot} = \|\cdot\|_1$:}
Beginning with the elementwise $\ell_1$-norm, given an arbitrary
subset $\Sset \subseteq \{1, 2, \ldots, \mdima \} \times \{1, 2,
\ldots, \mdimb \}$ of matrix indices, consider the subspace pair
\begin{align}
\label{EqnDecompSparse}
\Model(\Sset) & \defn \; \big \{ U \in \real^{\mdima \times \mdimb} \,
\mid \, U_{jk} = 0 \quad \mbox{for all $(j,k) \notin \Sset$} \big \},
\quad \mbox{and} \quad \ModelPerp(\Sset) \defn (\Model(\Sset))^\perp.
\end{align}
It is then easy to see that for any pair $U \in \Model(\Sset), U' \in
\ModelPerp(\Sset)$, we have the splitting \mbox{$\ellnorm{U + U'} =
  \ellnorm{U} + \ellnorm{U'}$,} showing that the elementwise
$\ell_1$-norm is decomposable with respect to the pair
$(\Model(\Sset), \ModelPerp(\Sset))$.

\paragraph{Decomposability of $\Reg{\cdot} = \|\cdot\|_{2,1}$:}
Similarly, the columnwise $(2,1)$-norm is also decomposable with
respect to appropriately defined subspaces, indexed by subsets
$\Colset \subseteq \{1, 2, \ldots, \mdimb \}$ of column indices.
Indeed, using $V_k$ to denote the $k^{th}$ column of the matrix $V$,
define
\begin{align}
\label{EqnColset}
\Model(\Colset) & \defn \; \big \{ V \in \real^{\mdima \times \mdimb}
\, \mid \, V_{k} = 0 \quad \mbox{for all $k \notin \Colset$} \big \},
\end{align}
and $\ModelPerp(\Colset) \defn (\Model(\Colset))^\perp$.  Again, it is
easy to verify that for any pair \mbox{$V \in \Model(\ColSet), V' \in
  \ModelPerp(\ColSet)$,} we have $\gennorm{V + V'}{2,1} =
\gennorm{V}{2,1} + \gennorm{V'}{2, 1}$, thus verifying the
decomposability property. \\

\vspace*{.2in}

For any decomposable regularizer and subspace $\Model \neq \{0\}$, we
define the compatibility constant
\begin{align}
\label{EqnSubspaceCompat}
\Compat(\Model, \Regplain) & \defn \sup_{U \in \Model, U \neq 0}
\frac{\Reg{U}}{\frob{U}}.
\end{align}
This quantity measures the compatibility between the Frobenius norm
and the regularizer over the subspace $\Model$.  For example, for the
$\ell_1$-norm and the set $\Model(\Sset)$ previously
defined~\eqref{EqnDecompSparse}, an elementary calculation yields
$\Compat \big(\Model(\Sset); \ellnorm{\cdot} \big) = \sqrt{\kdim}$.


\subsection{Restricted strong convexity}
\label{SecRSC}

Given a loss function, the general notion of strong convexity involves
establishing a quadratic lower bound on the error in the first-order
Taylor approximation~\cite{Boyd02}.  In our setting, the loss is the
quadratic function $\mathcal{L}(\Omega) = \frac{1}{2} \frob{Y -
  \Xop{\Omega}}^2$ (where we use $\Omega = \Theta + \Gamma$), so that
the first-order Taylor series error at $\Omega$ in the direction of
the matrix $\Delta$ is given by
\begin{align}
\Loss(\Omega + \Delta) - \Loss(\Omega) - \Loss(\Omega)^T \, \Delta & =
\frac{1}{2} \frob{\Xop{\Delta}}^2.
\end{align}
Consequently, strong convexity is equivalent to a lower bound of the
form $\frac{1}{2} \|\Xop{\Delta}\|_2^2 \geq \frac{\lossrsc}{2}
\frob{\Delta}^2$, where $\lossrsc > 0$ is the strong convexity
constant. \\

Restricted strong convexity is a weaker condition that also involves a
norm defined by the regularizers.  In our case, for any pair
$(\tworeg, \regpar)$ of positive numbers, we first define the weighted
combination of the two regularizers---namely
\begin{align}
\label{EqnDefnBigReg}
\BIGREG(\LowRank, \Sparse) & \defn \nuclear{\LowRank} +
\frac{\tworeg}{\regpar} \Reg{\Sparse}.
\end{align}
For a given matrix $\Delta$, we can use this weighted combination to
define an associated norm
\begin{align}
\label{EqnDefnRegInfConv}
\RegInfConv{\Delta} & \defn \inf_{\LowRank + \Sparse = \Delta}
\BIGREG(\LowRank, \Sparse),
\end{align}
corresponding to the minimum value of $\BIGREG(\LowRank, \Sparse)$
over all decompositions of $\Delta$\footnote{Defined this way,
  $\RegInfConv{\Delta}$ is the infimal-convolution of the two norms
  $\nuclear{\cdot}$ and $\Regplain$, which is a very well studied
  object in convex analysis (see e.g.~\cite{Rockafellar})}.
\bdes[RSC]
The quadratic loss with linear operator $\XopPlain : \real^{\mdima
  \times \mdimb} \to \real^{\numobs_1 \times \numobs_2}$ satisfies
restricted strong convexity with respect to the norm
$\RegInfConvPlain$ and with parameters $(\lossrsc, \FUNLOW)$ if
\begin{align}
  \label{EqnDefnRSC}
  \frac{1}{2} \frob{\Xop{\Delta}}^2 & \geq \frac{\lossrsc}{2}
  \frob{\Delta}^2 - \FUNLOW \, \RegInfConvSq{\Delta} \qquad \mbox{for
    all $\Delta \in \real^{\mdima \times \mdimb}$}.
\end{align}
\edes
\noindent Note that if condition~\eqref{EqnDefnRSC} holds with
$\FUNLOW = 0$ and any $\lossrsc > 0$, then we recover the usual
definition of strong convexity (with respect to the Frobenius norm).
In the special case of the identity operator (i.e., $\Xop{\Theta}=
\Theta$), such strong convexity does hold with $\lossrsc = 1$.  More
general observation operators require different choices of the
parameter $\lossrsc$, and also non-zero choices of the tolerance
parameter $\FUNLOW$.

While RSC establishes a form of (approximate) identifiability in
general, here the error $\Delta$ is a combination of the error in
estimating $\TrueLowRank$ ($\DelL$) and $\TrueSparse$
($\DelS$). Consequently, we will need a further lower bound on
$\frob{\Delta}$ in terms of $\frob{\DelL}$ and $\frob{\DelS}$ in the
proof of our main results to demonstrate the (approximate)
identifiability of our model under the RSC condition~\ref{EqnDefnRSC}.

\subsection{Results for general regularizers and noise}
\label{SecGenResult}
We begin by stating a result for a general observation operator
$\XopPlain$, a general decomposable regularizer $\Regplain$ and a
general noise matrix $\Wmat$.  In later subsections, we specialize
this result to particular choices of observation operator,
regularizers, and stochastic noise matrices.  In all our results, we
measure error using the \emph{squared Frobenius norm} summed across
both matrices
\begin{align}
\FERRSQ{\LowRankHat}{\SparseHat} & \defn \frob{\LowRankHat -
  \TrueLowRank}^2 + \frob{\SparseHat - \TrueSparse}^2.
\end{align}
With this notation, the following result applies to the observation
model $Y = \Xop{\TrueSparse + \TrueLowRank} + W$, where the low-rank
matrix satisfies the constraint \mbox{$\genspike(\TrueLowRank) \leq
  \DALCON$.}    Our upper bound on the squared Frobenius error consists
of three terms
\begin{subequations}
\begin{align}
\HACK_\TrueLowRank & \defn \frac{\regpar^2}{\lossrsc^2} \biggr \{\rdim
+ \frac{\lossrsc}{\regpar} \sum_{j = \rdim+1}^\mdim
\sigma_j(\TrueLowRank) \biggr \} \\
\HACK_\TrueSparse & \defn \frac{\twopar^2}{\lossrsc^2} \biggr \{
\Compat^2(\Model; \Regplain) + \frac{\lossrsc}{\twopar}
\Reg{\ProjPerp{\TrueSparse}} \biggr \} \\
\HACK_\FUNLOW & \defn \frac{\FUNLOWL}{\lossrsc} \biggr \{ \EXCESS \biggr
\}^2.
\end{align}
\end{subequations}
As will be clarified shortly, these three terms correspond to the
errors associated with the low-rank term ($\HACK_\TrueLowRank$), the
sparse term ($\HACK_\TrueSparse$), and additional error
($\HACK_\FUNLOW$) associated with a non-zero tolerance $\FUNLOW \neq
0$ in the RSC condition~\eqref{EqnDefnRSC}.

 \btheos
\label{ThmConvergence}
Suppose that the observation operator $\XopPlain$ satisfies the RSC
condition~\eqref{EqnDefnRSC} with curvature $\lossrsc > 0$, and a
tolerance $\FUNLOW$ such that there exist integers \mbox{$\rdim = 1,
  2, \ldots, \min \{\mdima, \mdimb \}$,} for which
\begin{equation}
  \label{EqnTauChoice}
  128 \, \FUNLOWL \, \rdim < \frac{\lossrsc}{4}, \quad \mbox{and}
  \quad 64 \, \FUNLOWL \, \biggr( \Compat(\Model;\Regplain) \,
  \frac{\tworeg}{\regpar} \biggr)^2 < \frac{\lossrsc}{4}.
\end{equation}
Then if we solve the convex program~\eqref{EqnGenProg} with
regularization parameters $(\regpar, \tworeg)$ satisfying
\begin{equation}
  \label{EqnRegChoice}
  \regpar \geq 4 \opnorm{\XopDual{\Wmat}}, \quad \mbox{and} \quad
  \tworeg \geq 4 \, \Dualreg{\XopDual{\Wmat}} + \frac{4 \, \lossrsc \;
    \DALCON}{\DCON},
\end{equation}
there are universal constant $\plaincon_j, j = 1, 2, 3$ such that for
any matrix pair $(\TrueLowRank, \TrueSparse)$ satisfying
$\genspike(\TrueLowRank) \leq \DALCON$ and any
$\Regplain$-decomposable pair $(\Model, \ModelPerp)$, any optimal
solution $(\LowRankHat, \SparseHat)$ satisfies
\begin{align}
\label{EqnGenRate}
\FERRSQ{\LowRankHat}{\SparseHat} & \leq \plaincon_1 \HACK_\TrueLowRank
+ \plaincon_2 \HACK_\TrueSparse + \plaincon_3 \HACK_{\FUNLOW}.
\end{align}
\etheos

\noindent Let us make a few remarks in order to interpret the meaning
of this claim.
\paragraph{Deterministic guarantee:}
To be clear, Theorem~\ref{ThmConvergence} is a deterministic statement
that applies to any optimum of the convex program~\eqref{EqnGenProg}.
Moreover, it actually provides a whole family of upper bounds, one for
each choice of the rank parameter $\rdim$ and each choice of the
subspace pair $(\Model, \ModelPerp)$.  In practice, these choices are
optimized so as to obtain the tightest possible upper bound.  As for
the condition~\eqref{EqnTauChoice}, it will be satisfied for a
sufficiently large sample size $\numobs$ as long as $\lossrsc > 0$,
and the tolerance $\FUNLOW$ decreases to zero with the sample size.
In many cases of interest---including the identity observation
operator and multi-task cases---the RSC condition holds with $\FUNLOW
= 0$, so that condition~\eqref{EqnTauChoice} holds as long as
$\lossrsc > 0$.

\paragraph{Interpretation of different terms:}
Let us focus first on the term $\HACK_\TrueLowRank$, which corresponds
to the complexity of estimating the low-rank component.  It is further
sub-divided into two terms, with the term $\regpar^2 \: \rdim$
corresponding to the \emph{estimation error} associated with a rank
$\rdim$ matrix, whereas the term $ \regpar \: \sum_{j = \rdim+1}^\mdim
\sigma_j(\TrueLowRank)$ corresponds to the \emph{approximation error}
associated with representing $\TrueLowRank$ (which might be full rank)
by a matrix of rank $\rdim$.  A similar interpretation applies to the
two components associated with $\TrueSparse$, the first of which
corresponds to a form of estimation error, whereas the second
corresponds to a form of approximation error. \\

\paragraph{A family of upper bounds:}
Since the inequality~\eqref{EqnGenRate} corresponds to a family of
upper bounds indexed by $\rdim$ and the subspace $\Model$, these
quantities can be chosen adaptively, depending on the structure of the
matrices $(\TrueLowRank, \TrueSparse)$, so as to obtain the tightest
possible upper bound.  In the simplest case, the RSC conditions hold
with tolerance $\FUNLOW = 0$, the matrix $\TrueLowRank$ is exactly low
rank (say rank $\rdim$), and $\TrueSparse$ lies within a
$\Regplain$-decomposable subspace $\Model$. In this case, the
approximation errors vanish, and Theorem~\ref{ThmConvergence}
guarantees that the squared Frobenius error is at most
\begin{align}
\label{EqnEasy}
e^2(\LowRankHat; \SparseHat) & \; \precsim \regpar^2 \rdim + \tworeg^2
\Compat^2(\Model; \Regplain),
\end{align}
where the $\precsim$ notation indicates that we ignore constant
factors.

\subsection{Results for $\ell_1$-norm regularization}
\label{SecEllOne}

Theorem~\ref{ThmConvergence} holds for any regularizer that is
decomposable with respect to some subspace pair.  As previously noted,
an important example of a decomposable regularizer is the elementwise
$\ell_1$-norm, which is decomposable with respect to subspaces of the
form~\eqref{EqnDecompSparse}. 
\bcors
\label{CorSparseDet}
Consider an observation operator $\XopPlain$ that satisfies the RSC
condition~\eqref{EqnDefnRSC} with $\lossrsc > 0$ and $\FUNLOW = 0$.
Suppose that we solve the convex program~\eqref{EqnSparseProb} with
regularization parameters $(\regpar, \tworeg)$ such that
\begin{equation}
\label{EqnRegChoiceSparse}
\regpar \geq 4 \opnorm{\XopDual{\Wmat}}, \quad \mbox{and} \quad
\tworeg \geq 4 \, \infnorm{\XopDual{\Wmat}} + \frac{ 4 \lossrsc \,
  \DALCON}{\sqrt{\mdima \mdimb}}.
\end{equation}
Then there are universal constants $\plaincon_j$ such that for any
matrix pair $(\TrueLowRank, \TrueSparse)$ with
\mbox{$\infnorm{\TrueLowRank} \leq \frac{\DALCON}{\sqrt{\mdima
      \mdimb}}$} and for all integers \mbox{$\rdim = 1, 2, \ldots,
  \min \{\mdima, \mdimb \}$}, and $\kdim = 1, 2, \ldots, (\mdima
\mdimb)$, we have
\begin{align}
\label{EqnSparseDet}
\FERRSQ{\LowRankHat}{\SparseHat} & \leq \plaincon_1
\frac{\regpar^2}{\lossrsc^2} \, \biggr \{ \rdim + \frac{1}{\regpar} \,
\sum_{j = \rdim+1}^\mdim \sigma_j(\TrueLowRank) \biggr \} +
\plaincon_2 \, \frac{\twopar^2}{\lossrsc^2} \, \biggr \{ \kdim +
\frac{1}{\twopar} \sum_{(j,k) \notin \Sset} |\TrueSparse_{jk}| \biggr
\},
\end{align}
where $\Sset$ is an arbitrary subset of matrix indices of cardinality
at most $\kdim$.
\ecors

\paragraph{Remarks:}    This result follows directly by specializing
Theorem~\ref{ThmConvergence} to the elementwise \mbox{$\ell_1$-norm.}
As noted in Example~\ref{ExaEllOne}, for this norm, we have $\DCON =
\sqrt{\mdima \mdimb}$, so that the choice~\eqref{EqnRegChoiceSparse}
satisfies the conditions of Theorem~\ref{ThmConvergence}.  The dual
norm is given by the elementwise $\ell_\infty$-norm $\Dualreg{\cdot} =
\infnorm{\cdot}$.  As observed in Section~\ref{SecDecomposable}, the
$\ell_1$-norm is decomposable with respect to subspace pairs of the
form $(\Model(\Sset), \ModelPerp(\Sset))$, for an arbitrary subset
$\Sset$ of matrix indices.  Moreover, for any subset $\Sset$ of
cardinality $\kdim$, we have $\Compat^2(\Model(\Sset)) = \kdim$.  It
is easy to verify that with this choice, we have
$\ProjPerp{\TrueSparse} = \sum \limits_{(j,k) \notin \Sset}
|\TrueSparse_{jk}|$, from which the claim follows. \\

It is worth noting the inequality~\eqref{EqnGenRate} corresponds to a
family of upper bounds indexed by $\rdim$ and the subset $\Sset$.  For
any fixed integer $\kdim \in \{1, 2, \ldots, (\mdima \mdimb) \}$, it
is natural to let $\Sset$ index the largest $\kdim$ values (in
absolute value) of $\TrueSparse$.  Moreover, the choice of the pair
$(\rdim, \kdim)$ can be further adapted to the structure of the
matrix.  For instance, when $\TrueLowRank$ is exactly low rank, and
$\TrueSparse$ is exactly sparse, then one natural choice is $\rdim =
\rank(\TrueLowRank)$, and $\kdim = |\supp(\TrueSparse)|$.  With this
choice, both the approximation terms vanish, and
Corollary~\ref{CorSparseDet} guarantees that any solution
$(\LowRankHat, \SparseHat)$ of the convex
program~\eqref{EqnSparseProb} satisfies
\begin{align}
\label{EqnExactCase}
\frob{\LowRankHat - \TrueLowRank}^2 + \frob{\SparseHat -
  \TrueSparse}^2 & \; \lesssim \; \regpar^2 \; \rdim + \twopar^2 \;
\kdim.
\end{align}
Further specializing to the case of noiseless observations ($\Wmat =
0$), yields a form of approximate recovery---namely
\begin{align}
\label{EqnSparseNoiseless}
\frob{\LowRankHat - \TrueLowRank}^2 + \frob{\SparseHat -
  \TrueSparse}^2 & \; \lesssim \; \DALCON^2 \frac{\kdim}{\mdima \mdimb}.
\end{align}
This guarantee is weaker than the exact recovery results obtained in
past work on the noiseless observation model with identity
operator~\cite{Chand09,CandesLiMaWr2009}; however, these papers
imposed incoherence requirements on the singular vectors of the
low-rank component $\TrueLowRank$ that are more restrictive than the
conditions of Theorem~\ref{ThmConvergence}.

\vspace*{.1in}

Our elementwise $\ell_\infty$ bound is a weaker condition than
incoherence, since it allows for singular vectors to be coherent as
long as the associated singular value is not too large.  Moreover, the
bound~\eqref{EqnSparseNoiseless} is unimprovable up to constant
factors, due to the non-identifiability of the observation
model~\eqref{EqnLinearObs}, as shown by the following example for the
identity observation operator $\XopPlain = I$.
\bexs
\label{ExaMild}[Unimprovability for elementwise sparse model]
Consider a given sparsity index $\kdim \in \{1, 2, \ldots, (\mdima
\mdimb) \}$, where we may assume without loss of generality that
$\kdim \leq \mdimb$.  We then form the matrix
\begin{align}
\label{EqnBad}
\TrueLowRank & \defn \frac{\DALCON}{\sqrt{\mdima \mdimb}}
\begin{bmatrix} 1 \\ 0 \\ \vdots  \\ 0
\end{bmatrix}
\underbrace{\begin{bmatrix} 1 & 1 & 1 & \ldots & 0 & \ldots & 0
\end{bmatrix}}_{f^T},
\end{align}
where the vector $f \in \real^{\mdimb}$ has exactly $\kdim$ ones.
Note that $\infnorm{\TrueLowRank} = \frac{\DALCON}{\sqrt{\mdima
    \mdimb}}$ by construction, and moreover $\TrueLowRank$ is rank
one, and has $\kdim$ non-zero entries.  Since up to $\kdim$ entries of
the noise matrix $\TrueSparse$ can be chosen arbitrarily, ``nature''
can always set $\TrueSparse = - \TrueLowRank$, meaning that we would
observe $\Ymat = \TrueLowRank + \TrueSparse = 0$.  Consequently, based
on observing only $\Ymat$, the pair $(\TrueLowRank, \TrueSparse)$ is
indistinguishable from the all-zero matrices $(0_{\mdima \times
  \mdimb}, 0_{\mdima \times \mdimb})$.  This fact can be used to show
that no method can have squared Frobenius error lower than $\approx
\frac{\DALCON^2 \, \kdim}{\mdima \mdimb}$; see
Section~\ref{SecMinimax} for a precise statement.  Therefore, the
bound~\eqref{EqnSparseNoiseless} cannot be improved unless one is
willing to impose further restrictions on the pair $(\TrueLowRank,
\TrueSparse)$.  We note that the singular vector incoherence
conditions, as imposed in past
work~\cite{Chand09,CandesLiMaWr2009,HUPAPER} and used to guarantee
exact recovery, would exclude the matrix~\eqref{EqnBad}, since its
left singular vector is the unit vector $e_1 \in \real^{\mdima}$.
\hfill \goodendex
\eexs

\subsubsection{Results for stochastic noise matrices}

Our discussion thus far has applied to general observation operators
$\XopPlain$, and general noise matrices $\Wmat$.  More concrete
results can be obtained by assuming particular forms of $\XopPlain$,
and that the noise matrix $\Wmat$ is stochastic.  Our first stochastic
result applies to the identity operator $\XopPlain = I$ and a noise
matrix $\Wmat$ generated with i.i.d. $N(0, \noisevar^2/(\mdima
\mdimb))$ entries.\footnote{To be clear, we state our results in terms
  of the noise scaling $\noisevar^2/(\mdima \mdimb)$ since it
  corresponds to a model with constant signal-to-noise ratio when the
  Frobenius norms of $\TrueLowRank$ and $\TrueSparse$ remain bounded,
  independently of the dimension.  The same results would hold if the
  noise were not rescaled, modulo the appropriate rescalings of the
  various terms.}

\bcors
\label{CorSparseNoisy}
Suppose $\XopPlain = I$, the matrix $\TrueLowRank$ has rank at most
$\rdim$ and satisfies \mbox{$\infnorm{\TrueLowRank} \leq
  \frac{\DALCON}{\sqrt{\mdima \mdimb}}$}, and $\TrueSparse$ has at
most $\kdim$ non-zero entries.  If the noise matrix $\Wmat$ has
i.i.d. $N(0, \noisevar^2/(\mdima \mdimb))$ entries, and we solve the
convex program~\eqref{EqnSparseProb} with regularization parameters
\begin{equation}
\label{EqnRegChoiceSparseNoisy}
\regpar = \frac{8 \noisevar}{\sqrt{\mdima}} + \frac{8
  \noisevar}{\sqrt{\mdimb}}, \quad \mbox{and} \quad \twopar = 16
\noisevar \, \sqrt{ \frac{\log (\mdima \mdimb)}{\mdima \mdimb}} +
\frac{4 \DALCON}{\sqrt{\mdima \mdimb}},
\end{equation}
then with probability greater than \mbox{$1 - \exp \big(-2 \log
  (\mdima \mdimb) \big)$,} any optimal solution $(\LowRankHat,
\SparseHat)$ satisfies
\begin{equation}
\label{EqnSparseNoisy}
e^2(\LowRankHat, \SparseHat) \leq \underbrace{\plaincon_1 \noisevar^2
  \, \biggr( \frac{\rdim \, (\mdima + \mdimb)}{\mdima \mdimb} \biggr
  )}_{\HACK_\TrueLowRank} + \underbrace{\plaincon_1 \noisevar^2 \,
  \biggr ( \frac{\kdim \, \log( \mdima \mdimb)}{\mdima \mdimb} \biggr)
  + \plaincon_1 \frac{\DALCON^2 \, \kdim}{\mdima
    \mdimb}}_{\HACK_\TrueSparse}
\end{equation}
\ecors

\paragraph{Remarks:}  In the statement of this corollary, 
the settings of $\regpar$ and $\twopar$ are based on upper bounding
$\infnorm{\Wmat}$ and $\opnorm{\Wmat}$, using large deviation bounds
and some non-asymptotic random matrix theory.  With a slightly
modified argument, the bound~\eqref{EqnSparseNoisy} can be sharpened
slightly by reducing the logarithmic term to $\log (\frac{\mdima
  \mdimb}{\kdim})$.  As shown in Theorem~\ref{ThmMinimax} to follow in
Section~\ref{SecMinimax}, this sharpened bound is minimax-optimal,
meaning that no estimator (regardless of its computational complexity)
can achieve much better estimates for the matrix classes and noise
model given here.  

It is also worth observing that both terms in the
bound~\eqref{EqnSparseNoisy} have intuitive interpretations.
Considering first the term $\HACK_\TrueLowRank$, we note that the
numerator term $\rdim (\mdima + \mdimb)$ is of the order of the number
of free parameters in a rank $\rdim$ matrix of dimensions $\mdima
\times \mdimb$.  The multiplicative factor $\frac{\noisevar^2}{\mdima
  \mdimb}$ corresponds to the noise variance in the problem.  On the
other hand, the term $\HACK_\TrueSparse$ measures the complexity of
estimating $\kdim$ non-zero entries in a $\mdima \times \mdimb$
matrix.  Note that there are ${\mdima \mdimb \choose \kdim}$ possible
subsets of size $\kdim$, and consequently, the numerator includes a
term that scales as $\log {\mdima \mdimb \choose \kdim} \approx \kdim
\log(\mdima \mdimb)$.  As before, the multiplicative pre-factor
$\frac{\noisevar^2}{\mdima \mdimb}$ corresponds to the noise variance.
Finally, the second term within $\HACK_\TrueSparse$---namely the
quantity $\frac{\DALCON^2 \, \kdim}{\mdima \mdimb}$---arises from the
non-identifiability of the model, and as discussed in
Example~\ref{ExaMild}, it cannot be avoided without imposing further
restrictions on the pair $(\TrueSparse, \TrueLowRank)$. \\

We now turn to analysis of the sparse factor analysis problem: as
previously introduced in Example~\ref{ExaFactor}, this involves
estimation of a covariance matrix that has a low-rank plus elementwise
sparse decomposition.  In this case, given $\numobs$ i.i.d. samples
from the unknown covariance matrix $\Sigma = \TrueLowRank +
\TrueSparse$, the noise matrix $\Wmat \in \real^{\mdim \times \mdim}$
is a recentered Wishart noise (see equation~\eqref{EqnWishNoise}).  We
can use tail bounds for its entries and its operator norm in order to
specify appropriate choices of the regularization parameters $\regpar$
and $\tworeg$.  We summarize our conclusions in the following
corollary:
\bcors
\label{CorFactor}
Consider the factor analysis model with $\numobs \geq \mdim$ samples,
and regularization parameters
\begin{equation}
\label{EqnRegparFactor}
\regpar \; = \; 16 \matsnorm{\sqrt{\Sigma}}{2}
\sqrt{\frac{\mdim}{\numobs}}, \quad \mbox{and} \quad \twopar \; = \;
32 \rho(\Sigma) \; \sqrt{\frac{\log \mdim}{\numobs}} + \frac{4
  \DALCON}{\mdim}, \quad \mbox{where $\rho(\Sigma) = \max_{j}
  \Sigma_{jj}$.}
\end{equation}
Then with probability greater than \mbox{$1 - c_2\exp \big(-c_3 \log
  (\mdim) \big)$,} any optimal solution $(\LowRankHat, \SparseHat)$
satisfies
\begin{align*}
e^2(\LowRankHat, \SparseHat) \leq \plaincon_1 \big \{
\matsnorm{\Sigma}{2} \frac{\rdim \mdim}{\numobs} + \rho(\Sigma)
\frac{\kdim \log \mdim}{\numobs} \big \} + \plaincon_1 \frac{\DALCON^2
  \kdim}{\mdim^2}.
\end{align*}
\ecors
\noindent We note that the condition $\numobs \geq \mdim$ is necessary
to obtain consistent estimates in factor analysis models, even in the
case with $\TrueSparse = I_{\mdim \times \mdim}$ where PCA is possible
(e.g., see Johnstone~\cite{John01}).  Again, the terms in the bound
have a natural interpretation: since a matrix of rank $\rdim$ in
$\mdim$ dimensions has roughly $\rdim \mdim$ degrees of freedom, we
expect to see a term of the order $\frac{\rdim \mdim}{\numobs}$.
Similarly, since there are $\log {\mdim^2 \choose \kdim} \approx \kdim
\log \mdim$ subsets of size $\kdim$ in a $\mdim \times \mdim$ matrix,
we also expect to see a term of the order $\frac{\kdim \log
  \mdim}{\numobs}$.  Moreover, although we have stated our choices of
regularization parameter in terms of $\matsnorm{\Sigma}{2}$ and
$\rho(\Sigma)$, these can be replaced by the analogous versions using
the sample covariance matrix $\widehat{\Sigma}$.  (By the
concentration results that we establish, the population and empirical
versions do not differ significantly when $\numobs \geq \mdim$.)

\subsubsection{Comparison to Hsu et al.~\cite{HUPAPER}} 

This recent work focuses on the problem of matrix decomposition with
the $\|\cdot\|_1$-norm, and provides results both for the noiseless
and noisy setting.  All of their work focuses on the case of exactly
low rank and exactly sparse matrices, and deals only with the identity
observation operator; in contrast, Theorem~\ref{ThmConvergence} in
this paper provides an upper bound for general matrix pairs and
observation operators.  Most relevant is comparison of our
$\ell_1$-results with exact rank-sparsity constraints to their Theorem
3, which provides various error bounds (in nuclear and Frobenius norm)
for such models with additive noise.  These bounds are obtained using
an estimator similar to our program~\eqref{EqnSparseProb}, and in
parts of their analysis, they enforce bounds on the $\ell_\infty$-norm
of the solution. However, this is not done directly with a constraint
on $\LowRank$ as in our estimator, but rather by penalizing the
difference $\|Y - \Sparse\|_\infty$, or by thresholding the solution.

Apart from these minor differences, there are two major differences
between our results, and those of Hsu et al.  First of all, their
analysis involves three quantities ($\alpha$, $\beta$, $\gamma$) that
measure singular vector incoherence, and must satisfy a number of
inequalities.  In contrast, our analysis is based only on a single
condition: the ``spikiness'' condition on the low-rank component
$\TrueLowRank$.  As we have seen, this constraint is weaker than
singular vector incoherence, and consequently, unlike the result of
Hsu et al., we do not provide exact recovery guarantees for the
noiseless setting.  However, it is interesting to see (as shown by our
analysis) that a very simple spikiness condition suffices for the
approximate recovery guarantees that are of interest for noisy
observation models.  Given these differing assumptions, the underlying
proof techniques are quite distinct, with our methods leveraging the
notion of restricted strong convexity introduced by Negahban et
al.~\cite{NegRavWaiYu09}.

The second (and perhaps most significant) difference is in the
sharpness of the results for the noisy setting, and the permissible
scalings of the rank-sparsity pair $(\rdim, \kdim)$. As will be
clarified in Section~\ref{SecMinimax}, the rates that we establish for
low-rank plus elementwise sparsity for the noisy Gaussian model
(Corollary~\ref{CorSparseNoisy}) are minimax-optimal up to constant
factors.  In contrast, the upper bounds in Theorem 3 of Hsu et al.
involve the \emph{product} $\rdim \kdim$, and hence are sub-optimal as
the rank and sparsity scale.  These terms appear only additively both
our upper and minimax lower bounds, showing that an upper bound
involving the product $\rdim \kdim$ is sub-optimal.  Moreover, the
bounds of Hsu et al. (see Section IV.D) are limited to matrix
decompositions for which the rank-sparsity pair $(\rdim, \kdim)$ are
bounded as
\begin{align}
\label{EqnHsuBound}
\rdim \kdim & \precsim \frac{\mdima \mdima}{\log(\mdima) \,
  \log(\mdimb)}
\end{align}
This bound precludes many scalings that are of interest.  For
instance, if the sparse component $\TrueSparse$ has a nearly constant
fraction of non-zeros (say $\kdim \asymp \frac{\mdima
  \mdimb}{\log(\mdima) \, \log(\mdimb)}$ for concreteness), then the
bound~\eqref{EqnHsuBound} restricts to $\TrueLowRank$ to have constant
rank.  In contrast, our analysis allows for high-dimensional scaling
of both the rank $\rdim$ and sparsity $\kdim$ simultaneously; as can
be seen by inspection of Corollary~\ref{CorSparseNoisy}, our Frobenius
norm error goes to zero under the scalings \mbox{$\kdim \asymp
  \frac{\mdima \mdimb}{\log(\mdima) \, \log(\mdimb)}$} and
\mbox{$\rdim \asymp \frac{\mdimb}{\log (\mdimb)}$}.

\subsubsection{Results for multi-task regression}

Let us now extend our results to the setting of multi-task regression,
as introduced in Example~\ref{ExaMultitask}.  The observation model is
of the form $Y = \EX \BEESTAR + W$, where $\EX \in \real^{\numobs
  \times \mdima}$ is a known design matrix, and we observe the matrix
$\Ymat = \real^{\numobs \times \mdimb}$.  Our goal is to estimate the
the regression matrix $\BEESTAR \in \real^{\mdima \times \mdimb}$,
which is assumed to have a decomposition of the form $\BEESTAR =
\TrueLowRank + \TrueSparse$, where $\TrueLowRank$ models the shared
characteristics between each of the tasks, and the matrix
$\TrueSparse$ models perturbations away from the shared structure.  If
we take $\TrueSparse$ to be a sparse matrix, an appropriate choice of
regularizer $\Regplain$ is the elementwise $\ell_1$-norm, as in
Corollary~\ref{CorSparseNoisy}. We use $\mineig$ and $\maxeig$ to
denote the minimum and maximum singular values (respectively) of the
rescaled design matrix $\EX/\sqrt{\numobs}$; we assume that $\EX$ is
invertible so that $\mineig > 0$, and moreover, that its columns are
uniformly bounded in $\ell_2$-norm, meaning that $\max_{j = 1, \ldots,
  \mdima} \|X_j\|_2 \leq \maxcol\sqrt{\numobs}$. We note that these
assumptions are satisfied for many common examples of random design.

\bcors
\label{CorMultiTask}
Suppose that the matrix $\TrueLowRank$ has rank at most $\rdim$ and
satisfies \mbox{$\infnorm{\TrueLowRank} \leq
  \frac{\DALCON}{\sqrt{\mdima \mdimb}}$}, and the matrix $\TrueSparse$
has at most $\kdim$ non-zero entries.  If the entries of $\Wmat$ are
i.i.d. $N(0,\noisevar^2)$, and we solve the convex
program~\eqref{EqnSparseProb} with regularization parameters
\begin{equation}
  \label{EqnRegChoiceMultiTask}
  \regpar = 8 \noisevar \, \maxeig \sqrt{\numobs}(\sqrt{\mdima} +
  \sqrt{\mdimb}), \quad \mbox{and} \quad \twopar = 16 \noisevar \,
  \maxcol \, \sqrt{ \numobs\log (\mdima \mdimb)} + \frac{4 \DALCON \,
    \mineig\sqrt{\numobs}}{\sqrt{\mdima \mdimb}},
\end{equation}
then with probability greater than \mbox{$1 - \exp \big(-2 \log
  (\mdima \mdimb) \big)$,} any optimal solution $(\LowRankHat,
\SparseHat)$ satisfies
\begin{equation}
\label{EqnSparseMultiTask}
e^2(\LowRankHat, \SparseHat) \; \leq \; \plaincon_1
\underbrace{\frac{\noisevar^2 \maxeig^2}{\mineig^4} \, \biggr(
  \frac{\rdim \, (\mdima + \mdimb)}{\numobs} \biggr
  )}_{\HACK_\TrueLowRank} \, + \, \plaincon_2 \, \underbrace{\big [
    \frac{\noisevar^2 \, \maxcol^2}{\mineig^4} \, \biggr ( \frac{\kdim
      \, \log( \mdima \mdimb)}{\numobs} \biggr) + \frac{\DALCON^2 \,
      \kdim}{\mdima \mdimb} \big ]}_{\HACK_\TrueSparse}.
\end{equation}
\ecors

\paragraph{Remarks:} We see that the results presented above are
analogous to those presented in
Corollary~\ref{CorSparseNoisy}. However, in this setting, we leverage
large deviations results in order to find bounds on
$\infnorm{\XopDual{\Wmat}}$ and $\opnorm{\XopDual{\Wmat}}$ that hold
with high probability given our observation model.


\subsection{An alternative two-step method}
\label{SecTwoStep}

As suggested by one reviewer, it is possible that a simpler two-step
method---namely, based on first thresholding the entries of the
observation matrix $Y$, and then performing a low-rank
approximation---might achieve similar rates to the more complex convex
relaxation~\eqref{EqnSparseProb}.  In this section, we provide a
detailed analysis of one version of such a procedure in the case of
nuclear norm combined with $\ell_1$-regularization.  We prove that in
the special case of $\XopPlain = I$, this procedure can attain the
same form of error bounds, with possibly different constants.
However, there is also a cautionary message here: we also give an
example to show that the two-step method will not necessarily perform
well for general observation operators $\XopPlain$.\\

\noindent In detail, let us consider the following two-step estimator:
\begin{enumerate}
\item[(a)] Estimate the sparse component $\TrueSparse$ by solving
\begin{equation}
\label{EqnTwoStepSparse}
\SparseHat \; \in \argmin_{\Sparse \in \real^{\mdima \times \mdimb}}
\big \{ \frac{1}{2} \matsnorm{\Ymat - \Sparse}{F}^2 \; + \tworeg \,
\|\Sparse\|_1 \big \}.
\end{equation}
As is well-known, this convex program has an explicit solution based
on soft-thresholding the entries of $\Ymat$.
\item[(b)] Given the estimate $\SparseHat$, estimate the low-rank
  component $\TrueLowRank$ by solving the convex program
\begin{equation}
  \label{EqnTwoStepLowRank}
  \LowRankHat \; \in \argmin_{\LowRank \in \real^{\mdima \times
      \mdimb}} \big \{ \frac{1}{2} \matsnorm{\Ymat - \LowRank -
    \SparseHat}{F}^2 \; + \regpar \, \nuclear{\LowRank} \big \}.
\end{equation}
\end{enumerate}
Interestingly, note that this method can be understood as the first
two steps of a blockwise co-ordinate descent method for solving the
convex program~\eqref{EqnSparseProb}.  In step (a), we fix the
low-rank component, and minimize as a function of the sparse
component.  In step (b), we fix the sparse component, and then
minimize as a function of the low-rank component.  The following
result that these two steps of co-ordinate descent achieve the same
rates (up to constant factors) as solving the full convex
program~\eqref{EqnSparseProb}:

\bprops
\label{PropTwoStep}
Given observations $\Ymat$ from the model $\Ymat = {\TrueLowRank +
  \TrueSparse} + W$ with \mbox{$\infnorm{\TrueLowRank} \leq
  \frac{\DALCON}{\sqrt{\mdima \mdimb}}$}, consider the two-step
procedure~\eqref{EqnTwoStepSparse} and~\eqref{EqnTwoStepLowRank} with
regularization parameters $(\regpar, \tworeg)$ such that
\begin{equation}
  \label{EqnRegChoiceTwoStep}
  \regpar \geq 4 \opnorm{\Wmat}, \quad \mbox{and} \quad \tworeg \geq 4
  \, \|\Wmat\|_\infty + \frac{4 \; \DALCON}{\sqrt{\mdima \mdimb}}.
\end{equation}
Then the error bound~\eqref{EqnSparseDet} from
Corollary~\ref{CorSparseDet} holds with $\lossrsc=1$.
\eprops
\noindent Consequently, in the special case that $\XopPlain = I$, then
there is no need to solve the convex program~\eqref{EqnSparseProb} to
optimality; rather, two steps of co-ordinate descent are
sufficient. \\

On the other hand, the simple two-stage method will not work for
general observation operators $\XopPlain$.  As shown in the proof of
Proposition~\ref{PropTwoStep}, the two-step method relies critically
on having the quantity $\|\Xop{\TrueLowRank + W}\|_\infty$ be upper
bounded (up to constant factors) by $\max \{ \|\TrueLowRank\|_\infty,
\|W\|_\infty\}$.  By triangle inequality, this condition holds
trivially when $\XopPlain = I$, but can be violated by other choices
of the observation operator, as illustrated by the following example.

\bexs[Failure of two-step method]
\label{ExaMultiTaskInfty}
Recall the multi-task observation model first introduced in
Example~\ref{ExaMultitask}.  In Corollary~\ref{CorMultiTask}, we
showed that the general estimator~\eqref{EqnSparseProb} will recover
good estimates under certain assumptions on the observation matrix.
In this example, we provide an instance for which the assumptions of
Corollary~\ref{CorMultiTask} are satisfied, but on the other hand,
the two-step method will not return a good estimate.

More specifically, let us consider the observation model $Y = \ex
(\TrueLowRank + \TrueSparse) + W$, in which $Y \in \real^{\mdim \times
  \mdim}$, and the observation matrix $\ex \in \real^{\mdim \times
  \mdim}$ takes the form
\begin{align*}
\ex & \defn I_{\mdim \times \mdim} + \frac{1}{\sqrt{\mdim}} \, e_1
\myone^T,
\end{align*}
where $e_1 \in \real^{\mdim}$ is the standard basis vector with a $1$
in the first component, and $\myone$ denotes the vector of all ones.
Suppose that the unknown low-rank matrix is given by $\TrueLowRank =
\frac{1}{\mdim} \myone \, \myone^T$.  Note that this matrix has rank
one, and satisfies $\|\TrueLowRank\|_\infty = \frac{1}{\mdim}$.

We now verify that the conditions of Corollary~\ref{CorMultiTask} are
satisfied.  Letting $\mineig$ and $\maxeig$ denote (respectively) the
smallest and largest singular values of $\ex$, we have $\mineig = 1$
and $\maxeig \leq 2$.  Moreover, letting $X_j$ denote the $j^{th}$
column of $\ex$, we have $\max_{j=1, \ldots, \mdim} \|X_j\|_2 \leq 2$.
Consequently, if we consider rescaled observations with noise variance
$\noisevar^2/d$, the conditions of Corollary~\ref{CorMultiTask} are
all satisfied with constants (independent of dimension), so that the
$M$-estimator~\eqref{EqnSparseProb} will have good performance.

On the other hand, letting $\Exs$ denote expectation over any zero-mean
noise matrix $W$, we have
\begin{align*}
\Exs \big[ \|\ex (\TrueLowRank + W)\|_\infty \big] &
\stackrel{(i)}{\geq} \| \ex (\TrueLowRank +\Exs[W]) \|_\infty \; = \|
\ex (\TrueLowRank) \|_\infty \stackrel{(ii)}{\geq} \sqrt{\mdim}
\|\TrueLowRank\|_\infty,
\end{align*}
where step (i) exploits Jensen's inequality, and step (ii) uses the
fact that 
\begin{align*}
\infnorm{\ex(\TrueLowRank)} = 1/\mdim + 1/\sqrt{\mdim} \; = \; \big(1
+ \sqrt{\mdim} \big) \|\TrueLowRank\|_\infty.
\end{align*}
For any noise matrix $W$ with reasonable tail behavior, the variable
$\| \ex( \TrueLowRank + W)\|_\infty$ will concentrate around its
expectation, showing that $\|\ex(\TrueLowRank + W)\|_\infty$ will be
larger than $\|\TrueLowRank\|_\infty$ by an order of magnitude (factor
of $\sqrt{\mdim}$).  Consequently, the two-step method will have much
larger estimation error in this case.
 \hfill \goodendex \eexs

\vspace*{.1in}


\subsection{Results for $\gennorm{\cdot}{2,1}$ regularization}
\label{SecColSparse}

Let us return again to the general Theorem~\ref{ThmConvergence}, and
illustrate some more of its consequences in application to the
columnwise $(2,1)$-norm previously defined in
Example~\ref{ExaColSparse}, and methods based on solving the convex
program~\eqref{EqnColProb}.  As before, specializing
Theorem~\ref{ThmConvergence} to this decomposable regularizer yields a
number of guarantees.  In order to keep our presentation relatively
brief, we focus here on the case of the identity observation operator
$\XopPlain = I$.
\bcors
\label{CorColDet}
Suppose that we solve the convex program~\eqref{EqnColProb} with
regularization parameters $(\regpar, \tworeg)$ such that
\begin{equation}
\label{EqnRegChoiceCol}
\regpar \geq 4 \opnorm{\Wmat}, \quad \mbox{and} \quad \tworeg \geq 4
\, \gennorm{\Wmat}{2,\infty} + \frac{ 4 \DALCON}{\sqrt{\mdimb}}.
\end{equation}
Then there is a universal constant $\plaincon_1$ such that for any
matrix pair $(\TrueLowRank, \TrueSparse)$ with
\mbox{$\gennorm{\TrueLowRank}{2,\infty} \leq
  \frac{\DALCON}{\sqrt{\mdimb}}$} and for all integers \mbox{$\rdim =
  1, 2, \ldots, \mdim$} and $\kdim = 1, 2, \ldots, \mdimb$, we have
\begin{align}
\label{EqnColDet}
\frob{\LowRankHat - \TrueLowRank}^2 + \frob{\SparseHat -
  \TrueSparse}^2 & \leq \plaincon_1 \regpar^2 \, \biggr \{ \rdim +
\frac{1}{\regpar} \, \sum_{j = \rdim+1}^\mdim \sigma_j(\TrueLowRank)
\biggr \} + \plaincon_1 \, \twopar^2 \, \biggr \{ \kdim +
\frac{1}{\twopar} \sum_{k \notin \Colset} \|\TrueSparse_{k}\|_2 \biggr
\},
\end{align}
where $\Colset \subseteq \{1,2, \ldots, \mdimb\}$ is an arbitrary subset
of column indices of cardinality at most $\kdim$.
\ecors

\paragraph{Remarks:}  
This result follows directly by specializing
Theorem~\ref{ThmConvergence} to the columnwise $(2,1)$-norm and
identity observation model, previously discussed in
Example~\ref{ExaColSparse}.  Its dual norm is the columnwise
$(2,\infty)$-norm, and we have $\DCON = \sqrt{\mdimb}$.  As discussed
in Section~\ref{SecDecomposable}, the $(2,1)$-norm is decomposable
with respect to subspaces of the type $\Model(\ColSet)$, as defined in
equation~\eqref{EqnColset}, where $\ColSet \subseteq \{1, 2, \ldots,
\mdimb \}$ is an arbitrary subset of column indices.  For any such
subset $\Colset$ of cardinality $\kdim$, it can be calculated that
$\Compat^2(\Model(\Colset)) = \kdim$, and moreover, that
$\gennorm{\Pi_{\ModelPerp}(\TrueSparse)}{2,1} = \sum_{k \notin
  \Colset} \|\TrueSparse_k\|_2$.  Consequently, the
bound~\eqref{EqnColDet} follows from Theorem~\ref{ThmConvergence}.

As before, if we assume that $\TrueLowRank$ has exactly rank $\rdim$
and $\TrueSparse$ has at most $\kdim$ non-zero columns, then both
approximation error terms in the bound~\eqref{EqnColDet} vanish, and
we recover an upper bound of the form $\frob{\LowRankHat -
  \TrueLowRank}^2 + \frob{\SparseHat - \TrueSparse}^2 \lesssim
\regpar^2 \rdim + \tworeg^2 \kdim$. If we further specialize to the
case of exact observations ($\Wmat = 0$), then
Corollary~\ref{CorColDet} guarantees that
\begin{align*}
\frob{\LowRankHat - \TrueLowRank}^2 + \frob{\SparseHat -
  \TrueSparse}^2 & \; \lesssim \; \DALCON^2 \frac{\kdim}{\mdimb}.
\end{align*}
The following example shows, that given our conditions, even in the
noiseless setting, no method can recover the matrices to precision
more accurate than $\DALCON^2 \kdim/\mdimb$. \\

\bexs[Unimprovability for columnwise sparse model]
\label{ExaColMild}
In order to demonstrate that the term $\DALCON^2 \kdim/\mdimb$ is
unavoidable, it suffices to consider a slight modification of
Example~\ref{ExaMild}.  In particular, let us define the matrix
\begin{align}
\label{EqnBadTwo}
\TrueLowRank & \defn \frac{\DALCON}{\sqrt{\mdima \mdimb}}
\begin{bmatrix} 1 \\ 1 \\ \vdots  \\ 1
\end{bmatrix}
\underbrace{\begin{bmatrix} 1 & 1 & 1 & \ldots & 0 & \ldots & 0
\end{bmatrix}}_{f^T},
\end{align}
where again the vector $f \in \real^{\mdimb}$ has $\kdim$ non-zeros.
Note that the matrix $\TrueLowRank$ is rank one, has $\kdim$ non-zero
columns, and moreover $\gennorm{\TrueLowRank}{2,\infty} =
\frac{\DALCON}{\sqrt{\mdimb}}$.  Consequently, the matrix
$\TrueLowRank$ is covered by Corollary~\ref{CorColDet}.  Since $\kdim$
columns of the matrix $\TrueSparse$ can be chosen in an arbitrary
manner, it is possible that $\TrueSparse = -\TrueLowRank$, in which
case the observation matrix $\Ymat = 0$.  This fact can be exploited
to show that \emph{no method} can achieve squared Frobenius error much
smaller \mbox{than $\approx \frac{\DALCON^2 \kdim}{\mdimb}$;} see
Section~\ref{SecMinimax} for the precise statement.  Finally, we note
that it is difficult to compare directly to the results of Xu et
al.~\cite{XuCaSa2010}, since their results do not guarantee exact
recovery of the pair $(\TrueLowRank, \TrueSparse)$.  \hfill \goodendex
\eexs

As with the case of elementwise $\ell_1$-norm, more concrete results
can be obtained when the noise matrix $\Wmat$ is stochastic.
\bcors
\label{CorColNoisy}
Suppose $\TrueLowRank$ has rank at most $\rdim$ and satisfies
\mbox{$\gennorm{\TrueLowRank}{2,\infty} \leq
  \frac{\DALCON}{\sqrt{\mdimb}}$}, and $\TrueSparse$ has at most
$\kdim$ non-zero columns.  If the noise matrix $\Wmat$ has
i.i.d. $N(0, \noisevar^2/(\mdima \mdimb))$ entries, and we solve the
convex program~\eqref{EqnColProb} with regularization parameters
$\regpar = \frac{8 \noisevar}{\sqrt{\mdima}} + \frac{8
  \noisevar}{\sqrt{\mdimb}}$ and
\begin{equation*}
\twopar = 8 \noisevar \sqrt{ \frac{1}{\mdimb}} +
\sqrt{\frac{\log\mdimb}{\mdima\mdimb}} + \frac{4 
  \DALCON}{\sqrt{\mdimb}},
\end{equation*}
then with probability greater than \mbox{$1 - \exp \big(-2 \log
  (\mdimb) \big)$,} any optimal solution $(\LowRankHat, \SparseHat)$
satisfies
\begin{align}
\label{EqnColNoisy}
e^2(\LowRankHat, \SparseHat) \leq \plaincon_1 \underbrace{\noisevar^2
  \frac{\rdim \, (\mdima + \mdimb)}{\mdima
    \mdimb}}_{\HACK_\TrueLowRank} + \underbrace{\noisevar^2 \biggr \{
  \frac{\kdim \mdima}{\mdima \mdimb} + \frac{\kdim\,
    \log\mdimb}{\mdima\mdimb} \biggr \} + \plaincon_2 \frac{\DALCON^2
    \kdim}{\mdimb}}_{\HACK_\TrueSparse}.
\end{align}
\ecors

\paragraph{Remarks:}  
Note that the setting of $\regpar$ is the same as in
Corollary~\ref{CorSparseNoisy}, whereas the parameter $\tworeg$ is
chosen based on upper bounding $\gennorm{\Wmat}{2,\infty}$,
corresponding to the dual norm of the columnwise $(2,1)$-norm.
With a slightly modified argument, the
bound~\eqref{EqnColNoisy} can be sharpened slightly by reducing the
logarithmic term to $\log (\frac{\mdimb}{\kdim})$.  As shown in
Theorem~\ref{ThmMinimax} to follow in Section~\ref{SecMinimax}, this
sharpened bound is minimax-optimal.

As with Corollary~\ref{CorSparseNoisy}, both terms in the
bound~\eqref{EqnColNoisy} are readily interpreted.  The term
$\HACK_\TrueLowRank$ has the same interpretation, as a combination of
the number of degrees of freedom in a rank $\rdim$ matrix (that is, of
the order $\rdim ( \mdima + \mdimb)$) scaled by the noise variance
$\frac{\noisevar^2}{\mdima \mdimb}$.  The second term
$\HACK_\TrueSparse$ has a somewhat more subtle interpretation.  The
problem of estimating $\kdim$ non-zero columns embedded within a
$\mdima \times \mdimb$ matrix can be split into two sub-problems:
first, the problem of estimating the $\kdim \mdima$ non-zero
parameters (in Frobenius norm), and second, the problem of column
subset selection---i.e., determining the location of the $\kdim$
non-zero parameters.  The estimation sub-problem yields the term
$\frac{\noisevar^2 \kdim \mdima}{\mdima \mdimb}$, whereas the column
subset selection sub-problem incurs a penalty involving $\log {\mdimb
  \choose \kdim} \approx \kdim \log \mdimb$, multiplied by the usual
noise variance.  The final term $\DALCON^2 \kdim/\mdimb$ arises from
the non-identifiability of the model.  As discussed in
Example~\ref{ExaColMild}, it is unavoidable without further restrictions.\\

\vspace*{.2in}

We now turn to some consequences for the problem of robust covariance
estimation formulated in Example~\ref{ExaRobustCovariance}.  As seen
from equation~\eqref{EqnSparseRC}, the disturbance matrix in this
setting can be written as a sum $(\TrueSparse)^T + \TrueSparse$, where
$\TrueSparse$ is a column-wise sparse matrix.  Consequently, we can
use a variant of the estimator~\eqref{EqnColProb}, in which the loss
function is given by $\matsnorm{\Ymat - \{ \TrueLowRank +
  (\TrueSparse)^T + \TrueSparse \}}{F}^2$. The following result
summarizes the consequences of Theorem~\ref{ThmConvergence} in this
setting:

 \bcors
\label{CorRobustCovariance}
Consider the problem of robust covariance estimation with $\numobs
\geq \mdim$ samples, based on a matrix $\TrueLowRank$ with rank at
most $\rdim$ that satisfies \mbox{$\gennorm{\TrueLowRank}{2,\infty}
  \leq \frac{\DALCON}{\sqrt{\mdim}}$}, and a corrupting matrix
$\TrueSparse$ with at most $\kdim$ rows and columns corrupted.  If we
solve SDP~\eqref{EqnColProb} with regularization parameters
\begin{equation}
\label{EqnRegparRobustCovariance}
\regpar^2 \; = \; 8 \opnorm{\TrueLowRank}^2 \; \frac{\rdim}{\numobs},
\quad \mbox{and} \quad \twopar^2 \; = \; 8 \opnorm{\TrueLowRank}^2
\frac{\rdim}{\numobs} + \frac{16 \DALCON^2}{\mdim},
\end{equation}
then with probability greater than \mbox{$1 - c_2\exp \big(-c_3 \log
  (\mdim) \big)$,} any optimal solution $(\LowRankHat, \SparseHat)$
satisfies
\begin{align*}
e^2(\LowRankHat, \SparseHat) \leq \plaincon_1 \opnorm{\TrueLowRank}^2
\big \{ \frac{\rdim^2}{\numobs} + \, \frac{\kdim \rdim}{\numobs} \big
\} + \plaincon_2 \frac{\DALCON^2 \kdim}{\mdim}.
\end{align*}
\ecors
\noindent 

Some comments about this result: with the motivation of being
concrete, we have given an explicit
choice~\eqref{EqnRegparRobustCovariance} of the regularization
parameters, involving the operator norm $\opnorm{\TrueLowRank}$, but
any upper bound would suffice. As with the noise variance in
Corollary~\ref{CorColNoisy}, a typical strategy would choose this
pre-factor by cross-validation.


\subsection{Lower bounds} 
\label{SecMinimax}

For the case of i.i.d Gaussian noise matrices,
Corollaries~\ref{CorSparseNoisy} and~\ref{CorColNoisy} provide results
of an achievable nature, namely in guaranteeing that our estimators
achieve certain Frobenius errors.  In this section, we turn to the
complementary question: what are the fundamental
(algorithmic-independent) limits of accuracy in noisy matrix
decomposition?  One way in which to address such a question is by
analyzing statistical minimax rates.

More formally, given some family $\Fam$ of matrices, the
associated minimax error is given by
\begin{align}
\MINIMAX(\Fam) & \defn \inf_{(\Ltil, \Stil)} \sup_{(\TrueLowRank,
  \TrueSparse)} \Exs \big[ \frob{\Ltil - \TrueLowRank}^2 + \frob{\Stil
    - \TrueSparse}^2 \big],
\end{align}
where the infimum ranges over all estimators $(\Ltil, \Stil)$ that are
(measurable) functions of the data $\Ymat$, and the supremum ranges
over all pairs $(\TrueLowRank, \TrueSparse) \in \Fam$.  Here the
expectation is taken over the Gaussian noise matrix $\Wmat$, under the
linear observation model~\eqref{EqnLinearObs}.

Given a matrix $\TrueSparse$, we define its support set
$\supp(\TrueSparse) \defn \{ (j,k) \, \mid \, \TrueSparse_{jk} \neq 0
\}$, as well as its column support $\colsupp(\TrueSparse) \defn \{ k
\, \mid \, \TrueSparse_k \neq 0 \big \}$, where $\TrueSparse_k$
denotes the $k^{th}$ column.  Using this notation, our interest
centers on the following two matrix families:
\begin{subequations}
\begin{align}
\Famsp(\rdim, \kdim, \DALCON) & \defn \biggr \{ (\TrueLowRank,
\TrueSparse) \; \mid \; \rank(\TrueLowRank) \leq \rdim, \;
|\supp(\TrueSparse)| \leq \kdim, \; \|\TrueLowRank\|_\infty \leq
\frac{\DALCON}{\sqrt{\mdima \mdimb}} \biggr \}, \quad \mbox{and} \\
\Famcol(\rdim, \kdim, \DALCON) & \defn \biggr \{ (\TrueLowRank,
\TrueSparse) \; \mid \; \rank(\TrueLowRank) \leq \rdim, \;
|\colsupp(\TrueSparse)| \leq \kdim, \;
\gennorm{\TrueLowRank}{2,\infty} \leq \frac{\DALCON}{\sqrt{\mdimb}}
\biggr \}.
\end{align}
\end{subequations}
By construction, Corollaries~\ref{CorSparseNoisy}
and~\ref{CorColNoisy} apply to the families $\Famsp$ and $\Famcol$
respectively. \\

\noindent The following theorem establishes lower bounds on the
minimax risks (in squared Frobenius norm) over these two families for
the identity observation operator:

\btheos
\label{ThmMinimax}

Consider the linear observation model~\eqref{EqnLinearObs} with
identity observation operator: $\Xop{\Theta + \Gamma} = \Theta +
\Gamma$. There is a universal constant $c_0 > 0$ such that for all
$\DALCON \geq 32 \sqrt{\log (\mdima \mdimb)}$, we have
\begin{align}
\label{EqnMinimaxSp}
\MINIMAX(\Famsp(\rdim, \kdim, \DALCON)) & \geq \plaincon_0 \noisevar^2
\, \biggr \{ \frac{\rdim \, (\mdima + \mdimb)}{\mdima \mdimb} +
\frac{\kdim \log (\frac{\mdima \mdimb - \kdim}{\kdim/2})}{\mdima
  \mdimb} \biggr \} + \plaincon_0 \frac{\DALCON^2 \, \kdim}{\mdima
  \mdimb},
\end{align}
and
\begin{align}
\label{EqnMinimaxCol}
\MINIMAX(\Famcol(\rdim, \kdim, \DALCON)) & \geq \plaincon_0
\noisevar^2 \, \biggr( \frac{\rdim \, (\mdima + \mdimb)}{\mdima
  \mdimb} + \frac{\kdim}{\mdimb} + \frac{\kdim \, \log( \frac{\mdimb -
    \kdim}{\kdim/2})}{\mdima \mdimb} \biggr) + \plaincon_0
\frac{\DALCON^2 \, \kdim}{\mdimb}.
\end{align}
\etheos
\noindent Note the agreement with the achievable rates guaranteed in
Corollaries~\ref{CorSparseNoisy} and~\ref{CorColNoisy} respectively.
(As discussed in the remarks following these corollaries, the
sharpened forms of the logarithmic factors follow by a more careful
analysis.)  Theorem~\ref{ThmMinimax} shows that in terms of squared
Frobenius error, the convex relaxations~\eqref{EqnSparseProb}
and~\eqref{EqnColProb} are minimax optimal up to constant factors.

In addition, it is worth observing that although
Theorem~\ref{ThmMinimax} is stated in the context of additive Gaussian
noise, it also shows that the radius of non-identifiability (involving
the parameter $\DALCON$) is a fundamental limit.  In particular, by
setting the noise variance to zero, we see that under our milder
conditions, even in the noiseless setting, no algorithm can estimate
to greater accuracy than $\plaincon_0 \frac{\DALCON^2 \, \kdim}{\mdima
  \mdimb}$, or the analogous quantity for column-sparse matrices.

\section{Simulation results}
\label{SecExperiment}

We have implemented the $M$-estimators based on the convex
programs~\eqref{EqnSparseProb} and~\eqref{EqnColProb}, in particular
by adapting first-order optimization methods due to
Nesterov~\cite{Nesterov07}.  In this section, we report simulation
results that demonstrate the excellent agreement between our
theoretical predictions and the behavior in practice.  In all cases,
we used square matrices ($\mdim = \mdima = \mdimb$), and a stochastic
noise matrix $\Wmat$ with i.i.d. $N(0, \frac{\noisevar^2}{\mdim^2})$
entries, with $\noisevar^2 = 1$.  For any given rank $\rdim$, we
generated $\TrueLowRank$ by randomly choosing the spaces of left and
right singular vectors.  We formed random sparse (elementwise or
columnwise) matrices by choosing the positions of the non-zeros
(entries or columns) uniformly at random.

Recall the estimator~\eqref{EqnSparseProb} from
Example~\ref{ExaEllOne}.  It is based on a combination of the nuclear
norm with the elementwise $\ell_1$-norm, and is motivated problem of
recovering a low-rank matrix $\TrueLowRank$ corrupted by an arbitrary
sparse matrix $\TrueSparse$.  In our first set of experiments, we
fixed the matrix dimension $\mdim = 100$, and then studied a range of
ranks $\rdim$ for $\TrueLowRank$, as well as a range of sparsity
indices $\kdim$ for $\TrueSparse$.  More specifically, we studied
linear scalings of the form $\rdim = \gamma \mdim$ for a constant
$\gamma \in (0,1)$, and $\kdim = \beta \mdim^2$ for a second constant
$\beta \in (0, 1)$.

\begin{figure}[h]
\begin{center}
\begin{tabular}{ccc}
\widgraph{.45\textwidth}{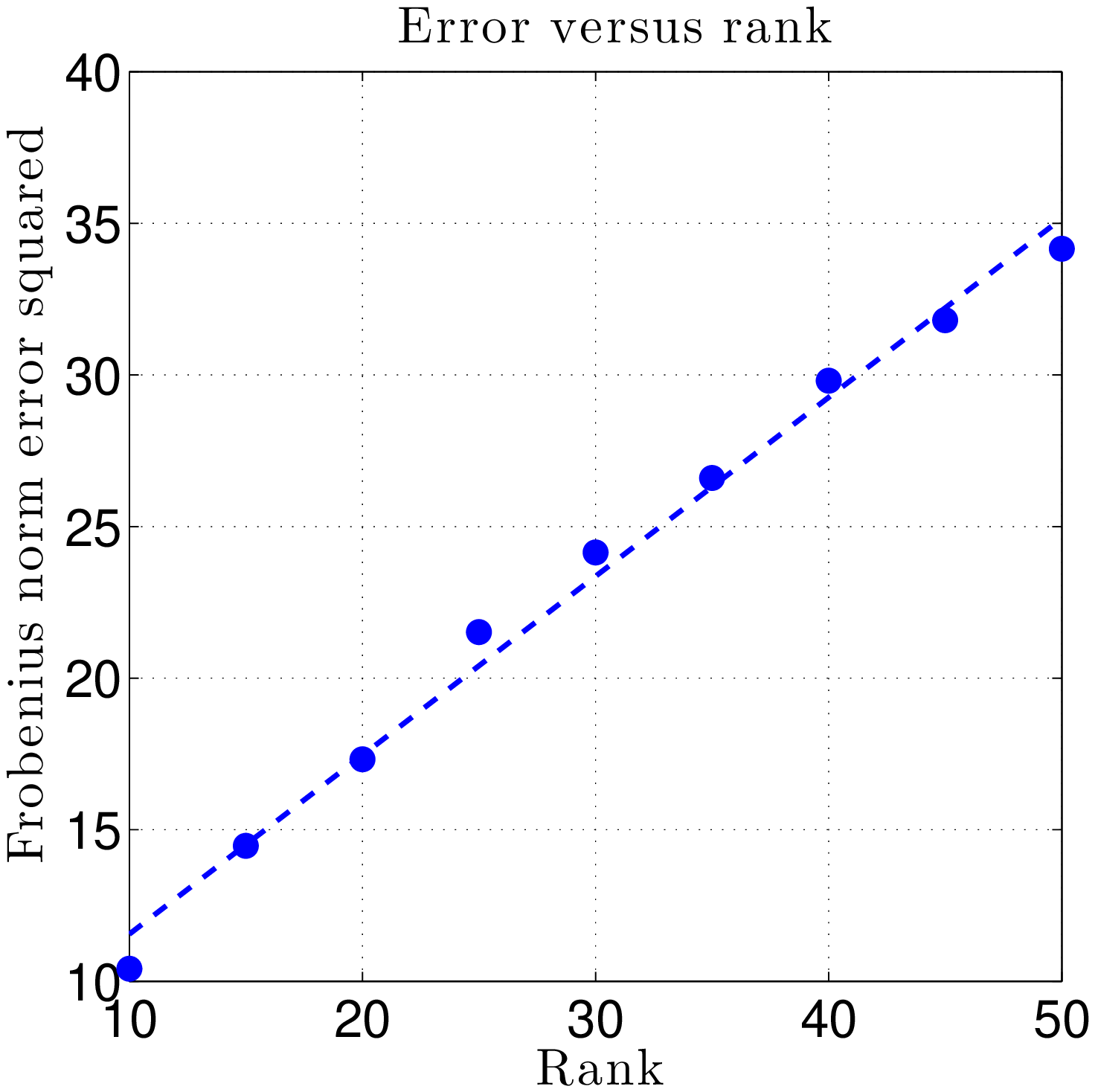} & & 
\widgraph{.45\textwidth}{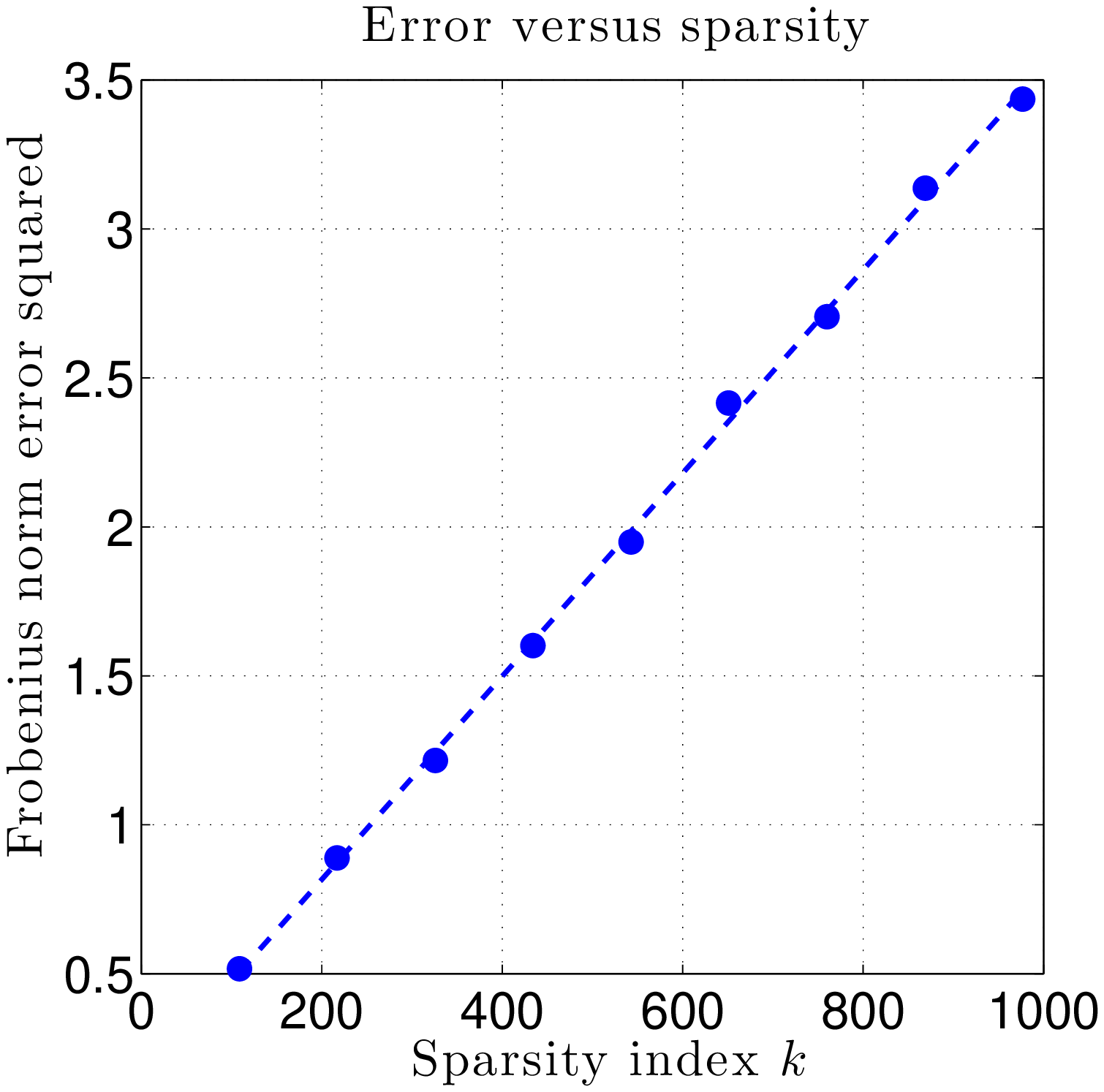} \\
(a) & & (b)
\end{tabular}
  \caption{Behavior of the estimator~\eqref{EqnSparseProb}. (a) Plot
    of the squared Frobenius error $e^2(\LowRankHat, \SparseHat)$
    versus the rank ratio $\gamma \in \{0.05:0.05:0.50 \}$, for
    matrices of size $100 \times 100$ and $\kdim = 2171$ corrupted
    entries.  The growth of the squared error is linear in $\gamma$,
    as predicted by the theory.  (b) Plot of the squared Frobenius
    error $e^2(\LowRankHat, \SparseHat)$ versus the sparsity parameter
    $\beta \in [0.01, 0.1]$ for matrices of size $100 \times 100$ and
    rank $\rdim = 10$. Consistent with the theory, the squared error
    scales approximately linearly in $\beta$ in a neighborhood around
    zero.}
  \label{FigSparse}
\end{center}
\end{figure}

Note that under this scaling, Corollary~\ref{CorSparseNoisy} predicts
that the squared Frobenius error should be upper bounded as $c_1
\gamma + c_2 \beta \log(1/\beta)$, for some universal constants $c_1$,
$c_2$.  Figure~\ref{FigSparse}(a) provides experimental confirmation
of the accuracy of these theoretical predictions: varying $\gamma$
(with $\beta$ fixed) produces linear growth of the squared error as a
function of $\gamma$.  In Figure~\ref{FigSparse}(b), we study the
complementary scaling, with the rank ratio $\gamma$ fixed and the
sparsity ratio $\beta$ varying in the interval $[.01, .1]$.  Since
$\beta \log(1/\beta) \approx \Theta(\beta)$ over this interval, we
should expect to see roughly linear scaling.  Again, the plot shows
good agreement with the theoretical predictions.

\begin{figure}[h]
\begin{center}
\begin{tabular}{ccc}
  \widgraph{0.45\textwidth}{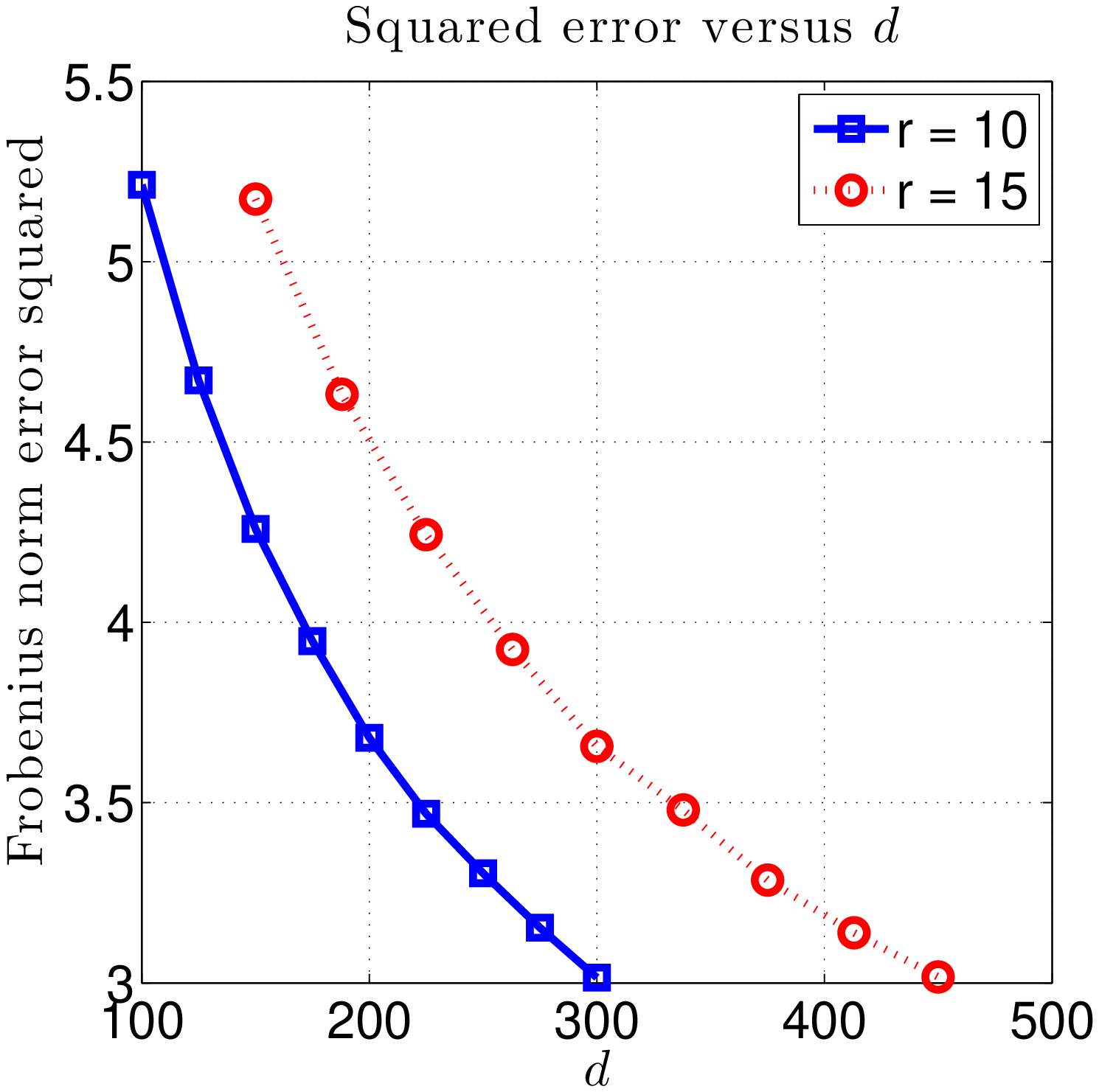} & &
  \widgraph{.45\textwidth}{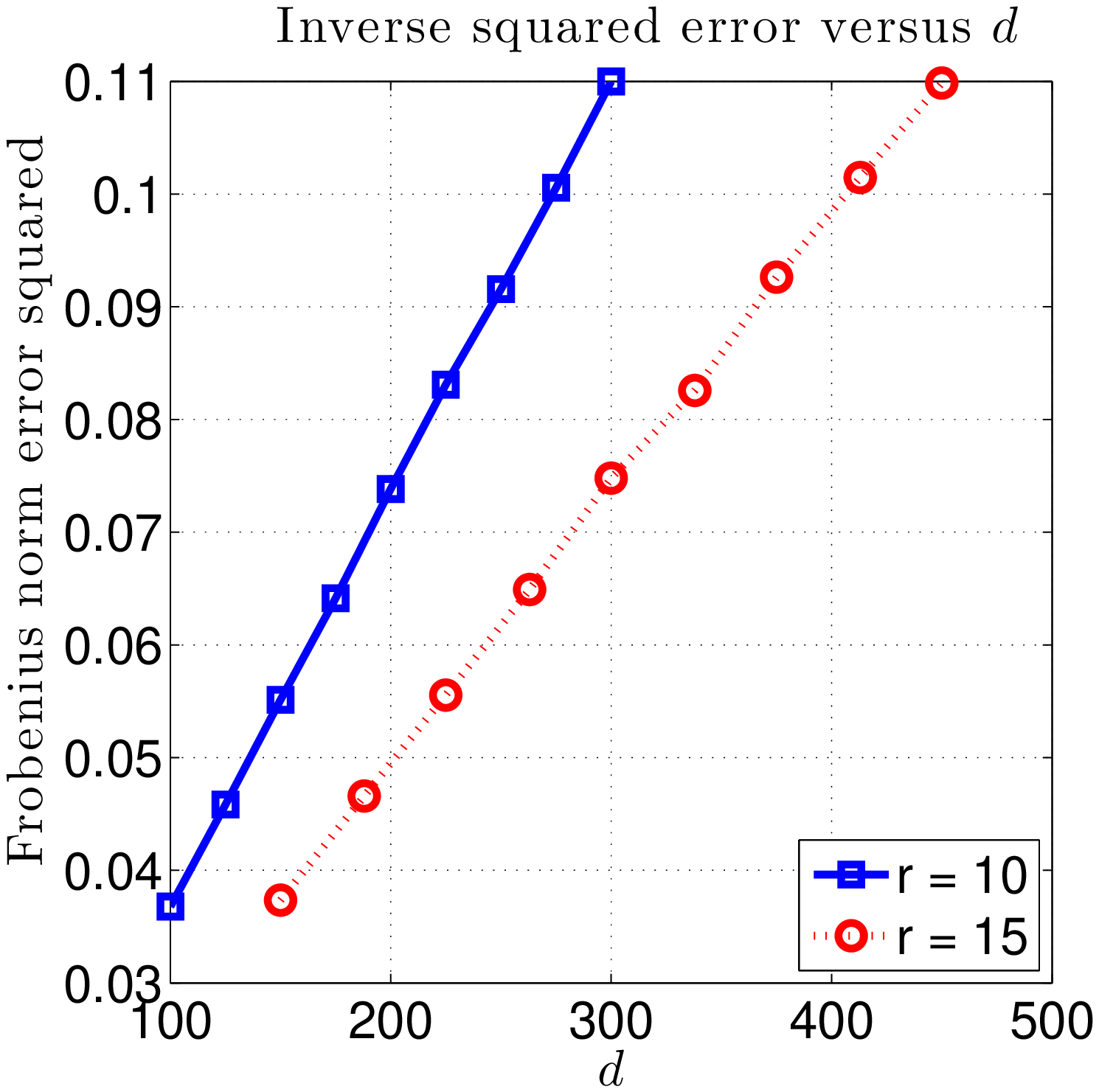} \\
(a) & & (b)
\end{tabular}
  \caption{ Behavior of the estimator~\eqref{EqnColProb}.  (a) Plot of
    the squared Frobenius error $e^2(\LowRankHat, \SparseHat)$ versus
    the dimension $\mdim \in \{100:25:300\}$, for two different
    choices of the rank ($\rdim = 10$ and $\rdim = 15$).  (b) Plot of
    the \emph{inverse} squared Frobenius error versus the dimension,
    confirming the linear scaling in $\mdim$ predicted by theory.  In
    addition, the curve for $\rdim = 15$ requires a matrix dimension
    that is $3/2$ times larger to reach the same error as the curve
    for $\rdim = 10$, consistent with theory.}
  \label{FigCol}
\end{center}
\end{figure}
Now recall the estimator~\eqref{EqnColProb} from
Example~\ref{ExaColSparse}, designed for estimating a low-rank matrix
plus a columnwise sparse matrix.  We have observed similar linear
dependence on the analogs of the parameters $\gamma$ and $\beta$, as
predicted by our theory.  In the interests of exhibiting a different
phenomenon, here we report its performance for matrices of varying
dimension, in all cases with $\TrueSparse$ having $\kdim = 3 \rdim$
non-zero columns.  Figure~\ref{FigCol}(a) shows plots of squared
Frobenius error versus the dimension for two choices of the rank
($\rdim = 10$ and $\rdim = 15$), and the matrix dimension varying in
the range $\mdim \in \{100:25:300\}$.  As predicted by our theory,
these plots decrease at the rate $1/\mdim$.  Indeed, this scaling is
revealed by replotting the inverse squared error versus $\mdim$, which
produces the roughly linear plots shown in panel (b).  Moreover, by
comparing the relative slopes of these two curves, we see that the
problem with rank $\rdim = 15$ requires roughly a dimension that is
roughly $\frac{3}{2}$ larger than the problem with $\rdim = 10$ to
achieve the same error.  Again, this linear scaling in rank is
consistent with Corollary~\ref{CorColNoisy}.


\section{Proofs}
\label{SecProofs}

In this section, we provide the proofs of our main results, with the
proofs of some more technical lemmas deferred to the appendices.

\subsection{Proof of Theorem~\ref{ThmConvergence}}

For the reader's convenience, let us recall here the two assumptions
on the regularization parameters:
\begin{align}
\label{EqnTworeg}
\tworeg \geq 4 \, \Dualreg{\XopDual{\Wmat}} + \frac{4 \,\lossrsc \,
  \DALCON}{\DCON}> 0, \quad \mbox{and} \quad \regpar \geq 4
\opnorm{\XopDual{\Wmat}}.
\end{align}
Furthermore, so as to simplify notation, let us define the error
matrices $\DelHatL \defn \LowRankHat - \TrueLowRank$ and
\mbox{$\DelHatS \defn \SparseHat - \TrueSparse$.}  Let $(\Model,
\ModelPerp)$ denote an arbitrary subspace pair for which the
regularizer $\Regplain$ is decomposable.  Throughout these proofs, we
adopt the convenient shorthand notation $\DelSModel \defn
\Pi_{\Model}(\DelHatS)$ and $\DelSPerp = \Pi_{\ModelPerp}(\DelHatS)$,
with similar definitions for $\TrueSparse_\Model$ and
$\TrueSparse_\ModelPerp$.

We now turn to a lemma that deals with the behavior of the error
matrices $(\DelHatL, \DelHatS)$ when measured together using a
weighted sum of the nuclear norm and regularizer $\Regplain$.  In
order to state the following lemma, let us recall that for any
positive $(\tworeg, \regpar)$, the weighted norm is defined as
$\BIGREG(\LowRank, \Sparse) \defn \nuclear{\LowRank} +
\frac{\tworeg}{\regpar} \Reg{\Sparse}$. \\

\noindent With this notation, we have the following: \blems
\label{LemNucDecomp}
For any $\rdim = 1, 2, \ldots, \mdim$, there is a decomposition
$\DelHatL = \DelHatLA + \DelHatLB$ such that:
\begin{enumerate}
\item[(a)] The decomposition satisfies
\begin{equation*}
\rank(\DelHatLA) \leq 2 \rdim, \quad \mbox{and} \quad (\DelHatLA)^T \;
\DelHatLB = (\DelHatLB)^T \; \DelHatLA = 0.
\end{equation*}
\item[(b)]  The difference
$\BIGREG(\TrueLowRank, \TrueSparse) - \BIGREG(\TrueLowRank + \DelHatL,
\TrueSparse + \DelHatS)$ is upper bounded by
\begin{align}
\label{EqnNarita}
\BIGREG(\DelHatLA, \DelSModel) - \BIGREG(\DelHatLB,\DelSPerp) + 2
\sum_{j=\rdim+1}^\mdim \sigma_j(\TrueLowRank) + \frac{2 \,
  \tworeg}{\regpar} \, \Reg{\TrueSparse_\ModelPerp}.
\end{align}
\item[(c)] Under conditions~\eqref{EqnTworeg} on $\tworeg$ and
  $\regpar$, the error matrices $\DelHatL$ and $\DelHatS$ satisfy
\begin{align}
\label{EqnNuclearCone}
\BIGREG \big(\DelHatLB, \DelHatS_\ModelPerp \big) & \leq 3 \BIGREG
\big(\DelHatLA, \DelHatS_\Model \big) + 4 \big \{ \sum_{j =
  \rdim+1}^\mdim \sigma_j(\TrueLowRank) + \frac{\tworeg}{\regpar} \,
\Reg{\TrueSparse_\ModelPerp} \big \}.
\end{align}
for any $\Regplain$-decomposable pair $(\Model, \ModelPerp)$.
\end{enumerate}
\elems
\noindent See Appendix~\ref{AppLemNucDecomp} for the proof of this
result.

\vspace*{.2in}

Our second lemma guarantees that the cost function $\Loss(\LowRank,
\Sparse) = \frac{1}{2} \frob{\Ymat - \Xop{\LowRank + \Sparse}}^2$ is
strongly convex in a restricted set of directions.  In particular, if
we let $\delta \Loss(\DelHatL, \DelHatS)$ denote the error in the
first-order Taylor series expansion around $(\TrueLowRank,
\TrueSparse)$, then some algebra shows that
\begin{align}
\label{EqnNewTayError}
\delta \Loss(\DelHatL, \DelHatS) & = \frac{1}{2} \frob{\Xop{\DelHatL +
    \DelHatS}}^2.
\end{align}
The following lemma shows that (up to a slack term) this Taylor error
is lower bounded by the squared Frobenius norm.

\blems[Restricted strong convexity]
\label{LemElementary}
Under the conditions of Theorem~\ref{ThmConvergence}, the first-order
Taylor series error~\eqref{EqnNewTayError} is lower bounded by
\begin{align}
\label{EqnElementary}
\frac{\lossrsc}{4} \big( \frob{\DelHatL}^2 + \frob{\DelHatS}^2 \big)
- \frac{\regpar}{2} \BIGREG(\DelHatL,\DelHatS) - 16 \, \FUNLOWL \Big
\{ \EXCESS \Big \}^2.
\end{align}
\elems
\noindent We prove this result in Appendix~\ref{AppLemElementary}. \\

\vspace*{.01in}

Using Lemmas~\ref{LemNucDecomp} and~\ref{LemElementary}, we can now
complete the proof of Theorem~\ref{ThmConvergence}.  By the optimality
of $(\LowRankHat, \SparseHat)$ and the feasibility of $(\TrueLowRank,
\TrueSparse)$, we have
\begin{align*}
  \frac{1}{2} \frob{\Ymat - \Xop{\LowRankHat + \SparseHat}}^2 + \regpar
  \nuclear{\LowRankHat} + \tworeg \Reg{ \SparseHat} & \leq \frac{1}{2}
  \frob{\Ymat - \Xop{\TrueLowRank + \TrueSparse}}^2 + \regpar
  \nuclear{\TrueLowRank} + \tworeg \Reg{\TrueSparse}.
\end{align*}
Recalling that $\Ymat = \Xop{\TrueLowRank + \TrueSparse} + \Wmat$, and
re-arranging in terms of the errors $\DelHatL = \LowRankHat -
\TrueLowRank$ and $\DelHatS = \SparseHat - \TrueSparse$, we obtain
\begin{align*}
  \label{EqnRearrange}
  \half\frob{\Xop{\DelHatL + \DelHatS}}^2 & \leq \tracer{\DelHatL +
    \DelHatS}{\XopDual{\Wmat}} + \regpar
  \BIGREG(\TrueLowRank,\TrueSparse) - \regpar
  \BIGREG(\TrueLowRank + \DelHatL,\TrueSparse + \DelHatS),
\end{align*}
where the weighted norm $\BIGREG$ was previously defined~\eqref{EqnDefnBigReg}.

We now substitute inequality~\eqref{EqnNarita} from
Lemma~\ref{LemNucDecomp} into the right-hand-side of the above
equation to obtain
\begin{multline*}
  \half\frob{\Xop{\DelHatL + \DelHatS}}^2 \; \leq \tracer{\DelHatL +
    \DelHatS}{\XopDual{\Wmat}} + \regpar \BIGREG(\DelHatLA, \DelSModel) - \regpar \BIGREG(\DelHatLB,\DelSPerp) \\+
  \EXCESSALL
\end{multline*}
Some algebra and an application of H\"older's inequality and the triangle inequality allows
us to obtain the upper bound
\begin{multline*}
  \big ( \nuclear{\DelHatLA} + \nuclear{\DelHatLB} \big ) \opnorm{\XopDual{\Wmat}} \, + \, \big ( \Reg{\DelSModel} + \Reg{\DelSPerp} \big )\, \Dualreg{\XopDual{\Wmat}} \\
  - \regpar \BIGREG(\DelHatLB,\DelSPerp) + \regpar \BIGREG(\DelHatLA, \DelSModel) + \EXCESSALL.
\end{multline*}
Recalling conditions~\eqref{EqnTworeg} for $\tworeg$ and $\regpar$, we
obtain the inequality
\begin{equation*}
  \half \frob{\Xop{\DelHatL + \DelHatS}}^2 \; \leq \; \frac{ 3
    \regpar}{2} \BIGREG(\DelHatLA, \DelSModel) + \EXCESSALL.
\end{equation*}
Using inequality~\eqref{EqnElementary} from Lemma~\ref{LemElementary}
to lower bound the right-hand side, and then rearranging terms yields
\begin{multline}
\label{EqnIntermediate}
\frac{\lossrsc}{4} \big( \frob{\DelHatL}^2 + \frob{\DelHatS}^2 \big)
\; \leq \; \frac{ 3 \regpar}{2} \BIGREG(\DelHatLA, \DelSModel) +
\frac{\regpar}{2} \BIGREG(\DelHatL,\DelHatS) \\ + 16 \, \FUNLOWL \big
\{ \EXCESS \big \}^2 + \EXCESSALL.
\end{multline}
Now note that by the triangle inequality $\BIGREG(\DelHatL,\DelHatS)
\leq \BIGREG(\DelHatLA, \DelSModel) + \BIGREG(\DelHatLB, \DelSPerp)$,
so that combined with the bound~\eqref{EqnNarita} from
Lemma~\ref{LemNucDecomp}, we obtain
\begin{align*}
\BIGREG(\DelHatL,\DelHatS) & \leq 4 \, \BIGREG(\DelHatLA, \DelSModel)
+ 4 \{ \EXCESS \}.
\end{align*}
Substituting this upper bound into equation~\eqref{EqnIntermediate}
yields
\begin{multline}
\label{EqnAlmost}
\frac{\lossrsc}{4} \big( \frob{\DelHatL}^2 + \frob{\DelHatS}^2 \big)
\; \leq \; 4 \, \BIGREG(\DelHatLA, \DelSModel) \\
+ 16 \, \FUNLOWL \big \{ \EXCESS \big \}^2 + 4 \big \{ \EXCESSNOTWO
\big \}.
\end{multline}
Noting that $\DelHatLA$ has rank at most $2 \rdim$ and that
$\DelSModel$ lies in the model space $\Model$, we find that
\begin{align*}
  \regpar \BIGREG(\DelHatLA,\DelSModel) & \leq \sqrt{2 \rdim} \,
  \regpar \frob{\DelHatLA} + \Compat(\Model) \tworeg \frob{\DelSModel}
  \\ 
& \leq \sqrt{2 \rdim} \, \regpar \frob{\DelHatL} +
  \Compat(\Model) \tworeg \frob{\DelHatS}.
\end{align*}
Substituting the above inequality into equation~\eqref{EqnAlmost} and
rearranging the terms involving $\FERRSQ{\DelHatL}{\DelHatS}$ yields
the claim.


\subsection{Proof of Corollaries~\ref{CorSparseNoisy} and~\ref{CorMultiTask}}
\label{SecProofCorSparseNoisy}

Note that Corollary~\ref{CorSparseNoisy} can be viewed as a special
case of Corollary~\ref{CorMultiTask}, in which $\numobs = \mdima$ and
$\ex = I_{\mdima \times \mdima}$.  Consequently, we may prove the
latter result, and then obtain the former result with this
specialization.  Recall that we let $\mineig$ and $\maxeig$ denote
(respectively) the minimum and maximum eigenvalues of $\ex$, and that
$\maxcol = \max_{j=1, \ldots, \mdima} \|X_j\|_2$ denotes the maximum
$\ell_2$-norm over the columns.  (In the special case $X = I_{\mdima
  \times \mdimb}$, we have $\mineig = \maxeig = \maxcol = 1$.)

Both corollaries are based on the regularizer, $\Reg{\cdot} =
\ellnorm{\cdot}$, and the associated dual norm $\Dreg{\cdot} =
\infnorm{\cdot}$.  We need to verify that the stated choices of
$(\regpar, \tworeg)$ satisfy the
requirements~\eqref{EqnRegChoiceSparse} of
Corollary~\ref{CorSparseDet}.  Given our assumptions on the pair
$(\ex, W)$, a little calculation shows that the matrix $Z \defn X^T W
\in \real^{\mdima \times \mdimb}$ has independent columns, with each
column $Z_j \sim N(0, \noisevar^2 \frac{X^T X}{\numobs})$.  Since
$\opnorm{X^T X} \leq \maxeig^2$, known results on the singular values
of Gaussian random matrices~\cite{DavSza01} imply that
\begin{align*}
  \mprob \biggr[ \opnorm{\ex^T \Wmat} \geq \frac{4 \noisevar \,
      \maxeig (\sqrt{\mdima} + \sqrt{\mdimb})}{\sqrt{\numobs}} \biggr]
  & \leq 2 \exp \big(-\plaincon (\mdima + \mdimb) \big).
\end{align*}
Consequently, setting $\regpar \geq \frac{16 \noisevar \, \maxeig
  (\sqrt{\mdima} + \sqrt{\mdimb})}{\sqrt{\numobs}}$ ensures that the
requirement~\eqref{EqnRegChoice} is satisfied.  As for the associated
requirement for $\tworeg$, it suffices to upper bound the elementwise
$\ell_\infty$ norm of $\ex^T \Wmat$.  Since the $\ell_2$ norm of the
columns of $\ex$ are bounded by $\maxcol$, the entries of $\ex^T \,
\Wmat$ are i.i.d. and Gaussian with variance at most $(\noisevar
\maxcol)^2/\numobs$. Consequently, the standard Gaussian tail bound
combined with union bound yields
\begin{align*}
\mprob \big[ \|\ex^T \, \Wmat\|_\infty \geq 4 \, \frac{\noisevar
    \maxcol}{\sqrt{\numobs}} \, \log(\mdima\mdimb) \big] & \leq \exp(-
\log \mdima\mdimb),
\end{align*}
from which we conclude that the stated choices of $(\regpar,\tworeg)$
are valid with high probability.  Turning now to the RSC condition, we
note that in the case of multivariate regression, we have
\begin{align*}
\frac{1}{2} \frob{\Xop{\Delta}}^2 & = \frac{1}{2} \frob{\ex \Delta}^2
\; \geq \; \frac{\mineig^2}{2} \frob{\Delta}^2,
\end{align*}
showing that the RSC condition holds with $\lossrsc = \mineig^2$. \\

In order to obtain the sharper result for $\ex = I_{\mdima \times
  \mdima}$ in Corollary~\ref{CorSparseNoisy}---in which $\log (\mdima
\mdimb)$ is replaced by the smaller quantity $\log(\frac{\mdima
  \mdimb}{\kdim})$--- we need to be more careful in upper bounding the
noise term $\tracer{\Wmat}{\DelHatS}$.  We refer the reader to
Appendix~\ref{AppCarefulSparse} for details of this argument.


\subsection{Proof of Corollary~\ref{CorFactor}}

For this model, the noise matrix is recentered Wishart noise---namely,
$W = \frac{1}{\numobs} \sum_{i=1}^\numobs Z_i Z_i^T - \Sigma$, where
each $Z_i \sim N(0, \Sigma)$.  Letting $U_i \sim N(0, I_{\mdim \times
  \mdim})$ be i.i.d. Gaussian random vectors, we have
\begin{align*}
\opnorm{W} & = \opnorm{\sqrt{\Sigma} \big( \frac{1}{\numobs}
  \sum_{i=1}^\numobs U_i U_i - I_{\mdim \times \mdim} \big)
  \sqrt{\Sigma}} \; \leq \; \opnorm{\Sigma} \,
\opnorm{\frac{1}{\numobs} \sum_{i=1}^\numobs U_i U_i^T - I_{\mdim
    \times \mdim}} \; \leq \; 4 \opnorm{\Sigma} \,
\sqrt{\frac{\mdim}{\numobs}},
\end{align*}
where the final bound holds with probability greater than $1 - 2
\exp(-c_1 \mdim)$, using standard tail bounds on Gaussian random
matrices~\cite{DavSza01}.  Thus, we see that the specified
choice~\eqref{EqnRegparFactor} of $\regpar$ is valid for
Theorem~\ref{ThmConvergence} with high probability.

We now turn to the choice of $\tworeg$.  The entries of $W$ are
products of Gaussian variables, and hence have sub-exponential tails
(e.g.,~\cite{BicLev08b}).  Therefore, for any entry $(i,j)$, we have
the tail bound $\mprob[|W_{ij}| > \rho(\Sigma) t ] \leq 2 \exp(-
\numobs t^2/20)$, valid for all $t \in (0,1]$.  By union bound over
  all $\mdim^2$ entries, we conclude that
\begin{align*}
\mprob \big[ \|\Wmat\|_\infty \geq 8 \rho(\Sigma) \sqrt{ \frac{\log
      \mdim}{\numobs}} \big] & \leq 2 \exp( - c_2 \log \mdim),
\end{align*}
which shows that the specified choice of $\tworeg$ is also valid
with high probability.


\subsection{Proof of Proposition~\ref{PropTwoStep}}

To begin, let us recall condition~\eqref{EqnTworeg} on the
regularization parameters, and that, for this proof, the matrices
$(\LowRankHat, \SparseHat)$ denote any optimal solutions to the
optimization problems~\eqref{EqnTwoStepSparse}
and~\eqref{EqnTwoStepLowRank} defining the two-step procedure.  We
again define the error matrices \mbox{$\DelHatL = \LowRankHat -
  \TrueLowRank$} and \mbox{$\DelHatS = \SparseHat - \TrueSparse$,} the
matrices $\DelSModel \defn \Pi_{\Model}(\DelHatS)$ and $\DelSPerp =
\Pi_{\ModelPerp}(\DelHatS)$, and the matrices $\TrueSparse_\Model$ and
$\TrueSparse_\ModelPerp$ as previously defined in the proof of
Theorem~\ref{ThmConvergence}. \\

Our proof of Proposition~\ref{PropTwoStep} is based on two lemmas, of
which the first provides control on the error $\DelHatS$ in estimating
the sparse component.
\blems
\label{LemTwoSparse}
Under the assumptions of Proposition~\ref{PropTwoStep}, for any subset
$S$ of matrix indices of cardinality at most $\kdim$, the sparse error
$\DelHatS$ in any solution of the convex
program~\eqref{EqnTwoStepSparse} satisfies the bound
\begin{align}
\label{EqnLemTwoSparse}
\frob{\DelHatS}^2 \; \leq \plaincon_1 \, \tworeg^2 \big \{ \kdim +
\frac{1}{\twopar} \sum_{(j,k) \notin S} |\TrueSparse_{jk}| \big \}.
\end{align}
\elems
\begin{proof}
Since $\SparseHat$ and $\TrueSparse$ are optimal and feasible
(respectively) for the convex program~\eqref{EqnTwoStepSparse}, we
have
\begin{equation*} 
\frac{1}{2} \frob{\SparseHat - \Ymat}^2 + \tworeg
\MYELLONE{\SparseHat} \; \leq \; \frac{1}{2} \frob{\TrueLowRank +
  \Wmat}^2 \, + \, \tworeg \MYELLONE{\TrueSparse}.
\end{equation*}
Re-writing this inequality in terms of the error $\DelHatS$ and
re-arranging terms yields
\begin{equation*}
\frac{1}{2} \frob{\DelHatS}^2 \; \leq \; |\tracer{\DelHatS}{\Wmat +
  \TrueLowRank}| \, + \tworeg \MYELLONE{\TrueSparse} - \tworeg
\MYELLONE{\TrueSparse + \DelHatS}.
\end{equation*}
By decomposability of the $\ell_1$-norm, we obtain
\begin{align*}
\frac{1}{2} \frob{\DelHatS}^2 & \leq  
|\tracer{\DelHatS}{\Wmat +
  \TrueLowRank}| \, + \tworeg \big \{ \MYELLONE{\MyModelSparse} +
\MYELLONE{\MyNoModelSparse} - \MYELLONE{\MyModelSparse + \DelHatS_S} -
\MYELLONE{\MyNoModelSparse + \DelHatS_{S^c}} \big \} \\
& \leq |\tracer{\DelHatS}{\Wmat + \TrueLowRank}| \, + 
\tworeg \big \{
2 \MYELLONE{\MyNoModelSparse} + \MYELLONE{ \DelHatS_S} -
\MYELLONE{\DelHatS_{S^c}}\big \},
\end{align*}
where the second step is based on two applications of the triangle
inequality.  Now by applying H\"older's inequality and the triangle
inequality to the first term on the right-hand side, we obtain
\begin{align*}
\frac{1}{2} \frob{\DelHatS}^2 & \leq \MYELLONE{\DelHatS} \,
     [\|\Wmat\|_\infty + \|\TrueLowRank\|_\infty] + \tworeg \big \{ 2
     \MYELLONE{\MyNoModelSparse} + \MYELLONE{ \DelHatS_S} -
     \MYELLONE{\DelHatS_{S^c}}\big \} \\
& = \MYELLONE{\DelHatS_S} \big \{ \|\Wmat\|_\infty +
     \|\TrueLowRank\|_\infty + \tworeg \} + \MYELLONE{\DelHatS_{S^c}}
     \big \{ \|\Wmat\|_\infty + \|\TrueLowRank\|_\infty - \tworeg \} +
     2 \tworeg \MYELLONE{\MyNoModelSparse} \\
& \leq 2 \tworeg \|\DelHatS_S\|_1 + 2 \tworeg
     \MYELLONE{\MyNoModelSparse},
\end{align*}
where the final inequality follows from our stated
choice~\eqref{EqnRegChoiceTwoStep} of the regularization parameter
$\tworeg$.  Since $\|\DelHatS_S\|_1 \leq \sqrt{\kdim} \frob{\DelHat_S}
\leq \sqrt{\kdim} \frob{\DelHatS}$, the claim~\eqref{EqnLemTwoSparse}
follows with some algebra.
\fpro
Our second lemma provides a bound on the low-rank error $\DelHatL$
in terms of the sparse matrix error $\DelHatS$.

\blems
\label{LemTwoLowRank}
If in addition to the conditions of Proposition~\ref{PropTwoStep}, the
sparse erorr matrix is bounded as $\frob{\DelHatS} \leq \delta$, then
the low-rank error matrix is bounded as
\begin{align}
  \label{EqnLemLowRankBound}
  \frob{\DelHatL}^2 & \leq \plaincon_1 \regpar^2 \biggr \{ \rdim +
  \frac{1}{\regpar} \sum_{j = \rdim+1}^\mdim \sigma_j(\TrueLowRank)
  \biggr \} \; + \; \plaincon_2 \delta^2.
\end{align}
\elems
\noindent As the proof of this lemma is somewhat more involved, we
defer it to Appendix~\ref{AppLemTwoLowRank}.
Finally, combining the low-rank bound~\eqref{EqnLemLowRankBound} with
the sparse bound~\eqref{EqnLemTwoSparse} from Lemma~\ref{LemTwoSparse}
yields the claim of Proposition~\ref{PropTwoStep}.


\subsection{Proof of Corollary~\ref{CorColNoisy}}

For this corollary, we have $\Reg{\cdot} = \gennorm{\cdot}{2,1}$ and
$\Dreg{\cdot} = \gennorm{\cdot}{2,\infty}$.  In order to establish the
claim, we need to show that the conditions of
Corollary~\ref{CorColDet} on the regularization pair $(\regpar,
\twopar)$ hold with high probability.  The setting of $\regpar$ is the
same as Corollary~\ref{CorSparseNoisy}, and is valid by our earlier
argument. Hence, in order to complete the proof, it remains to
establish an upper bound on $\gennorm{\Wmat}{2,\infty}$.

Let $\Wmat_k$ be the $k^{th}$ column of the matrix.  Noting that the
function $W_k \mapsto \|W_k\|_2$ is Lipschitz, by concentration of
measure for Gaussian Lipschitz functions~\cite{Ledoux01}, we have
\begin{align*}
\mprob \big[ \twonorm{\Wmat_k} \geq \Exs\twonorm{\Wmat_k} + t\big] &
\leq \exp \Big(-\frac{t^2\mdima\mdimb}{2 \noisevar^2} \Big) \qquad
\mbox{for all $t > 0$.}
\end{align*}
Using the Gaussianity of $\Wmat_k$, we have $\Exs\twonorm{\Wmat_k}
\leq \frac{\noisevar}{\sqrt{\mdima\mdimb}}\sqrt{\mdima} =
\frac{\noisevar}{\sqrt{\mdimb}}$.  Applying union bound over all
$\mdimb$ columns, we conclude that with probability greater than
\mbox{$1 - \exp \bigr(-\frac{t^2\mdima\mdimb}{2 \noisevar^2} +
  \log\mdimb\bigr)$,} we have $\max_k \twonorm{\Wmat_k} \leq
\frac{\noisevar}{\sqrt{\mdimb}} + t$.  Setting $t = 4
\noisevar\sqrt{\frac{\log\mdimb}{\mdima\mdimb}}$ yields
\begin{align*}
\mprob \biggr[ \gennorm{\Wmat}{2,\infty} \geq
  \frac{\noisevar}{\sqrt{\mdimb}} + 4
  \noisevar\sqrt{\frac{\log\mdimb}{\mdima\mdimb}} \biggr] & \leq \exp
\bigr(-3\log\mdimb\big),
\end{align*}
from which the claim follows.  

As before, a sharper bound (with $\log \mdimb$ replaced by $\log
(\mdimb/\kdim)$) can be obtained by a refined argument; we refer the
reader to Appendix~\ref{AppCarefulCol} for the details.


\subsection{Proof of Corollary~\ref{CorRobustCovariance}}

For this model, the noise matrix takes the form $\Wmat \defn
\frac{1}{\numobs} \sum_{i=1}^\numobs U_i U_i^T - \TrueLowRank$, where
\mbox{$U_i \sim N(0, \TrueLowRank)$.}   Since $\TrueLowRank$ is
positive semidefinite with rank at most $\rdim$, we can write
\begin{align*}
W & = Q \big \{ \frac{1}{\numobs} Z_i Z_i^T - I_{\rdim \times \rdim}
\big \} Q^T,
\end{align*}
where the matrix $Q \in \real^{\mdim \times \rdim}$ satisfies the
relationship $\TrueLowRank = Q Q^T$, and $Z_i \sim N(0, I_{\rdim
  \times \rdim})$ is standard Gaussian in dimension $\rdim$.
Consequently, by known results on singular values of Wishart
matrices~\cite{DavSza01}, we have $\opnorm{W} \leq \sqrt{8}
\opnorm{\TrueLowRank} \sqrt{\frac{\rdim}{\numobs}}$ with high
probability, showing that the specified choice of $\regpar$ is valid.
It remains to bound the quantity $\gennorm{\Wmat}{2, \infty}$.  By
known matrix norm bounds~\cite{Horn85}, we have $\gennorm{\Wmat}{2,
  \infty} \leq \opnorm{\Wmat}$, so that the claim follows by the
previous argument.

\subsection{Proof of Theorem~\ref{ThmMinimax}}
\label{SecProofThmMinimax}

Our lower bound proofs are based on a standard
reduction~\cite{Hasminskii78, Yu97,YanBar99} from estimation to a
multiway hypothesis testing problem over a packing set of matrix
pairs.  In particular, given a collection $\{(\LowRank^j, \Sparse^j),
j = 1, 2, \ldots, M\}$ of matrix pairs contained in some family
$\Fam$, we say that it forms a $\delta$-packing in Frobenius norm if,
for all distinct pairs $i,j \in \{1, 2, \ldots, M\}$, we have
\begin{align*}
\matsnorm{\LowRank^i - \LowRank^j}{F}^2 + \matsnorm{\Sparse^i -
  \Sparse^j}{F}^2 & \geq \delta^2.
\end{align*}
Given such a packing set, it is a straightforward consequence of
Fano's inequality that the minimax error over $\Fam$ satisfies the
lower bound
\begin{align}
\label{EqnFanoBasic}
\mprob \big[ \MINIMAX(\Fam) \geq \frac{\delta^2}{8} \big] & \geq 1 -
\frac{I(Y; J) + \log 2}{\log M},
\end{align}
where $I(Y; J)$ is the mutual information between the observation
matrix $Y \in \real^{\mdima \times \mdimb}$, and $J$ is an index
uniformly distributed over $\{1, 2, \ldots, M\}$.  In order to obtain
different components of our bound, we make different choices of the
packing set, and use different bounding techniques for the mutual
information.

\subsubsection{Lower bounds for elementwise sparsity}

We begin by proving the lower bound~\eqref{EqnMinimaxSp} for matrix
decompositions over the family $\Famsp(\rdim, \kdim, \DALCON)$.

\paragraph{Packing for radius of non-identifiability}

Let us first establish the lower bound involving the radius of
non-identifiability, namely the term scaling as $\frac{\DALCON^2
  \kdim}{\mdima \mdimb}$ in the case of $\kdim$-sparsity for
$\TrueLowRank$.  Recall from Example~\ref{ExaMild} the ``bad''
matrix~\eqref{EqnBad}, which we denote here by $\BAD$.  By
construction, we have $\matsnorm{\BAD}{F}^2 = \frac{\DALCON^2
  \kdim}{\mdima \mdimb}$.  Using this matrix, we construct a very
simple packing set with $M = 4$ matrix pairs $(\LowRank, \Sparse)$:
\begin{align}
\label{EqnSimplePack}
\big \{ (\BAD, -\BAD), \; (-\BAD, \BAD), \; (\frac{1}{\sqrt{2}} \BAD,
-\frac{1}{\sqrt{2}} \BAD), \; (0, 0) \big \}
\end{align}
Each one of these matrix pairs $(\LowRank, \Sparse)$ belongs to the
set $\Famsp(1, \kdim, \DALCON)$, so it can be used to establish a
lower bound over this set.  (Moreover, it also yields a lower bound
over the sets $\Famsp(\rdim, \kdim, \DALCON)$ for $\rdim > 1$, since
they are supersets.)  It can also be verified that for any two
distinct pairs of matrices in the set~\eqref{EqnSimplePack}, they
differ in squared Frobenius norm by at least $\delta^2 = \frac{1}{2}
\matsnorm{\BAD}{F}^2 = \frac{1}{2} \frac{\DALCON^2 \kdim}{\mdima
  \mdimb}$.  Let $J$ be a random index uniformly distributed over the
four possible models in our packing set~\eqref{EqnSimplePack}.  By
construction, for any matrix pair $(\LowRank, \Sparse)$ in the packing
set, we have $\LowRank + \Sparse = 0$.  Consequently, for any one of
these models, the observation matrix $Y$ is simply equal to pure noise
$\Wmat$, and hence $I(Y; J) = 0$.  Putting together the pieces, the
Fano bound~\eqref{EqnFanoBasic} implies that
\begin{align*}
\mprob \big[ \MINIMAX(\Famsp(1, \kdim, \DALCON)) \geq \frac{1}{16}
  \frac{\DALCON^2 \kdim}{\mdima \mdimb} \big] & \geq 1 - \frac{I(Y; J)
  + \log 2}{ \log 4} \; = \; \frac{1}{2}.
\end{align*}

\paragraph{Packing for estimation error:}
We now describe the construction of a packing set for lower bounding
the estimation error.  In this case, our construction is more subtle,
based on the the Cartesian product of two components, one for the low
rank matrices, and the other for the sparse matrices.  For the low
rank component, we re-state a slightly modified form (adapted to the
setting of non-square matrices) of Lemma 2 from the
paper~\cite{NegWai10b}:
\blems
\label{LemPacking}
For $\mdima, \mdimb \geq 10$, a tolerance $\delta > 0$, and for each
$\rdim = 1, 2, \ldots, \mdim$, there exists a set of $\mdima \times
\mdimb$-dimensional matrices $\{\Theta^1, \ldots, \Theta^{M} \}$ with
cardinality \mbox{$M \geq \frac{1}{4} \exp \big( \frac{\rdim
    \mdima}{256} + \frac{\rdim \mdimb}{256} \big)$} such that each
matrix has rank $\rdim$, and moreover
\begin{subequations}
\begin{align}
\label{EqnPackFrob}
\matsnorm{\Theta^\ell}{F}^2 & = \delta^2 \qquad \mbox{for all $\ell = 1,
  2, \ldots, M$,} \\
\label{EqnPackSpread}
\matsnorm{\Theta^\ell - \Theta^k}{F}^2 & \geq \delta^2 \qquad
\mbox{for all $\ell \neq k$,} \\
\label{EqnPackSpike}
\infnorm{\Theta^\ell} & \leq \delta \; \sqrt{\frac{32 \log (\mdima
    \mdimb)}{\mdima \mdimb}} \quad \mbox{for all $\ell = 1, 2,
  \ldots, M$.}
\end{align}
\end{subequations}
\elems
\noindent Consequently, as long as $\delta \leq 1$, we are guaranteed
that the matrices $\Theta^\ell$ belong to the set $\Famsp(\rdim,
\kdim, \DALCON)$ for all \mbox{$\DALCON \geq 32 \sqrt{\log(\mdima
    \mdimb)}$.}  \\

As for the sparse matrices, the following result is a modification, so
as to apply to the matrix setting of interest here, of Lemma 5 from
the paper~\cite{RasWaiYu09}:
\blems[Sparse matrix packing]
\label{LemSPPack}
For any $\delta > 0$, and for each integer $\kdim < \mdima \mdimb$,
there exists a set of matrices $\{\Sparse^1, \ldots, \Sparse^N\}$ with
cardinality $N \geq \exp \big(
\frac{\kdim}{2} \log \frac{\mdima \mdimb
  - \kdim}{\kdim/2} \big)$ such that
\begin{subequations}
\begin{align}
\frob{\Sparse^j - \Sparse^k}^2 & \geq \delta^2, \quad \mbox{and} \\
\frob{\Sparse^j}^2 & \leq 8 \,\delta^2,
\end{align}
\end{subequations}
and such that each $\Sparse^j$ has at most $\kdim$ non-zero entries.
\elems

We now have the necessary ingredients to prove the lower
bound~\eqref{EqnMinimaxSp}.  By combining Lemmas~\ref{LemPacking}
and~\ref{LemSPPack}, we conclude that there exists a set of matrices
with cardinality
\begin{align}
\label{EqnPackSize}
M \, N & \geq \frac{1}{4} \; \exp \biggr \{ \frac{\kdim}{2} \log
\frac{\mdima \mdimb - \kdim}{\kdim/2} + \frac{\rdim \mdima}{256} +
\frac{\rdim \mdimb}{256} \biggr \}
\end{align}
such that
\begin{subequations}
\begin{align}
\label{EqnBostonLower}
\frob{(\LowRank^\ell, \Sparse^k) - (\LowRank^{\ell'}, \Sparse^{k'})}^2
& \geq \delta^2 \qquad \mbox{for all pairs such that $\ell \neq \ell'$
  or $k \neq k'$, and} \\
\label{EqnBostonUpper}
\frob{(\LowRank^\ell, \Sparse^k)}^2 & \leq 9 \delta^2 \quad \mbox{for
  all $(\ell, k)$.}
\end{align}
\end{subequations}

Let $\mprob^{\ell, k}$ denote the distribution of the observation
matrix $\Ymat$ when $\LowRank^\ell$ and $\Sparse^k$ are the underlying
parameters.  We apply the Fano construction over the class of $M N$
such distributions, thereby obtaining that in order to show that the
minimax error is lower bounded by $\plaincon_0 \delta^2$ (for some
universal constant $\plaincon_0 > 0$), it suffices to show that
\begin{align}
\label{EqnFano}
\frac{\frac{1}{{{M N} \choose 2}} \sum \limits_{(\ell, k) \neq (\ell',
    k')} \kull{\mprob^{\ell,k}}{\mprob^{\ell', k'}} + \log 2}{ \log (M
  N) } & \leq \frac{1}{2},
\end{align}
where $\kull{\mprob^{\ell, k}}{\mprob^{\ell', k'}}$ denotes the
Kullback-Leibler divergence between the distributions $\mprob^{\ell,
  k}$ and $\mprob^{\ell', k'}$.  Given the assumption of Gaussian
noise with variance $\noisevar^2/(\mdima \mdimb)$, we have
\begin{align}
\label{EqnKLUpper}
\kull{\mprob^j}{\mprob^k} & = \frac{\mdima \mdimb}{2 \noisevar^2} \;
\frob{(\LowRank^{\ell},  \Sparse^{k}) - (\LowRank^{\ell'},
    \Sparse^{k'})}^2 \; \: \stackrel{(i)}{\leq} \; \: \frac{18 \mdima
    \mdimb \delta^2}{\noisevar^2},
\end{align}
where the bound (i) follows from the condition~\eqref{EqnBostonUpper}.
Combined with lower bound~\eqref{EqnPackSize}, we see that it suffices
to choose $\delta$ such that
\begin{align*}
\frac{ \frac{18 \mdima \mdimb \delta^2}{\noisevar^2} + \log 2} {\log
  \frac{1}{4} + \biggr \{ \frac{\kdim}{2} \log \frac{\mdima \mdimb -
    \kdim}{\kdim/2} + \frac{\rdim \mdima}{256} + \frac{\rdim
    \mdimb}{256} \biggr \}} & \leq \frac{1}{2}.
\end{align*}
For $\mdima, \mdimb$ larger than a finite constant (to exclude
degenerate cases), we see that the choice
\begin{align*}
\delta^2 & = c_0 \noisevar^2 \biggr \{ \frac{\rdim}{\mdima} +
\frac{\rdim}{\mdimb} + \frac{\kdim \log \frac{\mdima \mdimb -
    \kdim}{\kdim/2}}{\mdima \mdimb} \biggr \},
\end{align*}
for a suitably small constant $c_0 > 0$ is sufficient, thereby
establishing the lower bound~\eqref{EqnMinimaxSp}. \\

\subsubsection{Lower bounds for columnwise sparsity}
The lower bound~\eqref{EqnMinimaxCol} for columnwise follows from a
similar argument.  The only modifications are in the packing sets.

\paragraph{Packing for radius of non-identifiability}

In order to establish a lower bound of order $\frac{\DALCON^2
  \kdim}{d_2}$, recall the ``bad'' matrix~\eqref{EqnBadTwo} from
Example~\ref{ExaColMild}, which we denote by $\BAD$.  By construction,
it has squared Frobenius norm $\matsnorm{\BAD}{F}^2 = \frac{\DALCON^2
  \kdim}{\mdimb}$.  We use it to form the packing set
\begin{align}
\label{EqnSimplePackTwo}
\big \{ (\BAD, -\BAD), \; (-\BAD, \BAD), \; (\frac{1}{\sqrt{2}} \BAD,
-\frac{1}{\sqrt{2}} \BAD), \; (0, 0) \big \}
\end{align}
Each one of these matrix pairs $(\LowRank, \Sparse)$ belongs to the
set $\Famcol(1, \kdim, \DALCON)$, so it can be used to establish a
lower bound over this set.  (Moreover, it also yields a lower bound
over the sets $\Famcol(\rdim, \kdim, \DALCON)$ for $\rdim > 1$, since
they are supersets.)  It can also be verified that for any two
distinct pairs of matrices in the set~\eqref{EqnSimplePackTwo}, they
differ in squared Frobenius norm by at least $\delta^2 = \frac{1}{2}
\matsnorm{\BAD}{F}^2 = \frac{1}{2} \frac{\DALCON^2 \kdim}{ \mdimb}$.
Consequently, the same argument as before shows that
\begin{align*}
\mprob \big[ \MINIMAX(\Famcol(1, \kdim, \DALCON)) \geq \frac{1}{16}
  \frac{\DALCON^2 \kdim}{\mdimb} \big] & \geq 1 - \frac{I(Y; J) + \log
  2}{ \log 4} \; = \; \frac{1}{2}.
\end{align*}

\paragraph{Packing for estimation error:}
We now describe packings for the estimation error terms.  For the
low-rank packing set, we need to ensure that the $(2,\infty)$-norm is
controlled.  From the bound~\eqref{EqnPackSpike}, we have the
guarantee
\begin{align}
\gennorm{\Theta^\ell}{2,\infty} & \leq \delta \; \sqrt{\frac{32 \log
    (\mdima \mdimb)}{\mdimb}} \quad \mbox{for all $\ell = 1, 2,
  \ldots, M$,}
\end{align}
so that, as long as $\delta \leq 1$, the matrices $\Theta^\ell$ belong
to the set $\Famcol(\rdim, \kdim, \DALCON)$ for all \mbox{$\DALCON
  \geq 32 \sqrt{\log(\mdima \mdimb)}$.} \\

\noindent The following lemma characterizes a suitable packing set for
the columnwise sparse component:
\blems[Columnwise sparse matrix packing]
For all $\mdimb \geq 10$ and integers $\kdim$ in the set $\{1, 2,
\ldots, \mdimb-1\}$, there exists a family $\mdima \times \mdimb$
matrices $\{\Sparse^k, k = 1, 2, \ldots N \}$ with cardinality
\begin{align*}
N & \geq \exp \big( \frac{\kdim}{8} \log \frac{\mdimb -
  \kdim}{\kdim/2} + \frac{\kdim \mdima}{8} \big),
\end{align*}
satisfying the inequalities
\begin{subequations}
\begin{align}
\frob{\Sparse^j - \Sparse^k}^2 & \geq \delta^2, \quad \mbox{for all $j
  \neq k$, and} \\
\frob{\Sparse^j}^2 & \leq 64 \,\delta^2,
\end{align}
\end{subequations}
and such that each $\Sparse^j$ has at most $\kdim$ non-zero columns.
\elems
\noindent This claim follows by suitably adapting Lemma 5(b) in the
paper by Raskutti et al.~\cite{RasWaiYu10b} on minimax rates for
kernel classes.  In particular, we view column $j$ of a matrix
$\Sparse$ as defining a linear function in dimension $\real^{\mdima}$;
for each $j = 1, 2, \ldots, \mdima$, this defines a Hilbert space
$\mathcal{H}_j$ of functions. By known results on metric entropy of
Euclidean balls~\cite{Matousek}, this function class has logarithmic
metric entropy, so that part (b) of the above lemma applies, and
yields the stated result. \\

Using this lemma and the packing set for the low-rank component and
following through the Fano construction yields the claimed lower
bound~\eqref{EqnMinimaxSp} on the minimax error for the class
$\Famcol(\rdim, \kdim, \DALCON)$, which completes the proof of
Theorem~\ref{ThmMinimax}.

\section{Discussion}
\label{SecDiscuss}

In this paper, we analyzed a class of convex relaxations for solving a
general class of matrix decomposition problems, in which the goal is
recover a pair of matrices, based on observing a noisy contaminated
version of their sum.  Since the problem is ill-posed in general, it
is essential to impose structure, and this paper focuses on the
setting in which one matrix is approximately low-rank, and the second
has a complementary form of low-dimensional structure enforced by a
decomposable regularizer.  Particular cases include matrices that are
elementwise sparse, or columnwise sparse, and the associated matrix
decomposition problems have various applications, including robust
PCA, robustness in collaborative filtering, and model selection in
Gauss-Markov random fields.  We provided a general non-asymptotic
bound on the Frobenius error of a convex relaxation based on a
regularizing the least-squares loss with a combination of the nuclear
norm with a decomposable regularizer.  When specialized to the case of
elementwise and columnwise sparsity, these estimators yield rates that
are minimax-optimal up to constant factors.

Various extensions of this work are possible.  We have not discussed
here how our estimator would behave under a partial observation model,
in which only a fraction of the entries are observed.  This problem is
very closely related to matrix completion, a problem for which recent
work by Negahban and Wainwright~\cite{NegWai10b} shows that a form of
restricted strong convexity holds with high probability.  This
property could be adapted to the current setting, and would allow for
proving Frobenius norm error bounds on the low rank component.
Finally, although this paper has focused on the case in which the
first matrix component is approximately low rank, much of our theory
could be applied to a more general class of matrix decomposition
problems, in which the first component is penalized by a decomposable
regularizer that is ``complementary'' to the second matrix component.
It remains to explore the properties and applications of these
different forms of matrix decomposition.

\section*{Acknowledgements}

All three authors were partially supported by grant AFOSR-09NL184.  In
addition, SN and MJW were partially supported by grant
NSF-CDI-0941742, and AA was partially supported a Microsoft Research
Graduate Fellowship. All three authors would like to acknowledge the
Banff International Research Station (BIRS) in Banff, Canada for
hospitality and work facilities that stimulated and supported this
collaboration.


\appendix

\section{Proof of Lemma~\ref{LemNucDecomp}}
\label{AppLemNucDecomp}

The decomposition described in part (a) was established by Recht et
al.~\cite{RecFazPar10}, so that it remains to prove part (b).  With
the appropriate definitions, part (b) can be recovered by exploiting
Lemma 1 from Negahban et al.~\cite{NegRavWaiYu09}.  Their lemma
applies to optimization problems of the general form
\begin{equation*}
\min_{\theta \in \real^\pdim} \big \{ \Loss(\theta) + \gamma_\numobs
r(\theta) \big \},
\end{equation*}
where $\Loss$ is a loss function on the parameter space, and $r$ is
norm-based regularizer that satisfies a property known as
decomposability.  The elementwise $\ell_1$-norm as well as the nuclear
norm are both instances of decomposable regularizers.  Their lemma
requires that the regularization parameter $\gamma_\numobs$ be chosen
such that $\gamma_\numobs \geq 2 \,r^* \big( \nabla \Loss(\theta^*)
\big)$, where $r^*$ is the dual norm, and $\nabla \Loss(\theta^*)$
is the gradient of the loss evaluated at the true parameter.

We now discuss how this lemma can be applied in our special case.
Here the relevant parameters are of the form $\theta = (\LowRank,
\Sparse)$, and the loss function is given by 
\begin{align*}
\Loss(\LowRank, \Sparse) & = \frac{1}{2}\frob{\Ymat - (\LowRank +
  \Sparse)}^2.
\end{align*}
The sample size $\numobs = \mdim^2$, since we make one observation for
each entry of the matrix.  On the other hand, the regularizer is given
by the function
\begin{align*}
r(\theta) & = \BIGREG(\LowRank, \Sparse) \; \defn \;
\nuclear{\LowRank} + \frac{\tworeg}{\regpar} \Reg{\Sparse},
\end{align*}
coupled with the regularization parameter $\gamma_\numobs = \regpar$.
By assumption, the regularizer $\Regplain$ is decomposable, and as
shown in the paper~\cite{NegRavWaiYu09}, the nuclear norm is also
decomposable.  Since $\BIGREG$ is simply a sum of these decomposable
regularizers over separate matrices, it is also decomposable.

It remains to compute the gradient $\nabla \Loss(\TrueLowRank,
\TrueSparse)$, and evaluate the dual norm.  A straightforward
calculation yields that $\nabla \Loss(\TrueLowRank, \TrueSparse)
= \begin{bmatrix} \Wmat & \Wmat 
\end{bmatrix}^T$.    In addition, it can be verified by
standard properties of dual norms
\begin{align*}
\BIGREG^*(U, V) = \opnorm{U} + \frac{\regpar}{\twopar} \Dreg{V}.
\end{align*}
Thus, it suffices to choose the regularization parameter such that
\begin{align*}
\regpar & \geq 2 \BIGREG^*(\Wmat, \Wmat) = 2 \opnorm{\Wmat} + \frac{2
  \regpar}{\twopar} \Dreg{\Wmat}.
\end{align*}
Given our condition~\eqref{EqnTworeg}, we have
\begin{align*}
2 \opnorm{\Wmat} + \frac{2 \regpar}{\twopar} \Dreg{\Wmat} & \leq 2
\opnorm{\Wmat} + \frac{\regpar}{2},
\end{align*}
meaning that it suffices to have $\regpar \geq 4 \opnorm{\Wmat}$, as
stated in the second part of condition~\eqref{EqnTworeg}.


\section{Proof of Lemma~\ref{LemElementary}}
\label{AppLemElementary}

By the RSC condition~\eqref{EqnDefnRSC}, we have
\begin{align}
\label{EqnRSC1}
\frac{1}{2} \frob{\Xop{\DelHatL + \DelHatS}}^2 - \frac{\lossrsc}{2}
\frob{\DelHatL + \DelHatS}^2 & \geq - \FUNLOWL \RegInfConvSq{\DelHatL
  + \DelHatS} \; \geq \; - \FUNLOWL \BIGREG^2(\DelHatL,\DelHatS),
\end{align}
where the second inequality follows by the
definitions~\eqref{EqnDefnBigReg} and~\eqref{EqnDefnRegInfConv} of
$\BIGREG$ and $\RegInfConvPlain$ respectively.  We now derive a lower
bound on $\matsnorm{\DelHatL + \DelHatS}{F}$, and an upper bound on
$\BIGREG^2(\DelHatL, \DelHatS)$.  Beginning with the former term,
observe that
\begin{align*}
\frac{\lossrsc}{2} \big(\matsnorm{\DelHatL}{F}^2 +
\matsnorm{\DelHatS}{F}^2\big) - \frac{\lossrsc}{2} \matsnorm{\DelHatL
  + \DelHatS}{F}^2 & = - \lossrsc \tracer{\DelHatL}{\DelHatS},
\end{align*}
so that it suffices to upper bound $ \lossrsc
|\tracer{\DelHatL}{\DelHatS}|$.  By the duality of the pair
$(\Regplain, \Dregplain)$, we have
\begin{align*}
\lossrsc \big|\tracer{\DelHatL}{\DelHatS} \big| & \leq \lossrsc
\Dreg{\DelHatL} \; \Reg{\DelHatS}.
\end{align*}
Now since $\LowRankHat$ and $\TrueLowRank$ are both feasible for the
program~\eqref{EqnGenProg} and recalling that $\DelHatL = \LowRankHat
- \TrueLowRank$, an application of triangle inequality yields
\begin{align*}
\lossrsc \Dreg{\DelHatL} & \leq \lossrsc \big \{ \Dreg{\LowRankHat} +
\Dreg{\TrueLowRank} \big \} \; \leq \; \frac{2 \DALCON \,
  \lossrsc}{\DCON} \; \stackrel{(i)}{\leq} \; \frac{\tworeg}{2},
\end{align*}
where inequality (i) follows from our choice of $\tworeg$.  Putting
together the pieces, we have shown that
\begin{align*}
\frac{\lossrsc}{2} \frob{\DelHatL + \DelHatS}^2 & \geq
\frac{\lossrsc}{2} \big(\matsnorm{\DelHatL}{F}^2 +
\matsnorm{\DelHatS}{F}^2\big) - \frac{\tworeg}{2} \Reg{\DelHatS}.
\end{align*}
Since the quantity $\regpar \nuclear{\DelHatL} \geq 0$, we can write
\begin{align*}
\frac{\lossrsc}{2} \frob{\DelHatL + \DelHatS}^2 & \geq
\frac{\lossrsc}{2} \, \big(\matsnorm{\DelHatL}{F}^2 +
\matsnorm{\DelHatS}{F}^2\big) - \frac{\tworeg}{2} \Reg{\DelHatS} -
\frac{\regpar}{2} \nuclear{\DelHatL} \\
& = \frac{\lossrsc}{2} \big(\matsnorm{\DelHatL}{F}^2 +
\matsnorm{\DelHatS}{F}^2\big) - \frac{\regpar}{2}
\BIGREG(\DelHatL,\DelHatS),
\end{align*}
where the latter equality follows by the
definition~\eqref{EqnDefnBigReg} of $\BIGREG$.

Next we turn to the upper bound on $\BIGREG(\DelHatL,\DelHatS)$. By
the triangle inequality, we have
\begin{align*}
  \BIGREG(\DelHatL,\DelHatS) & \leq \BIGREG(\DelHatLA, \DelSModel) +
  \BIGREG(\DelHatLB, \DelSPerp).
\end{align*}
Furthermore, substituting in equation~\eqref{EqnNarita} into the above
equation yields
\begin{align}
\label{EqnNice}
\BIGREG(\DelHatL,\DelHatS) & \leq 4 \, \BIGREG(\DelHatLA, \DelSModel)
+ 4 \{ \EXCESS \}.
\end{align}
Since $\DelHatLA$ has rank at most $2 \rdim$ and $\DelSModel$ belongs
to the model space $\Model$, we have
\begin{align*}
\regpar \BIGREG(\DelHatLA,\DelSModel) & \leq \sqrt{2 \rdim} \, \regpar
\frob{\DelHatLA} + \Compat(\Model) \tworeg \frob{\DelSModel} \\ 
& \leq \sqrt{2 \rdim} \, \regpar \frob{\DelHatL} + \Compat(\Model)
\tworeg \frob{\DelHatS}.
\end{align*}
The claim then follows by substituting the above equation into
equation~\eqref{EqnNice}, and then substituting the result into the
earlier inequality~\eqref{EqnRSC1}.






\section{Refinement of achievability results}
\label{AppCareful}

In this appendix, we provide refined arguments that yield sharpened
forms of Corollaries~\ref{CorSparseNoisy} and~\ref{CorColNoisy}.
These refinements yield achievable bounds that match the minimax lower
bounds in Theorem~\ref{ThmMinimax} up to constant factors.  We note
that these refinements are significantly different \emph{only when}
the sparsity index $\kdim$ scales as $\Theta(\mdima \mdimb)$ for
Corollary~\ref{CorSparseNoisy}, or as $\Theta(\mdimb)$ for
Corollary~\ref{CorColNoisy}.

\subsection{Refinement of Corollary~\ref{CorSparseNoisy}}
\label{AppCarefulSparse}

In the proof of Theorem~\ref{ThmConvergence}, when specialized to the
$\ell_1$-norm, the noise term $|\tracer{\Wmat}{\DelHatS}|$ is simply
upper bounded by $\|\Wmat\|_\infty \|\DelHatS\|_1$.  Here we use a
more careful argument to control this noise term.  Throughout the
proof, we assume that the regularization parameter $\regpar$ is set in
the usual way, whereas we choose
\begin{align}
\label{EqnRefinedTwoparSparse}
\twopar & = 16 \noisevar \, \sqrt{ \frac{\log \frac{\mdima
      \mdimb}{\kdim}}{\mdima \mdimb}} + \frac{4 \DALCON}{\sqrt{\mdima
    \mdimb}}.
\end{align}

We split our analysis into two cases.

\paragraph{Case 1:}  First, 
suppose that $\ellnorm{\DelHatS} \leq \sqrt{\kdim} \frob{\DelHatS}$.
In this case, we have the upper bound
\begin{align*}
\big| \tracer{\Wmat}{\DelHatS} \big | & \leq
\sup_{\substack{\ellnorm{\Delta} \leq \sqrt{\kdim} \, \frob{\DelHatS}
    \\ \frob{\Delta} \leq \frob{\DelHatS}}} |\tracer{\Wmat}{\Delta}|
\; = \; \frob{\DelHatS} \;
\underbrace{\sup_{\substack{\ellnorm{\Delta} \leq \sqrt{\kdim}
      \\ \frob{\Delta} \leq 1}} |\tracer{\Wmat}{\Delta}|}_{Z(\kdim)}
\end{align*}
It remains to upper bound the random variable $Z(\kdim)$.  Viewed as a
function of $\Wmat$, it is a Lipschitz function with parameter
$\frac{\noisevar}{\sqrt{\mdima \mdimb}}$, so that
\begin{align*}
\mprob \big[Z(\kdim) \geq \Exs[Z(\kdim)] + \delta] & \leq \exp \Big(-
\frac{\mdima \,\mdimb\delta^2}{2 \noisevar^2} \Big).
\end{align*}
Setting $\delta^2 = \frac{4 \kdim \noisevar^2}{\mdima
  \mdimb}\log(\frac{\mdima \mdimb}{\kdim})$, we have
\begin{align*}
Z(\kdim) & \leq \Exs[Z(\kdim)] + \frac{2\kdim \noisevar}{\mdima
  \mdimb} \biggr[\log \biggr (\frac{\mdima \mdimb}{\kdim} \biggr ) \biggr]
\end{align*}
with probability greater than $1 - \exp \big(-2 \kdim
\log(\frac{\mdima \mdimb}{\kdim}) \big)$.

It remains to upper bound the expected value.  In order to do so, we
apply Theorem 5.1(ii) from Gordon et al.~\cite{Gor07} with $(q_0, q_1)
= (1,2)$, $n = \mdima \mdimb$ and $t = \sqrt{\kdim}$, thereby
obtaining
\begin{align*}
\Exs[Z(t)] & \leq \plaincon' \: \frac{\noisevar}{\sqrt{\mdima \mdimb}}
\; \sqrt{\kdim} \sqrt{2 + \log \biggr ( \frac{2 \mdima \mdimb}{\kdim}
  \biggr)} \; \leq \plaincon \; \frac{\noisevar}{\sqrt{\mdima\mdimb}} \;
\sqrt{\kdim \, \log \biggr(\frac{\mdima \mdimb}{\kdim} \biggr)}.
\end{align*}
With this bound, proceeding through the remainder of the proof yields
the claimed rate. \\

\paragraph{Case 2:}
Alternatively, we must have $\ellnorm{\DelHatS} > \sqrt{\kdim}
\frob{\DelHatS}$.  In this case, we need to show that the stated
choice~\eqref{EqnRefinedTwoparSparse} of $\tworeg$ satisfies $\tworeg
\ellnorm{\DelHatS} \geq 2 |\tracer{\Wmat}{\DelHatS}|$ with high
probability.  As can be seen from examining the proofs, this condition
is sufficient to ensure that Lemma~\ref{LemNucDecomp} and
Lemma~\ref{LemElementary} all hold, as required for our analysis.

We have the upper bound
\begin{align*}
\big| \tracer{\Wmat}{\DelHatS} \big | & \leq
\sup_{\substack{\ellnorm{\Delta} \leq \ellnorm{\DelHatS}
    \\ \frob{\Delta} \leq \frob{\DelHatS}}} |\tracer{\Wmat}{\Delta}|
\; = \; \frob{\DelHatS} \; Z \biggr(
\frac{\ellnorm{\DelHatS}}{\frob{\DelHatS}} \biggr),
\end{align*}
where for any radius $t > 0$, we define the random variable
\begin{align*}
Z(t) & \defn \sup_{\substack{\ellnorm{\Delta} \leq t \\ \frob{\Delta}
    \leq 1}} |\tracer{\Wmat}{\Delta}|.
\end{align*}
For each fixed $t$, the same argument as before shows that $Z(t)$ is
concentrated around its expectation, and Theorem 5.1(ii) from Gordon
et al.~\cite{Gor07} with $(q_0, q_1) = (1,2)$, $n = \mdima \mdimb$
yields
\begin{align*}
\Exs \big[Z(t) \big] & \leq \plaincon \;
\frac{\noisevar}{\sqrt{\mdima\mdimb}} \; t \, \sqrt{\log
  \biggr(\frac{\mdima \mdimb}{t^2} \biggr)}.
\end{align*}
Setting $\delta^2 = \frac{4 t^2 \noisevar^2}{\mdima
  \mdimb}\log(\frac{\mdima \mdimb}{\kdim})$ in the concentration
bound, we conclude that
\begin{align*}
Z(t) & \leq c' \, t \frac{\noisevar}{\sqrt{\mdima \mdimb}} \biggr
\{\sqrt{\log \biggr (\frac{\mdima \mdimb}{\kdim}\biggr)} + \sqrt{\log
  \biggr(\frac{\mdima \mdimb}{t^2} \biggr)} \biggr\}.
\end{align*}
with high probability.  A standard peeling argument
(e.g.,~\cite{vandeGeer}) can be used to extend this bound to a uniform
one over the choice of radii $t$, so that it applies to the random one
$t = \frac{\ellone{\DelHatS}}{\frob{\DelHatS}}$ of interest.  (The
only changes in doing such a peeling are in constant terms.) We thus
conclude that
\begin{align*}
Z \Bigg( \frac{\ellnorm{\DelHatS}}{\frob{\DelHatS}} \Bigg) & \leq c' \,
\frac{\ellnorm{\DelHatS}}{\frob{\DelHatS}}
\frac{\noisevar}{\sqrt{\mdima \mdimb}} \biggr
\{\sqrt{\log \biggr (\frac{\mdima \mdimb}{\kdim} \biggr )} + \sqrt{\log
  \bigg(\frac{\mdima \mdimb}{\ellnorm{\DelHatS}^2/\frob{\DelHatS}^2 }
  \bigg)} \biggr\}
\end{align*}
with high probability.  Since $\ellnorm{\DelHatS} > \sqrt{\kdim}
\frob{\DelHatS}$, we have
$\frac{1}{\ellnorm{\DelHatS}^2/\frob{\DelHatS}^2} \leq
\frac{1}{\kdim}$, and hence
\begin{align*}
\big| \tracer{\Wmat}{\DelHatS} \big| \; \leq \; \frob{\DelHatS} \; Z
\biggr( \frac{\ellnorm{\DelHatS}}{\frob{\DelHatS}} \biggr) & \leq c'' \,
\ellone{\DelHatS} \; \frac{\noisevar}{\sqrt{\mdima \mdimb}} \;
\sqrt{\log \biggr (\frac{\mdima \mdimb}{\kdim}\biggr )}
\end{align*}
with high probability.  With this bound, the remainder of the proof
proceeds as before.  In particular, the refined
choice~\eqref{EqnRefinedTwoparSparse} of $\twopar$ is adequate.

\subsection{Refinement of Corollary~\ref{CorColNoisy}}
\label{AppCarefulCol}

As in the refinement of Corollary~\ref{CorSparseNoisy} from
Appendix~\ref{AppCarefulSparse}, we need to be more careful in
controlling the noise term $\tracer{\Wmat}{\DelHatS}$.  For this
corollary, we make the refined choice of regularizer
\begin{align}
\label{EqnRefinedTwoparCol}
\twopar & = 16 \noisevar \sqrt{ \frac{1}{\mdimb}} + 16 \noisevar
\sqrt{\frac{\log(\mdimb/\kdim)}{\mdima\mdimb}} + \frac{4
  \DALCON}{\sqrt{\mdimb}}
\end{align}
As in Appendix~\ref{AppCarefulSparse}, we split our analysis into two
cases.

\paragraph{Case 1:}  First, suppose that 
$\gennorm{\DelHatS}{2,1} \leq \sqrt{\kdim} \frob{\DelHatS}$.  In this
case, we have
\begin{align*}
\big| \tracer{\Wmat}{\DelHatS} \big | & \leq
\sup_{\substack{\gennorm{\Delta}{2,1} \leq \sqrt{\kdim} \,
    \frob{\DelHatS} \\ \frob{\Delta} \leq \frob{\DelHatS}}}
|\tracer{\Wmat}{\Delta}| \; = \; \frob{\DelHatS} \;
\underbrace{\sup_{\substack{\gennorm{\Delta}{2,1} \leq \sqrt{\kdim}
      \\ \frob{\Delta} \leq 1}}
  |\tracer{\Wmat}{\Delta}|}_{\Ztil(\kdim)}
\end{align*}

The function $\Wmat \mapsto \Ztil(\kdim)$ is a Lipschitz function with
parameter $\frac{\noisevar}{\sqrt{\mdima \mdimb}}$, so that by
concentration of measure for Gaussian Lipschitz
functions~\cite{Ledoux01}, it satisfies the upper tail bound
\mbox{$\mprob \big[\Ztil (\kdim) \geq \Exs[\Ztil(\kdim)] + \delta]
  \leq \exp \big(- \frac{\mdima \mdimb \delta^2}{2 \noisevar^2}
  \big)$.}  Setting $\delta^2 = \frac{4 \kdim \noisevar^2}{\mdima
  \mdimb} \log(\frac{\mdimb}{\kdim})$ yields
\begin{align}
\label{EqnBARTConc}
\Ztil(\kdim) & \leq \Exs[\Ztil(\kdim)] + 2 \noisevar \sqrt{\frac{\kdim
    \log (\frac{\mdimb}{\kdim})}{\mdima \mdimb}}
\end{align}
with probability greater than $1 - \exp \big(-2 \kdim
\log(\frac{\mdimb}{\kdim}) \big)$.  

It remains to upper bound the expectation.  Applying the
Cauchy-Schwarz inequality to each column, we have
\begin{align*}
\Exs [\Ztil(\kdim)] & \leq \Exs
\biggr[\sup_{\substack{\gennorm{\Delta}{2,1} \leq \sqrt{\kdim}
      \\ \frob{\Delta} \leq 1}} \quad \sum_{k = 1}^\mdimb
  \|\Wmat_k\|_2 \; \|\Delta_k\|_2 \biggr] \\
& = \Exs \biggr[\sup_{\substack{\gennorm{\Delta}{2,1} \leq
      \sqrt{\kdim} \\ \frob{\Delta} \leq 1}} \quad \sum_{k = 1}^\mdimb
  \big(\|\Wmat_k\|_2 - \Exs[\|\Wmat_k\|_2] \big)\; \|\Delta_k\|_2
  \biggr] + \sup_{\substack{\gennorm{\Delta}{2,1} \leq \sqrt{\kdim}}}
\biggr (\sum_{k = 1}^\mdim \|\Delta_k\|_2 \biggr) \Exs[\|W_1\|_2] \\
& \leq \Exs \biggr[\sup_{\substack{\gennorm{\Delta}{2,1} \leq
      \sqrt{\kdim} \\ \frob{\Delta} \leq 1}} \quad \sum_{k = 1}^\mdimb
  \underbrace{\big(\|\Wmat_k\|_2 - \Exs[\|\Wmat_k\|_2] \big)}_{V_k}\;
  \|\Delta_k\|_2 \biggr] + 4 \noisevar \; \sqrt{\frac{\kdim}{\mdimb}},
\end{align*}
using the fact that $\Exs[\|\Wmat_1\|_2] \leq \noisevar
\sqrt{\frac{\mdima}{\mdimb \mdimb}} \; = \;
\frac{\noisevar}{\sqrt{\mdimb}}$.

Now the variable $V_k$ is zero-mean, and sub-Gaussian with parameter
$\frac{\noisevar}{\sqrt{\mdima \mdimb}}$, again using concentration of
measure for Lipschitz functions of Gaussians~\cite{Ledoux01}.
Consequently, by setting $\delta_k = \|\Delta_k\|_2$, we can write
\begin{align*}
\Exs [\Ztil(\kdim)] & \leq \Exs \biggr[\sup_{\substack{ \|\delta\|_1
      \leq 4 \sqrt{\kdim} \\ \|\delta\|_2 \leq 1}} \quad \sum_{k =
    1}^\mdimb V_k \delta_k \biggr] + 4 \noisevar \;
\sqrt{\frac{\kdim}{\mdimb}},
\end{align*}

Applying Theorem 5.1(ii) from Gordon et al.~\cite{Gor07} with $(q_0,
q_1) = (1,2)$, $n = \mdimb$ and $t = 4 \sqrt{\kdim}$ then yields
\begin{align*}
\Exs[\Ztil(\kdim)] & \leq \plaincon \: \frac{\noisevar}{\sqrt{ \mdima
    \mdimb}} \; \sqrt{\kdim} \sqrt{2 + \log \bigg( \frac{2 \mdimb}{16
    \kdim} \bigg)} + 4 \noisevar \; \sqrt{\frac{\kdim}{\mdimb}},
\end{align*}
which combined with the concentration bound~\eqref{EqnBARTConc}
yields the refined claim.

\paragraph{Case 2:}  Alternatively, we may assume that $\gennorm{\DelHatS}{2,1} > \sqrt{\kdim} \frob{\DelHatS}$.  In this
case, we need to verify that the choice~\eqref{EqnRefinedTwoparCol}
$\tworeg$ satisfies $\tworeg \gennorm{\DelHatS}{2,1} \geq 2
|\tracer{\Wmat}{\DelHatS}|$ with high probability.  We have the upper
bound
\begin{align*}
\big| \tracer{\Wmat}{\DelHatS} \big | & \leq
\sup_{\substack{\gennorm{\Delta}{2,1} \leq \gennorm{\DelHatS}{2,1}
    \\ \frob{\Delta} \leq \frob{\DelHatS}}} |\tracer{\Wmat}{\Delta}|
\; = \; \frob{\DelHatS} \; \Ztil \bigg(
\frac{\gennorm{\DelHatS}{2,1}}{\frob{\DelHatS}} \bigg),
\end{align*}
where for any radius $t > 0$, we define the random variable
\begin{align*}
\Ztil(t) & \defn \sup_{\substack{\gennorm{\Delta}{2,1} \leq t
    \\ \frob{\Delta} \leq 1}} |\tracer{\Wmat}{\Delta}|.
\end{align*}
Following through the same argument as in Case 2 of
Appendix~\ref{AppCarefulSparse} yields that for any fixed $t > 0$, we
have
\begin{align*}
\Ztil(t) & \leq \plaincon \: \frac{\noisevar}{\sqrt{ \mdima \mdimb}}
\; t \sqrt{2 + \log \biggr( \frac{2 \mdimb}{t^2} \biggr)} + 4 \noisevar \;
\frac{t}{\sqrt{\mdimb}} + 2 \noisevar t \, \sqrt{\frac{\log
    (\frac{\mdimb}{\kdim})}{\mdima \mdimb}}
\end{align*}
with high probability.  As before, this can be extended to a uniform
bound over $t$ by a peeling argument, and we conclude that
\begin{align*}
\big| \tracer{\Wmat}{\DelHatS} \big | & \leq \; = \; \frob{\DelHatS}
\; \Ztil \biggr( \frac{\gennorm{\DelHatS}{2,1}}{\frob{\DelHatS}} \biggr) \\
& \leq \plaincon \gennorm{\DelHatS}{2,1} \; \big \{
\frac{\noisevar}{\sqrt{ \mdima \mdimb}} \; \sqrt{2 + \log \bigg(
  \frac{2 \mdimb}{\gennorm{\DelHatS}{2,1}^2/\frob{\DelHatS}^2} \bigg)}
+ 4 \noisevar \; \frac{1}{\sqrt{\mdimb}} + 2 \noisevar \,
\sqrt{\frac{\log (\frac{\mdimb}{\kdim})}{\mdima \mdimb}} \big \}
\end{align*}
with high probability.  Since
$\frac{1}{\gennorm{\DelHatS}{2,1}^2/\frob{\DelHatS}^2} \leq \frac{1}{\kdim}$
by assumption, the claim follows.


\section{Proof of Lemma~\ref{LemTwoLowRank}}
\label{AppLemTwoLowRank}

Since $\LowRankHat$ and $\TrueLowRank$ are optimal and feasible
(respectively) for the convex program~\eqref{EqnTwoStepLowRank}, we
have
\begin{equation*}
  \frac{1}{2}\frob{Y - \LowRankHat - \SparseHat}^2 + \regpar
  \nuclear{\LowRankHat}\; \leq \frac{1}{2} \frob{Y - \TrueLowRank -
    \SparseHat}^2 + \regpar \nuclear{\TrueLowRank}.
\end{equation*}
Recalling that $Y = \TrueLowRank + \TrueSparse + W$ and re-writing in
terms of the error matrices $\DelHatS = \SparseHat - \TrueSparse$ and
$\DelHatL = \LowRankHat - \TrueLowRank$, we find that
\begin{align*}
  \frac{1}{2}\frob{\DelHatL + \DelHatS - W}^2 + \regpar
  \nuclear{\TrueLowRank + \DelHatL }\; \leq \frac{1}{2} \frob{\DelHatS
    - W}^2 + \regpar \nuclear{\TrueLowRank}.
\end{align*}
Expanding the Frobenius norm and reorganizing terms yields
\begin{equation*}
  \frac{1}{2} \frob{\DelHatL}^2 \; \leq \; |\tracer{\DelHatL}{\DelHatS
    + \Wmat}| + \regpar \big \{ \nuclear{\TrueLowRank} - \regpar
  \nuclear{\TrueLowRank + \DelHatL} \big \}.
\end{equation*}
From Lemma 1 in the paper~\cite{NegWai09}, there exists a
decomposition $\DelHatL = \DelHatLA + \DelHatLB$ such that the rank of
$\DelHatLA$ upper-bounded by $2\,\rdim$ and
\begin{equation*}
  \nuclear{\TrueLowRank} - \nuclear{\TrueLowRank + \DelHatLA +
    \DelHatLB} \leq 2 \sum_{j = \rdim+1}^\mdim \sigma_j(\TrueLowRank)
  + \nuclear{\DelHatLA} - \nuclear{\DelHatLB},
\end{equation*}
which implies that 
\begin{align*}
\frac{1}{2} \frob{\DelHatL}^2 & \leq |\tracer{\DelHatL}{\DelHatS +
  \Wmat}| + \regpar \big \{\nuclear{\DelHatLA} - \nuclear{\DelHatLB}
\big \} + 2 \regpar \sum_{j = \rdim+1}^\mdim \sigma_j(\TrueLowRank) \\
& \stackrel{(i)}{\leq} |\tracer{\DelHatL}{\DelHatS}| +
|\tracer{\DelHatL}{\Wmat}| + \regpar \nuclear{\DelHatLA} - \regpar
\nuclear{\DelHatLB} + 2 \regpar \sum_{j = \rdim+1}^\mdim
\sigma_j(\TrueLowRank) \\
& \stackrel{(ii)}{\leq} \frob{\DelHatL} \; \delta +
\nuclear{\DelHatL}\opnorm{\Wmat} + \regpar \nuclear{\DelHatLA} -
\regpar \nuclear{\DelHatLB} + 2 \regpar \sum_{j = \rdim+1}^\mdim
\sigma_j(\TrueLowRank) \\
& \stackrel{(iii)}{\leq} \frob{\DelHatL} \; \delta + \opnorm{\Wmat}
\big \{ \nuclear{\DelHatLA} + \nuclear{\DelHatLA} \big \} + \regpar
\nuclear{\DelHatLA} - \regpar \nuclear{\DelHatLB} + 2 \regpar \sum_{j
  = \rdim+1}^\mdim \sigma_j(\TrueLowRank) \\
& = \frob{\DelHatL} \; \delta + \nuclear{\DelHatLA} \big \{
\opnorm{\Wmat} + \regpar \} + \nuclear{\DelHatLB} \big \{
\opnorm{\Wmat} - \regpar \big \} + 2 \regpar \sum_{j = \rdim+1}^\mdim
\sigma_j(\TrueLowRank),
\end{align*}
where step (i) follows by triangle inequality; step (ii) by the
Cauchy-Schwarz and H\"{o}lder inequality, and our assumed bound
$\frob{\DelHatS} \leq \delta$; and step (iii) follows by substituting
\mbox{$\DelHatL = \DelHatLA + \DelHatLB$} and applying triangle
inequality.

Since we have chosen $\regpar \geq \opnorm{\Wmat}$, we conclude that
\begin{align*}
\frac{1}{2} \frob{\DelHatL}^2 & \leq \frob{\DelHatL} \; \delta + 2
\regpar \nuclear{\DelHatLA} + 2 \regpar \, \sum_{j = \rdim+1}^\mdim
\sigma_j(\TrueLowRank) \\
& \leq \frob{\DelHatL} \; \delta + 2 \regpar \sqrt{2 \rdim}
\frob{\DelHatL} + 2 \regpar \, \sum_{j = \rdim+1}^\mdim
\sigma_j(\TrueLowRank)
\end{align*}
where the second inequality follows since $\nuclear{\DelHatLA} \leq
\sqrt{2 \rdim} \frob{\DelHatLA} \leq \sqrt{2 \rdim} \frob{\DelHatL}$.
We have thus obtained a quadratic inequality in $\frob{\DelHatL}$, and
applying the quadratic formula yields the claim.


\bibliography{mjwain_super}


\end{document}